\documentclass[lettersize,journal]{IEEEtran}
\IEEEoverridecommandlockouts
\usepackage{cite}
\usepackage{amsmath,amssymb,amsfonts,amsthm}
\usepackage{algorithmic}
\usepackage{graphicx}
\usepackage{textcomp}
\usepackage{comment}
\usepackage{url}
\usepackage{authblk}
\usepackage{array}
\usepackage{multirow}
\usepackage{subfigure}
\usepackage[ruled,linesnumbered]{algorithm2e}

\usepackage[table]{xcolor}
\usepackage{booktabs}

\usepackage{enumitem} 

\usepackage[hyphenbreaks]{breakurl}

\def\BibTeX{{\rm B\kern-.05em{\sc i\kern-.025em b}\kern-.08em
    T\kern-.1667em\lower.7ex\hbox{E}\kern-.125emX}}
    
\begin{document}

\title{Industrial Internet of Things Intelligence Empowering Smart Manufacturing: A Literature Review}
\author{Yujiao Hu, \textit{Member, IEEE}, Qingmin Jia, Yuao Yao, Yong Lee, Mengjie Lee, Chenyi Wang, Xiaomao Zhou, Renchao Xie, \textit{Senior Member, IEEE}, F. Richard Yu, \textit{Fellow, IEEE}

\thanks{
Corresponding authors: Qingmin Jia, Yuan Yao

Yujiao Hu, Qingmin Jia and Xiaomao Zhou are with Future Network Research Center, Purple Mountain Laboratories, Nanjing 211111, China (email: huyujiao@pmlabs.com.cn, jiaqingmin@pmlabs.com.cn, zhouxiaomao@pmlabs.com.cn).
Yuao Yao and Mengjie Lee are with the Department of Computer Science, Northwestern Polytechnical University, Xi'an 710029, China (email: yaoyuan@nwpu.edu.cn, lmj1023@mail.nwpu.edu.cn).
Yong Lee is with School of Mechanical and Electronic Engineering, Wuhan University of Technology, Wuhan 430070, China (email: yonglee@whut.edu.cn).
Chenyi Wang is with the Department of Computer Science, Peking University, Beijing, China 100091 (email: wangchenyi@stu.pku.edu.cn)
Renchao Xie is with the State Key Laboratory of networking and Switching Technology, Beijing University of Posts and Telecommunications, Beijing 100876, China (email:renchao\_xie@bupt.edu.cn).
F. Richard Yu is with the Department of Systems and Computer Engineering, Carleton University, Ottawa, Canada (e-mail: richard.yu@carleton.ca).

This work was supported in part by the National Natural Science Foundation of China (Grant Number: 92267301, 92367104), the Purple Mountain Talents-Jiangning Baijia Lake Plan Program (Grant Number: 74072203-3), the Natural Science Foundation of Hubei Province (Grant Number: 2023AFB128) and the Teaching Research Project of Wuhan University of Technology (Grant Number: W2022093). 

Copyright (c) 20xx IEEE. Personal use of this material is permitted. However, permission to use this material for any other purposes must be obtained from the IEEE by sending a request to pubs-permissions@ieee.org.
}
}

\markboth{Journal of \LaTeX\ Class Files,~Vol.~14, No.~8, August~2021}%
{Shell \MakeLowercase{\textit{et al.}}: A Sample Article Using IEEEtran.cls for IEEE Journals}

\maketitle

\begin{abstract}
The fiercely competitive business environment and increasingly personalized customization needs are driving the digital transformation and upgrading of the manufacturing industry. IIoT intelligence, which can provide innovative and efficient solutions for various aspects of the manufacturing value chain, illuminates the path of transformation for the manufacturing industry.	
It's time to provide a systematic vision of IIoT intelligence. However, existing surveys often focus on specific areas of IIoT intelligence, leading researchers and readers to have biases in their understanding of IIoT intelligence, that is, believing that research in one direction is the most important for the development of IIoT intelligence, while ignoring contributions from other directions.
Therefore, this paper provides a comprehensive overview of IIoT intelligence. We first conduct an in-depth analysis of the inevitability of manufacturing transformation and study the successful experiences from the practices of Chinese enterprises. Then we give our definition of IIoT intelligence and demonstrate the value of IIoT intelligence for industries in fucntions, operations, deployments, and application. Afterwards, we propose a hierarchical development architecture for IIoT intelligence, which consists of five layers. The practical values of technical upgrades at each layer are illustrated by a close look on lighthouse factories. Following that, we identify seven kinds of technologies that accelerate the transformation of manufacturing, and clarify their contributions. \textcolor{black}{The ethical implications and environmental impacts of adopting IIoT intelligence in manufacturing are analyzed as well.} Finally, we explore the open challenges and development trends from four aspects to inspire future researches.  
\end{abstract}


\section{Introduction}

Manufacturing holds an important position in the global economy. It accounts for 16\% of global GDP, and its research and development (R\&D) spend accounts for 64\% of global R\&D expenditure \cite{theproportionmanufacturing}. However, manufacturing is suffering from many troubles. The stagnant productivity, increasingly personalized customization requirements, long overdue innovation, rapidly shrinking humanpower, and expectation for one-stop comfortable service, all of these are pushing manufacturing to upgrade and transform. At the same time, the rapid development of intelligent technologies in recent year also inspires governments and enterprises to carry out digitization transformation. Under the circumstance, governments issued a series of policies \cite{AmericanPolicy,JapanesePolicy,GermanPolicy,MadeInCHina2025}, such as \textit{New Robot Strategy}, \textit{Industrial Strategy 2030}, etc. Inspired by the policy dividend, enterprises begin to take the path of industrial revolution and innovation \cite{QingdaoHaier5GSmartFactory}\cite{MideaChinaUnicomandHuawei}. 

Industrial Internet of Things (IIoT), which makes all types of industrial equipment to be connected together through networks and can further support data collection, exchange and analysis \cite{qiu2020edge}, have realized great achievements for improve the digitalization level of industries. To make the research values of IIoT clear, some works surveyed the various definitions of IIoT and enabling technologies, and proposed open challenges to inspire future research directions. 
Boyes \textit{et al.} \cite{boyes2018industrial} differentiated the connotation of IIoT with industry 4.0, cyber physical system, industrial automation \& control systems, supervisory constrol and data acquisition, industrial internet, and emphasized \textit{interconnection} among smart objects, cyber-physical assets and computing platforms.  
Khan \textit{et al.} \cite{khan2020industrial} reviewed the research efforts in the areas of architectures and frameworks, communication protocols and data management schemes for IIoT. The general architecture for IIoT system in \cite{khan2020industrial} includes networked entities, cloud/edge computing platforms, and enterprise business domains. The researches of IIoT communication protocols are highlighted. Cloud/Edge computing is used to provide distributed high performance computing resources for IIoT. 
Malik \textit{et al.} \cite{malik2021industrial} gave the architecture of IIoT, which consists of physical layer, data link layer, network and Internet layer, transport layer and application layer, and analyzed the potential applications of IIoT, including environment monitoring, smart agriculture, healthcare, supply chain management and home automation. 
The common insight of these works for future researches is to combine more advanced technologies with IIoT, such as blockchain, big data analysis, deep learning, etc., so that IIoT can become more intelligent. 

As expected, in recent years, the integration of IIoT with many other advanced technologies makes IIoT emerge intelligence in terms of ubiquitous reliable communication, dynamic environment adaptation, flexible manufacturing, etc. This also accelerates the transformation progress of manufacturing industries. 
To be specific, when IIoT is combined with 5G, its demands of multiple device connectivity and high data rate, more bandwidth, low-latency quality of service (QoS) can be easily satisfied. Chettri \textit{et al.} \cite{8879484} and Mahmood \textit{et al.} \cite{9548837} provided detailed overview of challenges and vision of various communication industries in 5G IIoT systems. 
Time sensitive networking (TSN) is a set of IEEE 802.1 standards that defines mechanisms to provide deterministic services through IEEE 802 networks \cite{finn2018introduction,messenger2018time,farkas2018time}. Deterministic services include guaranteed packet transport with bounded latency, low packet delay variation, and low packet loss. Combined with TSN, IIoT can ensure the deterministic response for network traffics between any two devices. Bello \textit{et al.} \cite{bello2019perspective} described TSN protocols in detail and provided some cases while using TSN into IIoT. 
Deep learning has potentials to be combined with IIoT as well. Khalil \textit{et al.} \cite{9321458} presented the various deep learning techniques, including convolutional neural networks, autoencoders, and recurrent neural networks, as well as their use in different industries. 
The product R\&D cost in IIoT can be reduced through introducing digital experience \cite{wei2021consistency}\cite{gregorio2021digital}. The labour efficiency grows with the help of remote monitoring and control \cite{Remotehumanrobotcollaboration2020}. The industrial robots can gradually replace human in some physically demanding and routine jobs \cite{Germanrobots2017}\cite{Evolutionindustrialrobots2013}. 

It's time to have a systematic understanding for IIoT intelligence. However, existing surveys always focused on IIoT intelligence in specific domain, such as 5G in IIoT \cite{8879484,9548837}, TSN in IIoT \cite{bello2019perspective}, deep learning in IIoT \cite{9321458,ambika2020machine,de2022deep}, edge computing in IIoT \cite{qiu2020edge,laroui2021edge,hamdan2020edge}. 
This will lead to biases among researchers/readers in their understanding of IIoT intelligence, that is, believing that the research in one direction is the most important for the development of IIoT intelligence, while neglecting the contributions of other directions. Under the circumstance, the contributions/positions of advanced technologies in the architecture of IIoT intelligence cannot be clarified, either. 

Therefore, in this paper, we provide a comprehensive overview of IIoT intelligence. Considering that the development of IIoT intelligence integrates the wisdom of governments, enterprises, and researchers, we expand the information sources from academic papers to government documents, industry reports, and other related literature. 
\textcolor{black}{Specifically, to extract insights into IIoT intelligence, we meticulously study the World Economic Forum's Global Lighthouse Networks (GLN), which recognized the lighthouse use cases and factories under the Fourth Industrial Revolution. Then we try our best to carefully study associated customer stories from official sources.
By delving into GLN and customer stories, we discern technologies pivotal for their digital and intelligent transformation. For each identified technology, we primarily examine technical papers published within last 5 years, as well as highly cited papers beyond this timeframe, to gather comprehensive insights and consolidate relevant knowledge. Finally, we present this paper to answer four questions as follows. }

\begin{itemize}
    \item What is IIoT intelligence and what can IIoT intelligence do for smart manufacturing?
    \item What is the development architecture of IIoT intelligence?
    \item Which technologies are accelerating the development of IIoT intelligence?
    \item What are the open challenges and future trends of IIoT intelligence?
\end{itemize}

In response to the first question, we conduct an in-depth analysis of the inevitability of manufacturing transformation under the pressure of customization demand, increased labor costs, and competitive environment, as well as the incentives of government policies and rapid development of advanced technology. Then, based on the practical exploration of Chinese industries, the secrets of successful transformation are revealed, including building an effective production architecture, building a customer-centric value chain, building innovation business models, building digital capabilities of manufacturing and organization. 
The secrets inspire the emergence and development of IIoT intelligence. 
We then give our definition of IIoT intelligence, i.e. a series of techniques, methods, productions and platforms that can been taken within the entire value chain of manufacturing to build autonomous capabilities. The value chain of manufacturing involves R\&D, production, operation and maintenance, marketing, management, services, etc. The built autonomous capabilities through IIoT intelligence include digital connection and perception, intelligent analysis and cognition, immediate decisions, etc. Combining with practice, we demonstrate the value of IIoT intelligence for manufacturing industries in functions, operations, deployments and application scenarios. 

To answer the second question, we propose a hierarchical development architecture to build a system-level understanding about how IIoT intelligence empowers smart manufacturing. Specifically, the architecture consists of five layers. Each layer has own missions. The equipment layer builds the basis of automatic production. The network layer connects human-machine-things, and overcomes isolated islands of information. The software layer provides digital representation of industrial processes. The modeling layer makes digital twin for physical processes to connect virtual and physical spaces. The analysis and optimization layer is responsible for mining industrial big data and optimizing industrial processes. 
Each layer of IIoT intelligence has its own key technologies, making its capabilities stronger. Interactions will occur between levels to fully enhance the intelligence and digitization level of the manufacturing industry. 
In addition, we also closely observe the lighthouse factory \cite{IndustrialRevolution4,globalLighthouse,globalLighthouse2022,globalLighthouse2023} to demonstrate that technological upgrades at each layer enhance the company's returns.

To deal with the third question, we identify seven kinds of technologies and clarify their contributions in accelerating the development of IIoT intelligence. To be specific, (a)~industrial robots can become intelligent labors; (b)~machine vision systems have the potential to be the eyes of industries; (c)~networking facilitates ubiquitous connectivity; (d)~digital twins connects cyber spaces and physical spaces; (e)~deep learning boosts intelligence; (f)~smart hardware builds foundation of intelligence; (g)~cloud/edge computing inspires novel business and deployment. 

To reply to the fourth question, we explore open challenges and future trends in four directions, based on the vision of the future manufacturing industry, including \textit{digital control, deterministic response, cost-friendly operation and deployment, IIoT intelligence proliferation}. 

The main contributions of this paper can be summarized as follows. 
\begin{itemize}
    \item This paper provides a comprehensive literature review on IIoT intelligence to establish a systematic understanding of the positive impact of IIoT intelligence on smart manufacturing.
    \item This paper proposes a hierarchical development architecture for IIoT intelligence and combines it with practical lighthouse factories to demonstrate the research value of various levels in improving enterprise returns. 
    \item This paper identify seven kinds of non-trivial technologies and clarify their contributions in accelerating the development of IIoT intelligence. 
    \item This paper discovers open challenges and predicts future trends of IIoT intelligence to inspire future researches. 
\end{itemize}

This paper is structured as follows. 
Section~\ref{section:smartmanufacturing} presents background and transformation secrets of smart manufacturing. 
The connotation of IIoT intelligence is explained in Section~\ref{section:industrialIntelligence}. 
Section~\ref{section:hierarachical-architecture} provides a reference hierarchical development architecture to build a systematic understanding for IIoT intelligence.
Section~\ref{section:closelook} demonstrates the significant effects of IIoT intelligence through case studies on lighthouses and gradually released collaborations.
Section~\ref{section:Technology-Study} identities some of the most influential technologies that facilitate IIoT intelligence. 
Section~\ref{section:challenges} summarizes open challenges and future trends of IIoT intelligence.
\textcolor{black}{Section~\ref{section:ethical} presents the ethical implications and environmental impacts of adopting IIoT intelligence in manufacturing. }
The conclusion is given in Section~\ref{conclusion}.

\section{In-depth Perspectives on Smart Manufacturing} \label{section:smartmanufacturing}
In recent years, the Forth Industrial Revolution has attracted attentions of governments, enterprises and universities, and has been fully launched. Many manufacturers are exploring the way to intelligent transformation. 

\subsection{Background}
In this section, we will answer one important question, \textit{why the manufacturing industry needs to undergo intelligent transformation}, based on the factors in social environment and technical development. 

\begin{figure}
	\centerline{\includegraphics[width=0.4\textwidth]{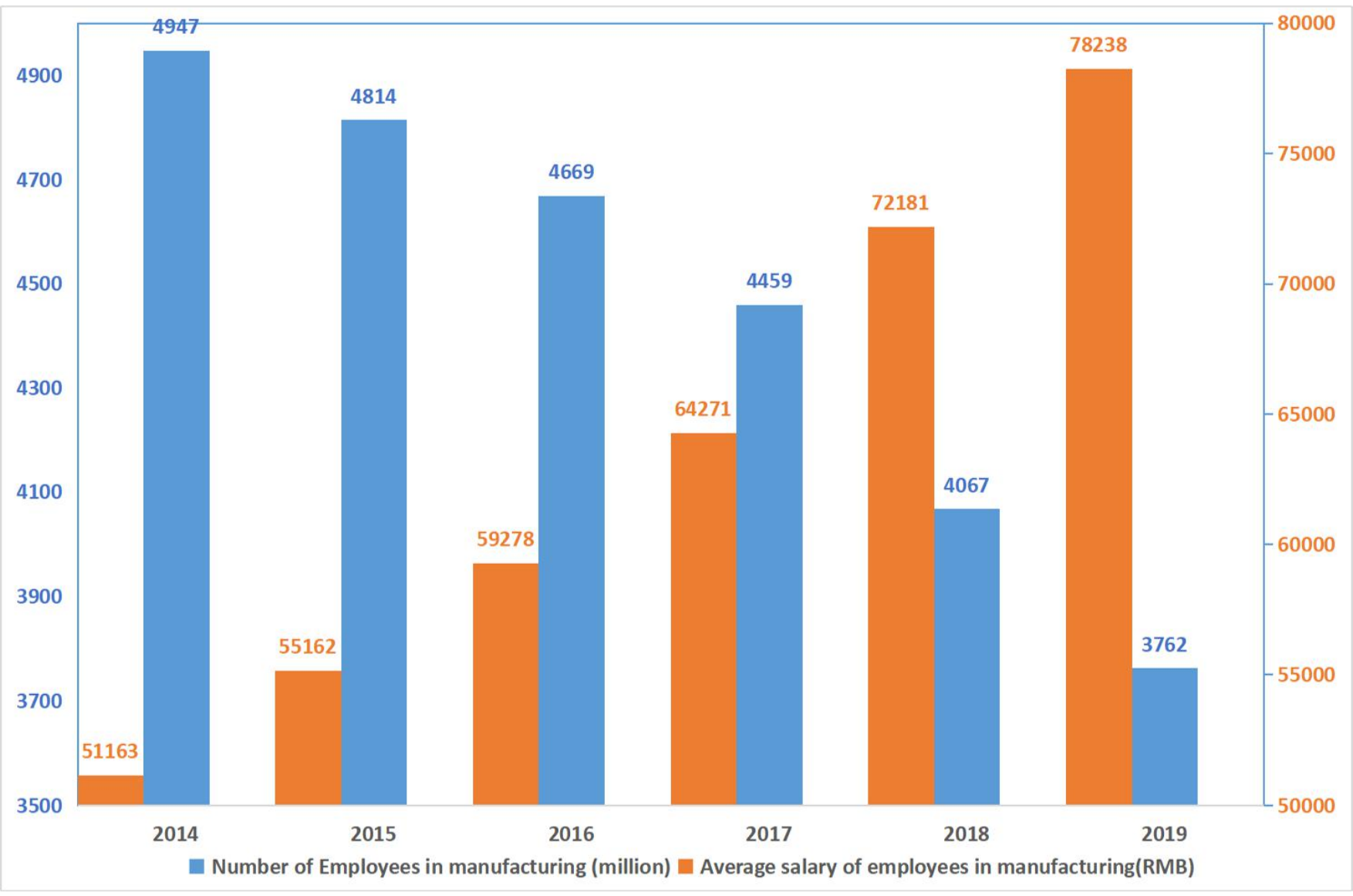}}
	\caption{The workforce in Chinese manufacturing has been falling from 2014 to 2019, while workforce costs have been rising significantly.  
	}
	\label{fig:ChinesePopularityCost}
\end{figure}

\begin{figure*}
	\centerline{\includegraphics[width=0.85\textwidth]{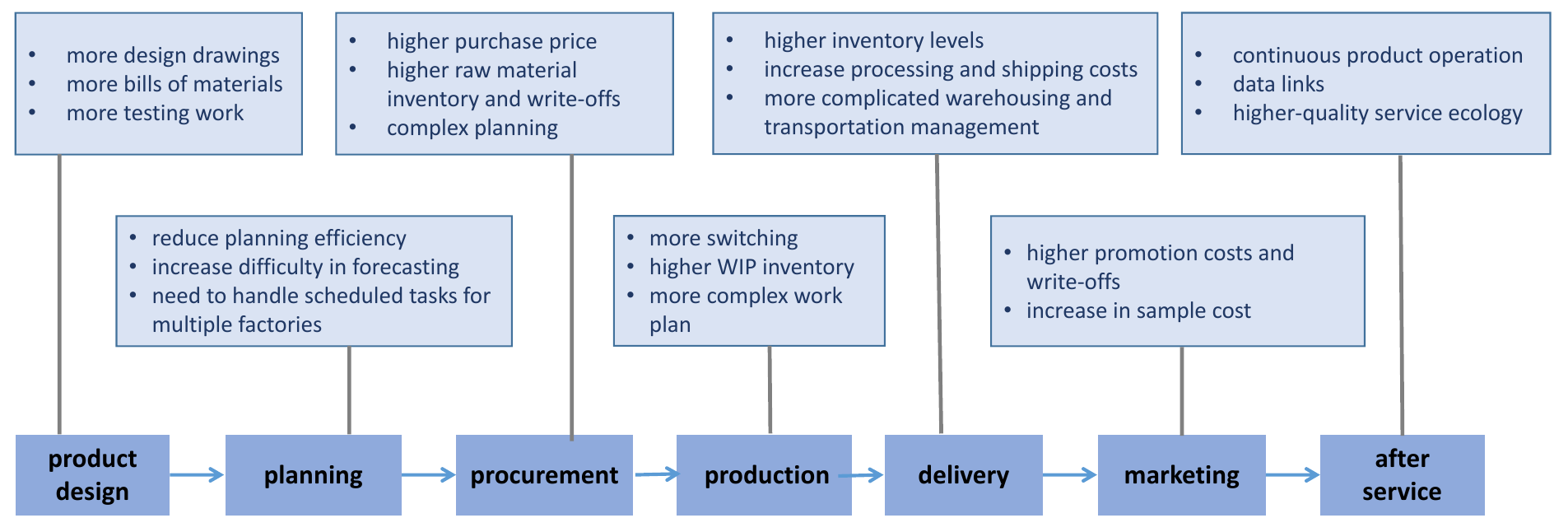}}
	\caption{Personalized customization and patient experience enhance complexity of the architectures for manufacturing systems.
	}
	\label{fig:complexityEnhancement}
\end{figure*}

In recent years, the population of many countries increases slowly, and meanwhile the workforce costs of manufacturing industries are rising. For example, according to the results of the \textit{Chinese National Bureau of Statistics}\cite{ChinesePopularity} in Fig.~\ref{fig:ChinesePopularityCost}, the workforce in manufacturing has been falling from 2014 to 2019, while workforce costs have been rising significantly. 
In addition, more and more customers are pursuing personalized customization and patient experience, which makes the architecture of manufacturing systems become more complex at every phase of end-to-end product life cycle throughout designing, production, and after service, as shown in Fig.~\ref{fig:complexityEnhancement}.
The capabilities of mass-producing customized products on demand and reacting rapidly to demand fluctuation will provide the essential competitive for companies. However, there are some challenges to build the capabilities, including adjusting the organizational structure,  operating, and business modes. 
In general, the rising workforce costs and customization demands force the manufacturing community to embark intelligent transformation, so as to increase productivity and realize appealing returns. 

Recent years have witnessed the rapid development of a large number of advanced intelligent technologies, including  cloud computing, 5G connection, industrial data analysis, industrial internet of things, etc. These technologies light the way for upgrading manufacturing. 
\textcolor{black}{Many countries worldwide are also actively supporting the intelligent transformation of manufacturing through policy initiatives. These efforts encompass various aspects such as promoting research and development, fostering innovation ecosystems, incentivizing investment in advanced technologies, and addressing regulatory barriers. For example, on October 30 2023, the Biden Administration released Executive Order (E.O.) 14110 on \textit{Safe, Secure, and Trustworthy Development and Use of Artificial Intelligence}. It establishes a government-wide effort to guide responsible artificial intelligence (AI) development and deployment through federal agency leadership, regulation of industry, and engagement with international partners \cite{AmericaAI2023Whitehouse,AmericaAI2023highlights}.  On February 1st 2023, European commission released \textit{Green Deal Industrial Plan for the Net-Zero Age} \cite{Europ2023GreenIndPlan}, and following with \textit{Net-Zero Industry Act} on February 6 2024 \cite{Europ2024NetIndustrialPlan}, to help the European Union (EU) become home to clean technologies and make significant strides towards building a strong domestic manufacturing capacity of those technologies in the EU. On August 23, 2023, China released implementation guidelines as part of standards for new emerging industries, vowing to continuously improve the technical level and internationalization of new industry standards, and to provide solid technical support for accelerating the high-quality development of new industries by 2035 \cite{ChinaStandarad2035,ChinaNewplan2023}.  
}

\subsection{Secrets to smart manufacturing} \label{section:secrets2SM}
\begin{figure*}[h!]
	\centerline{\includegraphics[width=0.85\textwidth]{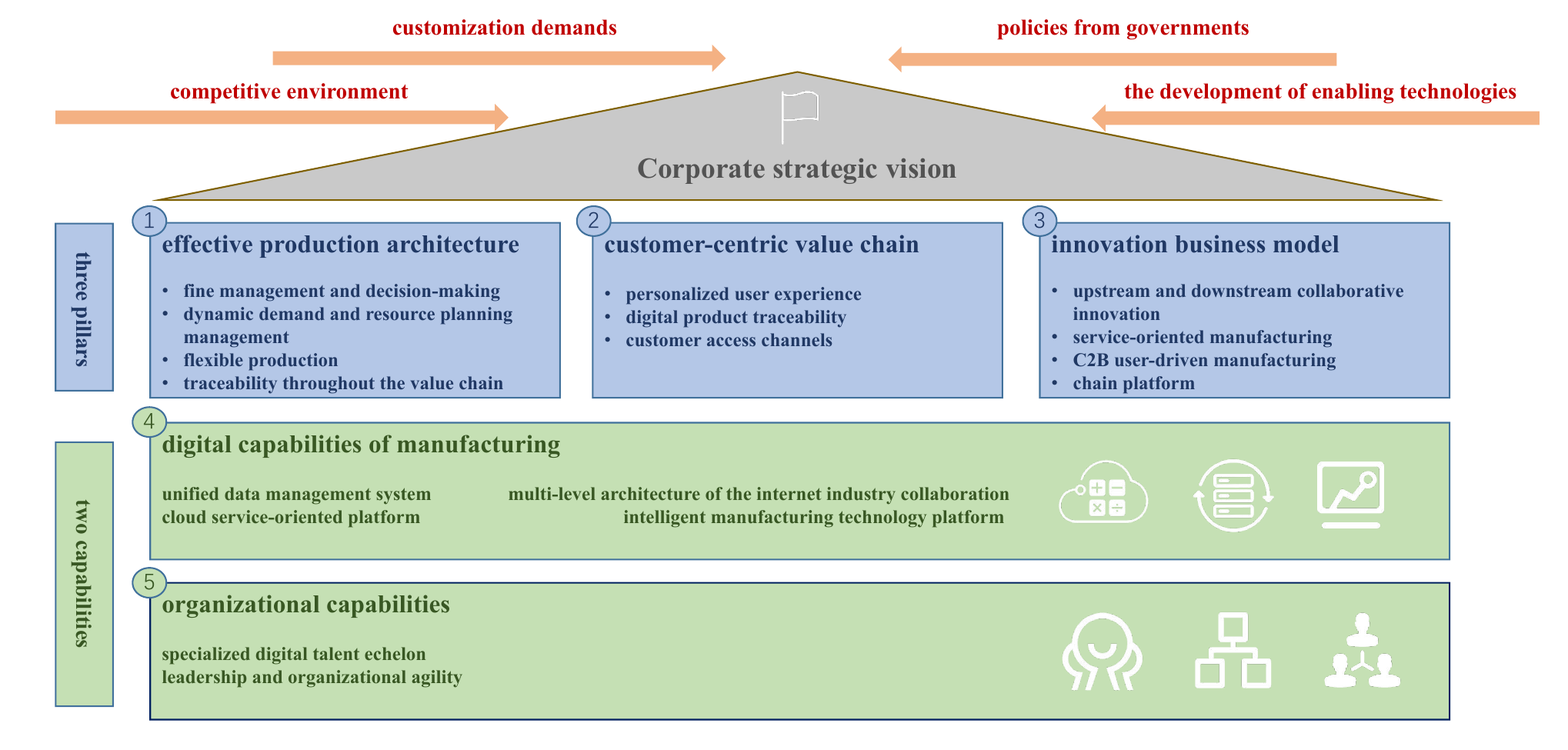}}
	\caption{The secrets to intelligent transformation of manufacturing}
	\label{fig:secrets-of-transformation}
\end{figure*}

In this section, we will answer the question, \textit{how to accelerate the progress of intelligent transformation for manufacturing industry}, based on the practical experiences of  \textit{Foxconn Industrial Internet}, which is a famous enterprise that has realized a high-level intelligent transformation in China. 
\textcolor{black}{
Please note that many other manufacturers, such as \textit{DHL Supply Chain, Siemens, and Unilever} in America, Germany, and India, respectively, have undergone similar transformation journeys as \textit{Foxconn Industrial Internet}. In this context, we draw insights from the experiences of \textit{Foxconn Industrial Internet} to illustrate the secrets to smart manufacturing. \textit{Foxconn Industrial Internet} has summarized its recipes into three pillars and two capabilities \cite{WhitePaperofFoxconn}, as presented in Fig.~\ref{fig:secrets-of-transformation}. }
\begin{itemize}
	\item  Building an effective production architecture to achieve intelligent transformation. Specifically, companies should improve flexible automated production lines, connect production data, and promote intelligent applications. As a result of these efforts, companies will improve their yield and throughput, energy efficiency and product quality, and reduce product cost, inventory, operating cost.  
	\item Building a customer-centric value chain. Companies are expected to develop new products and provide supporting service on the basis of accurately understanding customer needs. Enterprises may focus on optimizing different value chains, such as the value chains from suppliers to marketing, from enterprise management to after-sales service, from product development to consumers, etc.
	\item Building innovation business models to enable digital manufacturing ecosystem through standardized collaboration models with academia, industrial partners and government, such as technology ecosystem \cite{technologyEcosystem}, investment binding. 
	\item Building digital capabilities of manufacturing by combining traditional and new technologies. Although these technologies are at different development phases, they can solve particular problems in the processes and operation.
	\item Building capabilities proactively, both technical and soft, and managing talent. Companies can tailor the skills training programs to the entire organization, from front-line worker to managing director, based on their digital affinities. Meanwhile, building an agile architecture for  organization and performance management can support intelligent transformation and technology innovation as well.
\end{itemize}


\section{IIoT intelligence  to support manufacturing transformation} \label{section:industrialIntelligence}

IIoT intelligence can support the construction of three pillars and two capabilities required by manufacturing transformation, which are explain in Section~\ref{section:secrets2SM}. In the following subsections, we will explain the connotation of IIoT intelligence and its effects on smart manufacturing.

\subsection{Understanding IIoT intelligence }
We give our definition for IIoT intelligence:

\textit{IIoT intelligence refers to a series of techniques, methods, productions and platforms that can been taken throughout the entire value chain to build capabilities of digital connection and perception, intelligent analysis and cognition, real-time decision-making, etc. }

The value chain of manufacturing involves research and development (R\&D), production, operation and maintenance, marketing, management, services, etc. All segments of the value chain are the launching scenarios of the IIoT intelligence. 
That is, the essence of IIoT intelligence is to realize innovative industrial applications, such as digital R\&D, efficient and immediate decisions, rapid line reconfiguration, through in-depth combining intelligent technologies with industrial scenarios, mechanisms and knowledge.

IIoT intelligence is improving by iterating continuously.
At the beginning, data throughout the value chain is collected and some machine learning algorithms are used to learn specific knowledge to all segments of the value chain. Then, based on the data and knowledge, optimized models and algorithms are designed and verified. Afterwards, the optimized solution is scaled within the value chain. Finally, new data will be generated, a new round that improves the value chain will start.  

\subsection{How does IIoT intelligence achieve impacts in manufacturing?}
IIoT intelligence can provide some innovative solutions for segments of the value chain of manufacturing, and support to build the pillars and capabilities (shown in Fig.~\ref{fig:secrets-of-transformation}). 
\textcolor{black}{In the following sections, we delve into three case studies to illustrate the impacts of IIoT intelligence on smart manufacturing, focusing on economic, technical, and organizational aspects.}
\textcolor{black}{
\subsubsection{Economic impacts -- reducing product R\&D costs through digital experiments}
The traditional processes of the R\&D of products require several rounds of trial production, testing and optimization in the physical production lines. Each round of the physical experiment will cost a lot of manpower, material, money and time. Therefore, the traditional R\&D process is time-consuming, expensive and environmentally unfriendly. }

\textcolor{black}{
IIoT intelligence can play a crucial role in reducing the costs of product R\&D by digitizing a portion of physical experiments. Leveraging historical data and experiences, IIoT maps industrial processes into the digital realm, enabling cost-effective trial and error experiments. Moreover, the digital environment facilitates the automatic prediction of potential faults based on industrial knowledge and historical data.
Recommendation technologies contribute by offering improvement suggestions, allowing the project team to promptly revise and optimize product designs in response to detected or predicted faults. This streamlined process enhances efficiency and cost-effectiveness in the development of new products.}

\textcolor{black}{
In addition to optimizing the product development process and low-cost trial and error in the digital environment, IIoT intelligence can also support generative design in the product development stage, cross-departmental and cross-regional collaborative innovation and cooperation, thereby speeding up the design process and reducing design costs. }
\textcolor{black}{
\subsubsection{Technical impacts -- enhanced inspection efficiency through automated visual systems}
Prior to the integration of intelligent technologies in manufacturing quality control, inspections relied on human inspectors for manual defect identification during visual inspections. However, this method had several drawbacks. Human inspectors may struggle to maintain consistent attention and precision over extended periods.  Additionally, manual inspection processes may not match the speed of automated systems, resulting in slower production speeds and reduced efficiency. }

\textcolor{black}{
IIoT intelligence tackles these challenges through automated visual inspection. Cameras and sensors strategically positioned along the production line capture images and gather data on product attributes. Machine vision systems, fueled by AI algorithms, process these images. Trained to recognize components, detect defects, and assess assembly quality, the algorithms prompt corrective actions when defects are found. The benefits of automated visual inspection include enhanced accuracy, a decrease in defect rates, and accelerated inspection processes, contributing to overall production line efficiency.
}
\textcolor{black}{
\subsubsection{Organizational behaviour impacts -- increase labour efficiency through remote monitoring and control}
Traditionally, workers are required to be physically present in the production area to actively engage in manufacturing processes or oversee equipment operations. This hands-on approach is essential for promptly identifying and rectifying errors during production. Additionally, equipment maintenance is typically carried out through on-site services provided by technicians. The deployment of production lines is often based on manual analysis and the expertise of industry professionals. }

\textcolor{black}{
IIoT technologies can connect and perceive information across equipment, workers, and intermediate products in workshops. This enables remote monitoring of production conditions and control of manufacturing equipment, enhancing safety through unmanned operations and reducing exposure to hazardous sites. The technology improves Overall Equipment Effectiveness (OEE) and labor efficiency by transitioning from \textit{one-man-one-control} to a more efficient \textit{one-man-multi-control} approach.
Remote monitoring and control allow managers to adapt worker schedules based on real-time progress. Additionally, these technologies create new white-collar job opportunities, enabling workers to operate from comfortable offices. This enhances the appeal of the manufacturing industry, addressing labor shortages by offering an alternative and attractive working environment. }

\subsection{The overall effect of IIoT intelligence  on smart manufacturing}
In summary, IIoT intelligence can promote the rapid development of smart manufacturing from the perspectives of functions, operation, deployment and application scenarios.
\begin{itemize}
	\item Functions: IIoT intelligence can help companies connect human-cyber-physical spaces, collect industrial data synchronously and accurately, build effective data analysis models, propose scenario-oriented optimization methods, develop microservice architecture, etc. \textcolor{black}{For example, according to the lighthouse collections from the World Economic Forum \cite{globalLighthouse2023} in 2023, through adopting IIoT intelligence, \textit{ACG Capsules} in India realizes real-time quality batch insights and enhanced production planning and scheduling, which reduces 39\% of batch lead time and improve 13\% of on-time delivery in full. }
	\item Operation: IIoT intelligence builds cloud manufacturing business model for enterprises to enhance returns \cite{xu2012cloud,liu2017workload,bello2021cloud}. Cloud manufacturing create intelligent factory networks that encourage effective collaboration, through encouraging enterprises to encapsulate distributed manufacturing resources into cloud services and managing them in a centralized way. 
 \textcolor{black}{For example, \textit{China Resources Building Materials Technology} adopts cloud to enrich and make data easy to consume across the organization to improve business performance. \textit{Agilent Technologies} in Germany utilizes cloud to achieve predictive quality testing with AI, leading to a significant 13\% improvement in test station throughput \cite{globalLighthouse2023}. }
	\item Deployment: IIoT intelligence provides the cloud-edge-terminal architecture to deploy computing resources in a cost-friendly way. \textcolor{black}{Except that, IIoT intelligence can give efficient production deployment/scheduling suggestions to improve factory output and energy optimization. For example, \textit{ACG Capsules} optimizes its production schedule with a novel color-matching AI and validates it with a digital twin, which realizes 10\%-20\% on-time-in-full increase. }
	\item Application scenarios: IIoT intelligence can be used to solve specific problems in some application scenarios, such as predictive and remote maintenance, product and equipment monitoring, product design feedback optimization, product process traceability, renting manufacturing capacity, product quality inspection, etc. \textcolor{black}{For example, \textit{Johnson \& Johnson Consumer Health} in India deploys IIoT intelligence to achieve predictive maintenance,  resulting in improved asset reliability and a 50\% reduction in unplanned machine downtime \cite{globalLighthouse2022}.}
\end{itemize}

\section{Hierarchical development architecture of IIoT intelligence } \label{section:hierarachical-architecture}
\begin{figure} 
	\centerline{\includegraphics[width=0.45\textwidth]{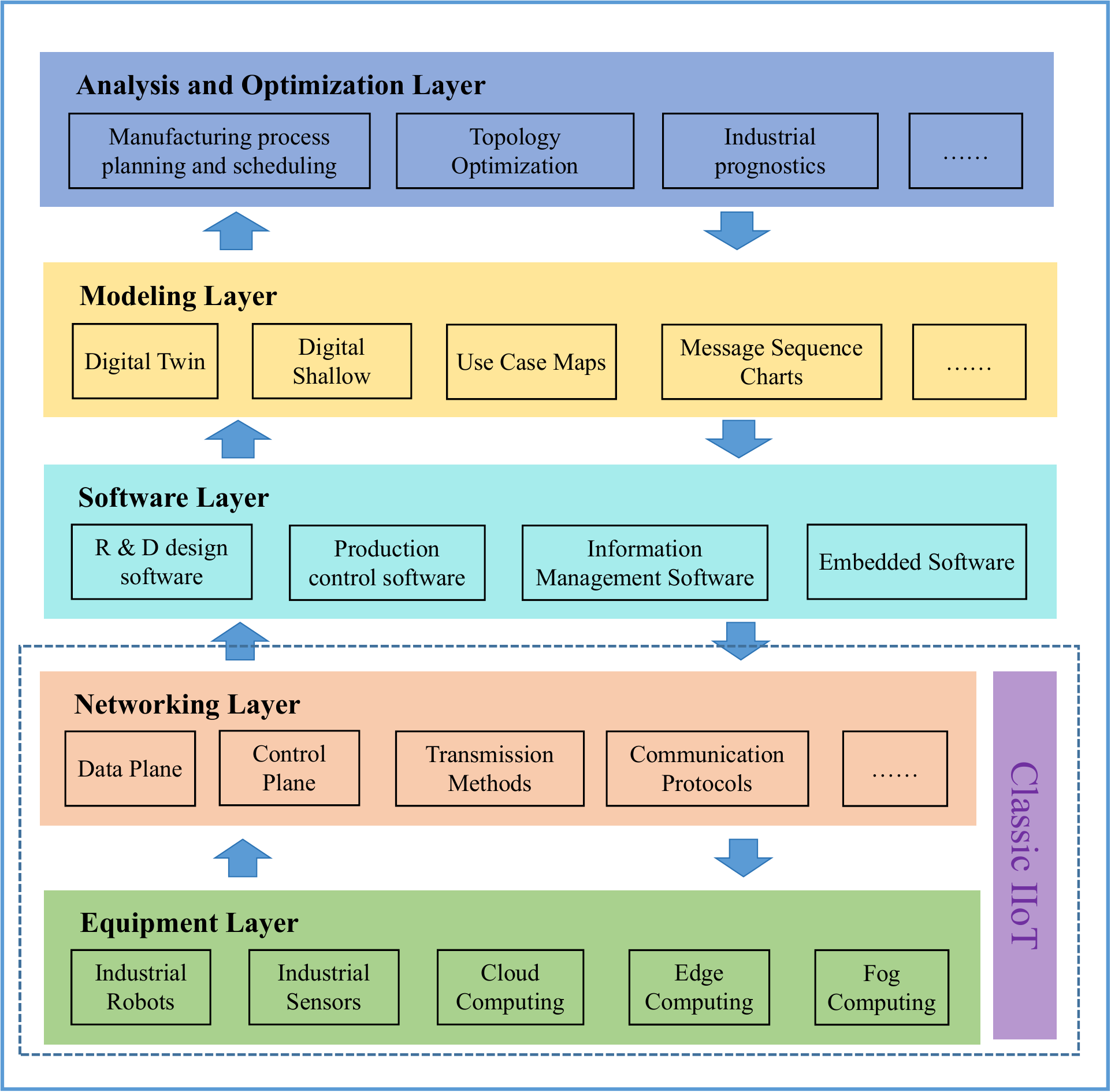}}
	\caption{The hierarchical development architecture of the IIoT intelligence.  The equipment layer builds the basis of automatic production. The network layer connects human-machine-things, and overcomes isolated islands of information. The software layer provides digital representation of industrial processes. The modeling layer makes digital twin for physical processes to connect virtual and physical spaces. The analysis and optimization layer is responsible for mining industrial big data and optimizing industrial processes. }
	\label{fig:hierarchicaldevelopment}
\end{figure}

To build a systematic understanding of how IIoT intelligence deals with issues in the transformation process of smart manufacturing, this paper presents a hierarchical development architecture for IIoT intelligence. The architecture consists of equipment, network, software, model and optimization from bottom to top, as shown in Fig.~\ref{fig:hierarchicaldevelopment}. The classic architecture of IIoT-enabled smart manufacturing usually include the bottom two layers \cite{boyes2018industrial,qiu2020edge,gupta2020overview,malik2021industrial}, however, it cannot bear the weight of the various intelligence that have emerged in recent years. Therefore, we extends software layer, modeling layer, and analysis and optimization layer over the classic architecture. 
The functions of each layer in the hierarchical architecture are as follows:
\begin{itemize}
	\item \textit{Equipment Layer}: Technological innovation in the equipment layer builds the basis of automatic production. Especially, industrial robots paired with smart sensors will replace human labour to achieve automatic production to some extent, or assist workers engaging in flexible production. Cloud/Edge/fog computing ensures sufficient computing resources to support the development of IIoT intelligence.  
	\item \textit{Networking Layer}: Communication technologies in the networking layer constructs a network that supports high reliability and low delay transmissions of human-cyber-physical information flows. The network can overcome isolated islands of information in the companies, which will greatly accelerates the development of smart manufacturing.
	\item \textit{Software Layer}: Industrial software provides digital expression of industrial processes by collecting and mining industrial data. Industrial software can meet the requirements of enterprises to strengthen control over various industrial processes, making industrial processes transparent, and also conducive to optimizing production and management. 
	\item \textit{Modeling Layer}: Modeling technologies in the modeling layer are expected to accurately model industrial processes and scenarios, based on the digital expression from industrial software.  At the same time, the model layer should support the analysis and optimization layer via providing the necessary data, model and management control interfaces.
	\item \textit{Analysis and Optimization Layer}: The analysis and optimization layers consists of some specific algorithms. The algorithms are designed to solve specific problems in the industrial processes based on the industrial big data coming up from the bottom and models built in the modeling layers.     
\end{itemize}

The transmissions of information flow in the hierarchical development architecture are bidirectional. 
\begin{itemize}
    \item \textit{The data is transmitted from bottom to top.} The industrial equipment and workers produce data constantly. The networks ubiquitously connecting human-machine-things transmit data to computing devices with specific industrial software. After collecting and roughly mining data, software provides required information to modeling layer, so that accurate models can be constructed. Finally, the information and models are used by the specific algorithms in the analysis and optimization layer. 
    \item \textit{The data is transmitted from top to bottom.} An optimized solution is firstly issued in the analysis and optimization layer, and then assigned to modeling layer. After modules in the modeling layer understand the solution, they will inform corresponding management and control software. Finally, the equipment layer will carry out the solutions. 
\end{itemize} 

\textcolor{black}{Compared with IIoT intelligence empowered smart manufacturing, traditional manufacturing models lack automatic controls and intelligent modeling. To be specific, traditional models often rely on manual control systems and fixed processes. Automation is limited, and adjustments to manufacturing processes typically require human intervention. This can lead to slower response times and increased susceptibility to human errors. When IIoT intelligence enabling smart manufacturing, IIoT sensors and devices are strategically deployed throughout the production process. These devices collect real-time data, enabling automatic control systems to respond dynamically to changing conditions. This results in enhanced automation, reduced manual intervention, and improved operational efficiency. Additionally, traditional models may lack sophisticated intelligent modeling capabilities. Decision-making often relies on historical data and experience, with limited ability to predict future events or proactively address issues. While IIoT intelligence provides advanced analytics and machine learning algorithms to process the vast amounts of data generated by sensors. These intelligent approaches help in forecasting maintenance needs, optimizing production schedules, and improving overall decision-making during the overall production processes. }

In the following subsections, the motivation and connotation of each layer in the hierarchical development architecture are introduced in detail. 

\subsection{Equipment layer}

\subsubsection{Motivations}

Facing the crisis of labour shortage and high labour cost, the manufacturing industry's desire for a machine that can work like or even better than humans has become increasingly strong, thus promoting the invention and innovation of industrial robots.
Except that, there are other reasons that motivate enterprises \cite{IFRPresentationPPT}. 
a)~There are many dull, dirty, dangerous and/or delicate tasks that can be done by robots, which can improve health, safety and job satisfaction of employees. 
b)~The rapid development of robotics is continuously improving the automated and flexible production capacity of robots, while also reducing the procurement and maintenance costs. 
c)~Modern robots will support a smaller carbon footprint \cite{carbonfootprintrobots2020}\cite{nilakantan2017multi}, which is profitable for building low-carbon factories according to national requirements.
d)~More and more robots support plug-and-play integration and their control programming becomes much easy, which allow operators to become proficient in using robots with general training \cite{progressonprogramming2012}. 
e)~Human-robot collaboration will strengthen flexible production \cite{vysocky2016human}\cite{villani2018survey}. 
f)~The aging society of the world has also released demands for liberation from manual labor.

Except robots, in order to achieve precise perception of the state of the production environment and high-precision control of the production process, it is necessary to introduce smart sensors to improve \textit{five senses}\footnote{taste, touch, smell, sight, hearing} of industry. In addition, with the deepening of industrial digital transformation, a large number of industrial data storage and analysis tasks have emerged. To complete these tasks, massive computing and storage resources are required.

\begin{figure*} 
	\centerline{\includegraphics[width=0.7\textwidth]{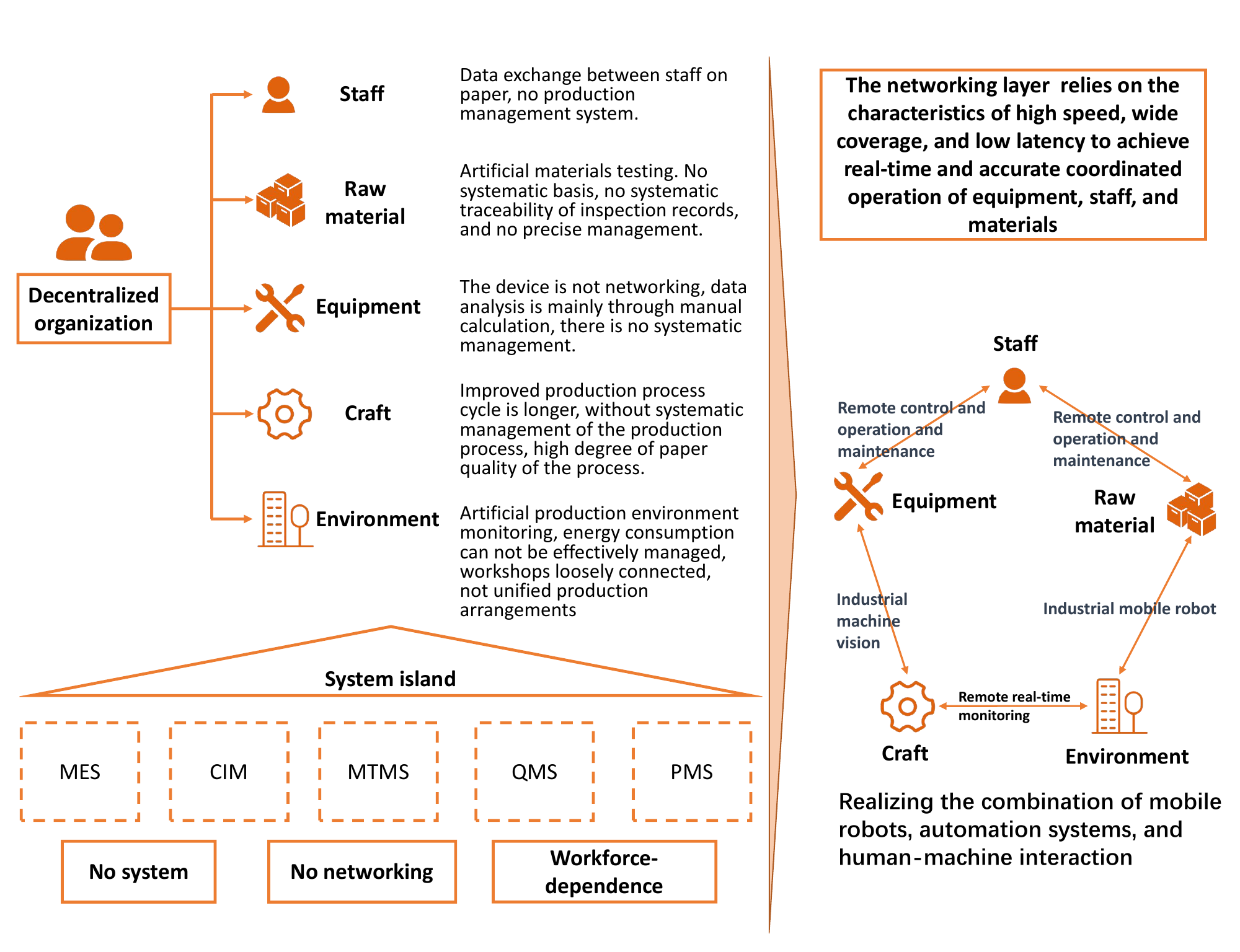}}
	\caption{The networking helps to realize smart factories from information islands to collaborative operations} 
	\label{fig:informationIsland}
\end{figure*}

\subsubsection{Connotation}
Equipment in the equipment layer can be divided into two categories, that is, production equipment, which includes industrial robots and sensors, and computation devices, which include computation resources and storage resources on cloud/edge/fog. 

Industrial sensors perceive a physical quantity at the input and send a corresponding electrical signal at the output. There are a lot of industrial sensors that can be used to improve five senses of industry, including air quality sensors, ambient light sensors, distance sensors, image sensors, industrial motion \& position sensors, infrared detectors, industrial pressure sensors, speed sensors, etc. Industrial sensors can be found in nearly every modern manufacturing process, regardless of complexity. As a result, it is nearly impossible to conceive an automated system without sensors \cite{industrialsensors2023}.  

An industrial robot refers to a robot system that can be used for manufacturing. Industrial robots are automated, programmable and capable of movement on three or more axes. Industrial robots are gradually replacing human to overtake dangerous and delicate tasks. \textcolor{black}{Typical applications where robots can assist operations} include welding, painting, assembly, disassembly pick and place for printed circuit boards, packaging and labeling, palletizing, product inspection, and testing. Moreover, with the help of smart sensors, robots can ensure accomplish tasks with high endurance, speed, and precision. 

Cloud computing \textcolor{black}{bringing computing services}, including servers, storage, databases, softwar, etc. over the Internet, has become a economic and flexible computation resources deployment solution for enterprises in recent years. Most cloud computing services fall into four broad categories: infrastructure as a service (IaaS), platform as a service (PaaS), serverless, and software as a service (SaaS) \cite{cloudcomputing-Azure}. However, because \textcolor{black}{cloud resources are usually far away from the data sources, resulting in a long latency for IIoT data analysis}, edge computing and fog computing that bring computation resources near data sources offer new ways of processing and analyzing data in real time. Currently, cloud-edge-fog architecture has become an effective architecture to provide sufficient computation and storage resources that are required by IIoT intelligence. 



\subsection{Networking layer}
\subsubsection{Motivations}
The revolution of IIoT networks is inspired by global trends\cite{GlobalNetworkingTrends2020}. 
Firstly, because of more and more smart equipment and staff accessing the network in the domain of smart manufacturing, and the connection between systems, human, processes, locations and devices getting more distributed and complex, the economic value of the network for enterprises are growing. 
Secondly, the inherent unpredictability of business that the network quickly corrects the transmission modes to adjust to the new services, processes and models. Thirdly, many manufacturing processes are time sensitive and mission critical, which requires the communication time of network to be more precise and predictable\cite{nasrallah2018ultra}. 

Technology improvements encourage IIoT network revolution as well. The applications and data of enterprises are leaving the premises and moving to the public/private clouds. Part of the monolithic applications in many cases are being modularized into microservices that are delivered via a variety of virtual and physical workloads across the entire enterprises. 
Under the background of applications and microservices revolution, the network is expected to be a growing set of nerve clusters that can be anywhere along the cloud-edge-terminals. 


\textit{CISILION} summarizes four primary objectives for the new network in terms of aligning to the business, abstracting complexity, assuring performance and reducing risks \cite{GlobalNetworkingTrends2020}. The network in IIoT intelligence is expected to achieve some similar characteristics. As presented in Fig.~\ref{fig:informationIsland}, which refers from \cite{WhitePaperZhongxingDeqin} and Prediction from TF security \cite{TFprediciton20200913}, the network connecting human-machine-things helps overcome the serious problem of information island, provides high-speed, wide-coverage and low-latency communication, and supports the development of overlay applications, such as industrial big-data mining and immediate decisions. 
To be specific, in the increasingly demanding manufacturing environment, the network needs to be able to adapt quickly to changing manufacturing requirements. The network should provide convenient access of fast-changing set of equipment, staff, applications and services, so that resource scheduling and management can be focus more on. Providing steady continuous service without interruption is also essential for the network used in the manufacturing. 

\begin{figure*}[h!]
	\centerline{\includegraphics[width=0.7\textwidth]{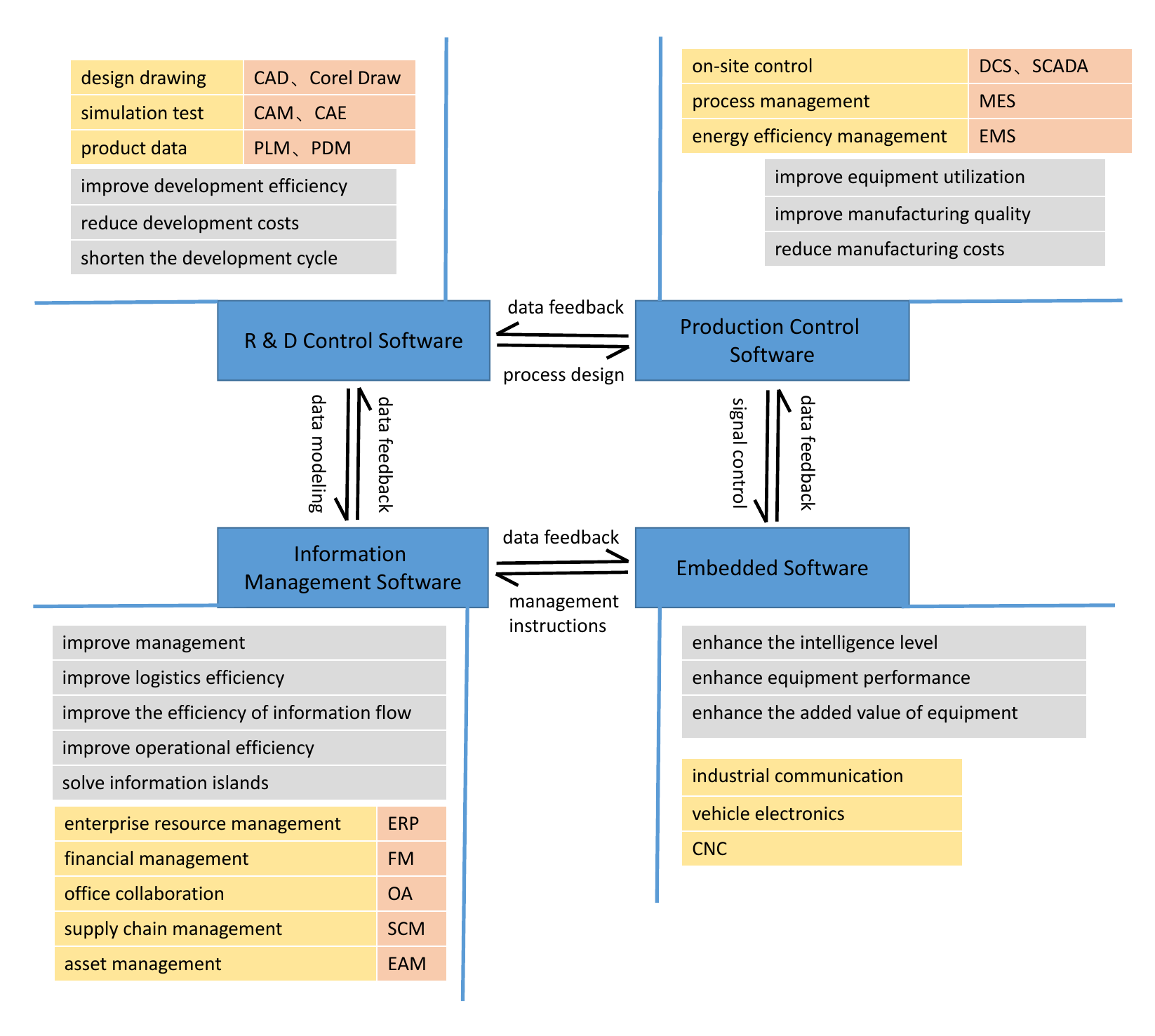}}
	\caption{The primary functions and application instances of R\&D software, production control software, information management software and embedded software. The grey blocks represent the expected effectiveness of the industrial softwares. The yellow blocks show the corresponding functions. The orange blocks include representative instances. } 
	\label{fig:softwarecategories}
\end{figure*}
\subsubsection{Connotation} 
All feasible networking methods that connects human, sensors, robots, cloud/edge/fog computing resources and other smart devices belong to the networking layer of IIoT intelligence. 

Some advanced networks are proposed in recent years. Representative progresses involve software-defined network (SDN), white box switch,  information centric network (ICN), deterministic networks, time sensitive networks, 5G and network slicing, etc. These achievements are driving the network to realize high performance, including rich bandwidth, high-speed and steady transmission, massive connectivity, and so on. Here some techniques are briefly introduced as examples to illustrate the effects of advanced technologies for the future IIoT network. 
SDN exploits the ability to decouple the data plane and the control plane in routers and switches \cite{fang2013loss}, which enables the network to be intelligently and centrally controlled and managed, regardless of the complexity of the underlying network technologies \cite{hu2014SDNsurvey}.  
ICN is a promising candidate for the paradigm of the future internet \cite{xylomenos2013survey}. Inspired by the fact that it is the information dissemination rather than pair-wise communication between end hosts, ICN switches the IP-based Internet to a information based architecture by in-network caching and storage \cite{xylomenos2013survey}\cite{vasilakos2015information}. 
The fifth generation (5G) wireless communication aims to achieve revolutionary leap forward in terms of latency, massive connectivity, network reliability and energy efficiency \cite{shafi20175g}\cite{agiwal2016next}. The network slicing is an unique technique that is firstly used in 5G to achieve the ambition of 5G. Network slicing facilitates multiple logical self-contained networks on top of a common physical infrastructure platform, so that 5G is able to support very diverse and sometimes extreme requirements in an isolated and transparent approach \cite{afolabi2018network}. Some industrial instances that network slicing is used are given in the work \cite{alliance2016description}. 

\subsection{Industrial software layer}
\subsubsection{Motivations}
In an increasingly digital manufacturing, human, smart sensors and robots are generating industrial data continuously, and networks are busy for transferring data. The foundation of IIoT intelligence has been built by equipment layer and networking layer. However, there is a lack of software that can be used to manage data and production process.

\subsubsection{Connotation}
The industrial software is the product of the integration of industrial knowledge and information technologies. The product is updating iteratively as the long-term accumulation and application of industrial innovative knowledge. According to the functions and characteristics, the industrial software can be divided into four categories: R\&D software, production control software, information management software and embedded software\cite{SoftwareWhitePaper2019}. Fig.~\ref{fig:softwarecategories} presents the primary functions and application instances of the four types of industrial softwares.


The industrial software is highly complex for several reasons. First of all, there are various industries, among which process industry and discrete industry contain 39 major categories, 191 medium-granularity categories and 525 fine-granularity categories. Secondly, many professional knowledge and experience in machinery, electronics, optics, acoustics, sound control, fluid heat treatment are required to support research and development of industrial softwares.  Thirdly, the complexity of industrial products is wide-span, from simple products such as clothing and toys to complex products in aviation and satellite. Moreover, the industrial softwares engage in all stages of industrial processes, while the industrial process is complex, including R\&D, production, marketing, operation, maintenance, supply chain management and others. Deployed equipments are also various, functions involve production, implementation, test and measurement. Then, it is difficult for manufacturers to cooperate. For example, Boeing747 is jointly produced by more than 16000 enterprises with different scales from six countries. Finally, industrial software has high requirements for real-time reliability, because  unreliable data is possible to cause huge problems in terms of cost and safety.

In recent years, governments are promoting the rapid development of industrial software by policy incentives, at the same time, innovative technologies also \textcolor{black}{provide opportunities to develop} industrial software. As the industry environment becomes more and more competitive and complex, the industrial software is upgrading in three ways, including systematic platform, cloud-based deployment and lightweight development. 
To be specific, early-stage industrial software are oriented specific scenarios, such as CAD for designing, CAE for simulation, ERP for enterprise management, etc. Current industry requires collaboration among multiple companies. Therefore, some systematic platforms are expected to support the cross-company interaction. The systematic platforms should integrate and utilize resources and tools to provide solutions for smart manufacturing. 
On the other hand, developing cloud-based industrial softwares \cite{wu2017digital}\cite{mora2017design} is an effective approach to deal with current operation models (IaaS, SaaS, PaaS) and customization demands. 
Guaranteeing data interconnection, real-time interaction and remote monitoring \cite{Remotehumanrobotcollaboration2020} is important topic in smart manufacturing, which promotes to develop lightweight mobile industrial APP.

\subsection{Modeling layer}
\subsubsection{Motivation}
In the architecture of IIoT intelligence, the modeling layer should achieve two functions, that is, modeling physical processes into the digital spaces, at the same time, transferring decisions made in the digital space to physical world. The development of modeling methods are driven by future intelligent applications, such as remote control, predictive maintenance, process optimization and so on. Specifically, in order to perform well on these scenarios, it is necessary to model particular objects accurately, and it is better to maintain uninterrupted connections with the objects to update the models in real time and transfer instructions. 

\subsubsection{Connotation}
Modeling methods, \textcolor{black}{which} can build connections between physical world and cyber world for IIoT,  belong to the modeling layer. 

\begin{figure} 
	\centerline{\includegraphics[width=0.4\textwidth]{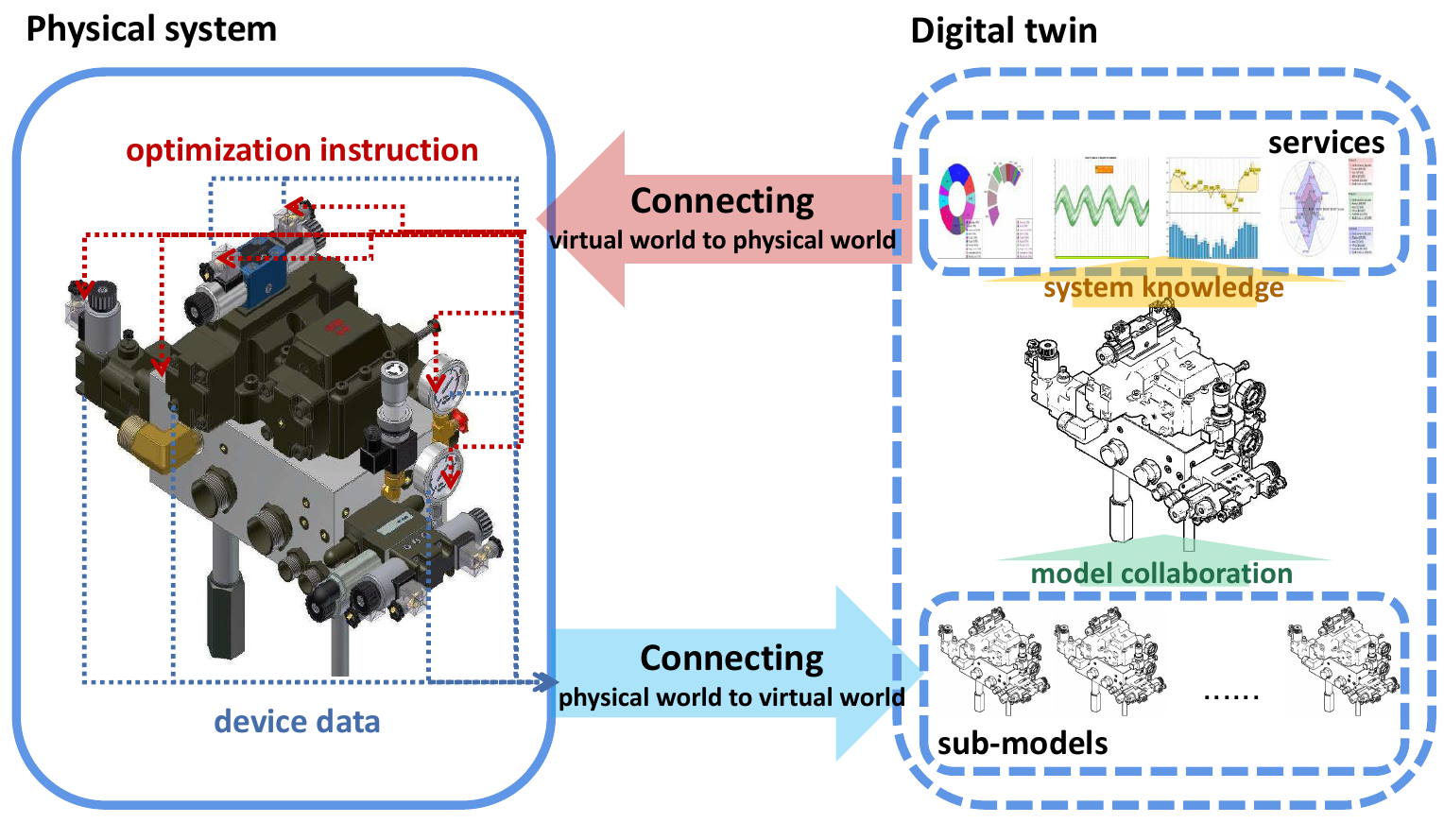}}
	\caption{The representation model of digital twin}
	\label{fig:digitalTwin}
\end{figure}

The digital twin is the most promising technology to realize the interactive between cyber spaces and physical spaces.  
\textcolor{black}{Its representation model} is shown in Fig.~\ref{fig:digitalTwin}. The digital twin has five necessary keypoints: data, connection, model, interaction, and application. Consequently, the digital twin can run through the entire life cycle of physical processes, and provide capabilities with description, analysis, diagnosis, prediction, prescription, cognition, and so on \cite{NationalDevelopmentandReformCommission2020}\cite{DHL2019DHL}.  Currently, the industrial-research-university communities explore the digital twin in two aspects, modeling and simulation. Modeling refers to establishing 3D virtual twins for physical entities. The twins are expected to reproduce appearance, geometric association, motion dynamics, and other attributes of the physical entities. The simulation is an activity to calculate, analyze and predict the future state of physical entities based on the virtual twins with industrial knowledge, mechanism, and data. As far as researches contributions at this stage, the modeling and simulation of digital twin depend on industrial software, including CAD, CAE, etc.

Besides the digital twin, there are some other methodologies that can model certain industrial scenarios, such as Use Case Maps (UCMs), message sequence chart (MSCs), digital shallow, etc. 
The UCMs is a real-time system oriented and scenario-based modeling method \cite{alsumait2002use}\cite{buhr1998use}. The UCMs uses abstract symbols to present the structure and interactive behaviors of complex systems, so that developers can understand the overall system easily. The UCMs provides an intuitive and clear visual system description method, which can realize flexible conversion between formal and informal description of the system.
The MSCs, also known as timing diagrams, message flow diagrams, are effective graphical and textual languages used to describe the system messaging and are widely used in software development practices. 
There are three popular MSC software versions, namely MSC-92, MSC-96 and MSC-2000 respectively. The MSC-92 standard refines the basic concepts in the system and uses formal expressions to describe the basic processes. In the MSC-96 version, structured language is added compared with the previous version. The MSC-2000 version has made improvements in four aspects: control flow, quantitative representation time, data description and conditions.

\subsection{Analysis and optimization layer}
\subsubsection{Motivations}
With the help of modeling layer, the connection between cyber space and physical space is constructed. However, the connection only provide a channel to control physical processes based on the optimization in the cyber space.  It cannot solve specific problems in the industrial environment, such as manufacturing processes planning and schedule, topology optimization, and prognosis, etc. Therefore, we design the analysis and optimization layer to support the development of IIoT intelligence in specific problems and scenarios. 

\subsubsection{Connotation}
intelligent algorithms that can solve specific problems in the industrial environment belong to the analysis and optimization layer. 
There are too many scenarios and problems to enumerate them all. Due to space limitation, this paper only identifies some issues of greater concerns, including manufacturing processes planning and schedule, topology optimization, and prognosis.

\paragraph{Manufacturing process planning and scheduling}
Manufacturing process planning determines how a product will be manufactured by selecting and sequencing manufacturing operations. The process planning aims at improving some specific performance (e.g., shorten processing time, minimize production cost) while satisfying a set of domain constraints \cite{shen2006agent,chang2000integrated,chen2023scheduling}. 
The manufacturing scheduling is the process of assigning manufacturing resources over time to the set of manufacturing processes in the process plan \cite{shen2006agent}. It is an optimization problem that schedules of resource allocation will be proposed with respect to criteria such as delay, throughput, or cost. 
The process planning and scheduling problem is becoming increasingly important to improve the flexible manufacturing. However, the problem is typically NP-hard, that is, it is impossible to get optimal solutions without the use of an essentially enumerative algorithm, while the computing complexity grows significantly as the problem scale increases.  


Computer-aided process planning (CAPP) \cite{niebel1965mechanized}\cite{yusof2014survey} based approaches are useful for the problems, such as petri-net-based approaches and knowledge-based expert system, agent-based methods, and heuristic algorithms.
Petri-net tools can also be used to analyze the performance of the system, generate code, simulate the system and perform model checking on it \cite{thong2015survey,zhang2011petri,quintanilla2016petri,guo2017timed}. 
Knowledge-based system provides necessary expert information, featuring information input, recovering comparable plan and making essential alteration for manufacturing process planning and scheduling \cite{kretschmer2017knowledge,tsai2010knowledge,li2011recent}.
Agent technology \cite{wooldridge1995intelligent} has received significant attention and achieved important progress, both in terms of algorithms \cite{Iglesias1998A,Tveit2001A,Sudeikat2004Evaluation} and platforms \cite{Tonn2010ASGARD,Lin2005Tool,Kravari2015A,Bergenti2017Agent}. Cooperative intelligent agents are also used in developing distributed CAPP systems for process planning and scheduling problems in manufacturing \cite{zhang2007agent,li2010agent,sarkar2018multi,torreno2017cooperative}. 
Heuristic approaches \cite{gendreau2005metaheuristics}\cite{blum2011hybrid}, like climbing, simulated annealing, Tabu search, and GAs, try to replace the exhaustive search strategies with some sophisticated experience. 


\paragraph{Topology optimization} 
Topology optimization is a mathematical method that answers the question of how to place material within a prescribed design domain in order to obtain the best structural performance \cite{sigmund2013topology}. The topology optimization happens at the initial stage of the product manufacturing process \cite{zuo2006manufacturing}. It is efficient because automated optimization processes are employed to generate the conceptual designs instead of the conventional trial-and-error approach \cite{liu2016survey}. Some IIoT intelligence  technologies are used to maximize the performance. Surveys of representative manufacturing-oriented topology optimization approaches \cite{liu2016survey}\cite{sigmund2013topology} summarize the efforts to parameterize the topology (SIMP method \cite{marck2012topology}\cite{long2016optimization}, level set method \cite{yamada2010topology}, spline based methods and some limitations), machining-oriented topology optimization methods (length scale control \cite{zhou2015minimum} and geometric feature based design \cite{liu20153d}), and injection casting oriented topology optimization (thickness control \cite{liu2018uniform}). 


\paragraph{Industrial prognostics} 
The researches about industrial prognostics mainly focus on estimating and anticipating events of interest regarding industrial assets and production processes \cite{diez2019data}. Particularly, the industrial prognostics are regraded as a data-based workflow in the work \cite{diez2019data}, and two important research topics are mentioned. $(a)$ The descriptive prognostics that mainly summarize data and infer insights from the data. Consequently, the gained information from data can help detecting events of interest or to estimate the health status of the industrial asset, product or process under study. Important research topics are pattern recognition and classification \cite{kuo2016study}\cite{chang2012development}, and health management \cite{vogl2014standards}\cite{shin2018framework}.  $(b)$ The predictive prognostics utilize a variety of data fusion, statistics, modeling, and machine learning techniques to study recent and historical data, to learn prognostic models, which make accurate predictions about the future status of the monitored asset.  The condition-based maintenance \cite{kumar2018big}\cite{rastegari2014implementation} and the predictive maintenance \cite{wang2017new}\cite{he2017integrated} are two of the interesting topics. 

Some prognostics approaches are summarized in the survey \cite{xia2018recent}, such as $(a)$ physics-based approaches \cite{luo2008modelprognostic} achieve prognostics by using mathematical models to describe the degradation mechanics or damage propagation of a machinery component. $(b)$ Data-driven approaches \cite{liu2015novelhiddensemiMarkov}\cite{soualhi2016hiddenMarkov} use the collected data (usually condition monitoring data) to map the characteristics of damage/degradation state to achieve prognostics. $(c)$ Hybrid approaches \cite{qian2017multiScaleApproach} combine the advantages of different approaches to improve the prediction accuracy or extend prognostics model applications. 


\begin{table*}
	\renewcommand\arraystretch{1.2} 
	\caption{Digitization transformation of 4 lighthouse factories with IIoT intelligence }
	\label{table:lighthouseInstance}
	\begin{center}
		\begin{tabular}{m{2cm}|m{2.5cm}|m{3.2cm}|m{3.2cm}|m{3.5cm}}
    \hline
    \hline
			& AGCO/Fendt, Marktoberdorf, Germany & Henkel Laundry and Home Care, Germany & Johnson \& Johnson Vision Care, Jacksonville, Florida & Weichai Power Company, Weifang, China \\
			\hline
   Analysis and Optimization Layer
			& 
   \begin{itemize}[itemsep=0pt,parsep=0pt,leftmargin=*]
        \item Advanced big-data analysis and quality monitoring 
    \end{itemize}
			& 
   \begin{itemize}[itemsep=0pt,parsep=0pt,leftmargin=*]
	\item Integrating data evaluation and optimization tools 
       \item Standardization of measurements and calculation of KPIs
    \end{itemize} 
			& 
   \begin{itemize}[itemsep=0pt,parsep=0pt,leftmargin=*]
        \item Adaptive process control
        \item  Advanced analysis algorithms to connect manufacturer, patient, professional opticians and retailer
   \end{itemize}
            & 
    \begin{itemize}[itemsep=0pt,parsep=0pt,leftmargin=*]
        \item Big-data analysis, remote maintenance
    \end{itemize} \\
			\hline
	Modeling Layer & 
    \begin{itemize}[itemsep=0pt,parsep=0pt,leftmargin=*]
        \item Digital twin 
    \end{itemize}
			&  
    \begin{itemize}[itemsep=0pt,parsep=0pt,leftmargin=*]
        \item Digital shadows of 33 sites and 10 distribution centres 
    \end{itemize}
			&
    \begin{itemize}[itemsep=0pt,parsep=0pt,leftmargin=*]
        \item Augmented-reality tools for testing product comfort
    \end{itemize} 
			&  
   \begin{itemize}[itemsep=0pt,parsep=0pt,leftmargin=*]
        \item Digital rapid modelling design
        \item Virtual development simulation 
        \item IIoT-based testing
    \end{itemize} \\
			\hline
    Software Layer & 
    \begin{itemize}[itemsep=0pt,parsep=0pt,leftmargin=*]
        \item Holistic manufacturing execution system (MES) 
    \end{itemize} 
			& 
   \begin{itemize}[itemsep=0pt,parsep=0pt,leftmargin=*]
        \item Laundry data foundation
        \item Henkel laundry digital backbone
    \end{itemize} 
			& 
    \begin{itemize}[itemsep=0pt,parsep=0pt,leftmargin=*]
        \item An adaptive flexible modular platform (FMP) for rapid new product introduction
        \item An interconnected visual-value-chain control tower
        \item A mobile/web platform 
    \end{itemize} 
			& 
   \begin{itemize}[itemsep=0pt,parsep=0pt,leftmargin=*]
        \item An end-to-end product development system
        \item An IoV system to collect real-time data on engine speed, fuel consumption and power under various working conditions
        \item An app-based car networking platform provides users with efficient service solutions
    \end{itemize} \\
			\hline
			Networking Layer & 
    \begin{itemize}[itemsep=0pt,parsep=0pt,leftmargin=*]
        \item Iot realizing a consistent data platform across all systems
    \end{itemize} 
    &  
    \begin{itemize}[itemsep=0pt,parsep=0pt,leftmargin=*]
        \item A digital backbone connecting factories to a central data repository in order to collect real-time energy consumption data
    \end{itemize}
    & 
    \begin{itemize}[itemsep=0pt,parsep=0pt,leftmargin=*]
        \item  IIoT-enabled and fully validated closed loop 
    \end{itemize} 
    & 
    \begin{itemize}[itemsep=0pt,parsep=0pt,leftmargin=*]
        \item  Intelligent IIoT-based testing 
    \end{itemize}\\
			\hline
    Equipment Layer & 
    \begin{itemize}[itemsep=0pt,parsep=0pt,leftmargin=*]
        \item  Manufacturing-related robots and sensors 
    \end{itemize}
			& 
   \begin{itemize}[itemsep=0pt,parsep=0pt,leftmargin=*]
        \item  3500 interconnected sensors are used to measure electricity, fossil fuel, water, compressed air, steam consumption
    \end{itemize} 
			& 
   \begin{itemize}[itemsep=0pt,parsep=0pt,leftmargin=*]
        \item Advanced robotics and intelligent material handling
    \end{itemize} 
			& 
   \begin{itemize}[itemsep=0pt,parsep=0pt,leftmargin=*]
        \item Manufacturing robots and increased sensors are used to ensure real-time acquisition and uploading of test results
    \end{itemize}\\
			\hline
			Returns & 
   \begin{itemize}[itemsep=0pt,parsep=0pt,leftmargin=*]
        \item 60\% lower cycle time
        \item 37\% assembly line volume increase
        \item 30\% reduction in time-to-identify quality issues
    \end{itemize}
        & 
   \begin{itemize}[itemsep=2pt,parsep=0pt,leftmargin=*]
			    \item 36\% lower sustain ability footprint
                    \item 10\% reduction in processing costs
                    \item 25\% reduction in logistic costs
                    \item Reducing Henkel’s energy consumption by 38\% and its water consumption by 28\%
			\end{itemize} 
        &
    \begin{itemize}[itemsep=0pt,parsep=0pt,leftmargin=*]
        \item 30\% faster time to market on new products
        \item Maximizing yield, reducing downtimes
        \item 11\% OEE improvement
        \item Enabling a 50\% reduction in the time to scale worldwide volume of new products
    \end{itemize}
	& 
    \begin{itemize}[itemsep=0pt,parsep=0pt,leftmargin=*]
        \item New product development is shortened from 24 months to 18 months
        \item The product design reuse rate has increased by 30\% 
        \item Reducing test costs by more than 20\%
        \item Reducing labour costs by 75\%, shortening the R\&D cycle by more than 20\% 
    \end{itemize} \\
    
    \hline 
    \hline
		\end{tabular}
	\end{center}
\end{table*}

\section{A close look on lighthouse factories with IIoT intelligence  } \label{section:closelook}
In this section, a close look on lighthouses and collaborations across companies is presented to demonstrate that IIoT intelligence is able to empower intelligence transformation and upgrading to smart manufacturing, and augment the confidence to explore and exploit IIoT intelligence.

Since 2016, the World Economic Forum, in collaboration with McKinsey \& Company, has been monitoring progress in advanced manufacturing worldwide. In 2017, the World Economic Forum identified 40 initial advanced manufacturing use cases that began to pilot intelligent technologies with some success. In 2018, 16 companies were recognized as leaders that achieve step-change results, both operational and financial, across individual sites \cite{IndustrialRevolution4}. In 2019, 28 additional facilities were identified as end-to-end (E2E) lighthouses \cite{globalLighthouse}. \textcolor{black}{Until 2023, the number of E2E lighthouses increases to 54, and 17 are also sustainability lighthouses \cite{globalLighthouse2023}. }

These advanced companies succeed in transforming factories, value chains and business models to generate compelling financial, operational and environment returns \cite{globalLighthouse}. The specific key performance indicator (KPI) improvements are reflected in factory output increase, productivity increase, overall equipment effectiveness (OEE) increase, quality cost reduction, waste reduction, energy efficiency, change-over shortening, etc. These impressive progresses depend on expanding IIoT intelligence within the manufacturing plant environment, for example, automation, digitization, big-data analysis, remote control, 3D printing, etc.

According to the inside perspectives from the global lighthouse network, this paper closely looks the contributions of IIoT intelligence  for smart manufacturing. This paper selects some lighthouse companies, and summarize the improvements from the perspective of hierarchical development of IIoT intelligence. The upgrade behaviours and corresponding returns are presented in Table~\ref{table:lighthouseInstance}.

Besides the successful end-to-end digital transformation lighthouse factories, \textcolor{black}{more and more collaborations are being gradually released}. For example,  
in 2019, Haier, China Mobile,  Huawei and Mstar released the world's first 5G smart factory joint solution \cite{QingdaoHaier5GSmartFactory}. The factory implements intelligence in terms of quality detection, maintenance, machine collaboration, material transportation, enery management and campus security under help of IIoT intelligence. 
In 2021, Midea Group, China Unicom, and Huawei rolled out Guangdong's first 5G fully-connected smart manufacturing demonstration site at the factory of Midea's Kitchen Appliance Division in Shunde district, Foshan city \cite{MideaChinaUnicomandHuawei}. 
\textcolor{black}{
In 2022, Siemens and GENERA announced a collaboration \cite{SiemensAndGENERA2022}, which involves the integration of Siemens technology in GENERA digital light processing systems, including operational technology, information technology, and automation. In addition, GENERA digital light processing solutions will be fully integrated into Siemens simulation and planning tools for factory design.
In 2023, Siemens and Intel released a collaboration \cite{SiemensAndIntel2023} to advance semiconductor manufacturing production efficiency and sustainability. Leveraging their respective portfolios of cutting-edge IoT solutions, along with Siemens automation solutions, the collaboration aimed to enhance semiconductor manufacturing efficiency and sustainability across scopes 1-3 of the value chain.
}


\section{Technology studies on IIoT intelligence } \label{section:Technology-Study}
There is no doubt that the rapid and effective development of IIoT intelligence cannot leave some crucial technologies.
Because of the space limit, we identify some of the most influential technologies for IIoT intelligence, and clarify their contributions in the development of IIoT intelligence. The application progresses of these technologies are described as follows. 

\subsection{Industrial robots as intelligent labors}
The industrial robot usually refers to the manufacturing-oriented multi-joint mechanical arms, or other mechanical devices with multiple axes. The primary characteristics involves automatically control, programmable, multi-purpose, fixed in place or mobile, etc. Based on the mechanical structure, the robots can be classified into six types, including articulated robot, SCARA robot,  cartesian robot, Delta robot, cylindrical robot and others \cite{IFRchapter1}\cite{Evolutionindustrialrobots2013}. 
The industrial robots are mainly used to replace human in automatic and/or flexible production to some extent \cite{Germanrobots2017}\cite{Evolutionindustrialrobots2013}. According to the statistics of \textit{operational sotck of industrial robots by customer industry and application - world} made by IFR(the International Federation of Robotics)\cite{IFRPresentationPPT},  the largest application of industries of industrial robots is automotive manufacturing, following that electronics is important as well. As a consequence, the automatic production and flexible manufacturing of these two areas are developing rapidly \cite{WhitePaperofFoxconn}\cite{IndustrialRevolution4}. 
In terms of usage, most industrial robots are used for handling, followed by welding and assembling. These tasks usually require lots of physical energy and part of content has bad impacts on health.  At the same time, the statistics about annual installations also demonstrate that industrial robots have played a significant positive role in promoting industrial processes, and there is still large room for further development.


For industrial robots at current stage, a relatively complete industrial chain has been constructed and each stage has leading enterprises. Except the basic materials, like metal, non-metal and basic components, the techniques of making three primary components are handled in some companies, such as, Siemens, Yaskawa, Eston, Nabotsk, Hammer, KEBA, ABB, etc.  After getting enough materials and primary components, some companies focus on integrating robots. Related leading enterprises include Fanuc, Yaskawa, Nizhi, Eston, and so on. In addition, some well-known companies, such as Siemens, Fanuc, ABB, etc., also overtake business of system integration, which refers to deploy production lines  using the robots and some software. 

\textcolor{black}{In practical manufacturing processes, robotics has played a crucial role in improving production efficiency. For example, in the World Economic Forum's global lighthouse network 2022 \cite{globalLighthouse2022}, \textit{Haier} refrigerator factory in Qingdao, China, uses collaborative robots for flexible manufacturing, resulting in a 52\% improvement in assembly efficiency. Similarly, \textit{Sany Heavy Industry} in Changsha, China, employs robotics in logistics execution, leading to an 11\% increase in the on-time delivery rate.}

\subsection{Machine vision system as eyes of industries}
\begin{table*}
    \renewcommand\arraystretch{2} 
    \caption{Comparison between human eyes and machine vision}
    \label{table:machinevisioncompare}
    \begin{center}
    \begin{tabular}{m{4cm}<{\centering}|m{6cm}<{\centering}|m{6cm}<{\centering}}
    \hline
    \hline
     & Human Eyes & Machine Vision \\
    \hline
    Adaptability & Strong & Relatively strong, less demanding on the environment \\
    \hline
    Speed &  The visual persistence of 0.1 second makes it impossible for human eyes to see faster moving targets & The shutter time can reach about 10 microseconds \\
    \hline
    Precision &	Low, unable to quantify &	High, micron accuracy, easily quantified \\
    \hline
    Grayscale resolution & Low, only 64 gray level resolution & High, currently 256 gray levels are generally used, and the acquisition system can getgray levels such as 10bit, 12bit, 16bit, etc. \\
    \hline
    Repeatability &	Low, easily affected by emotions, physical conditions, etc. &	Always uniform methods and standards \\
    \hline
    Data collection and the possibility of future intelligence applications & No, the human eye does not have the ability to analyze after collecting data & Strong, with strong data collection and analysis functions, it is a huge data entry at the edge layer \\
    \hline
    \hline
    \end{tabular}
\end{center}
\end{table*}


The goal of a machine vision system \cite{jain1995machine,davies2004machine,liu2022deep} is to create simplified models of the real world based on sensed images. These models should be understandable to computers and can restore real information (geometric shape, color, texture, category, printed matter, etc.) from images. Compared with human eyes, machine vision  has significant differences and benefits, as presented in Table~\ref{table:machinevisioncompare}. Thanks to these advantages, machine vision systems have been applied in industry. 



Machine vision systems have great differences between various industrial tasks. For general practical applications, there are some shared common performance requirements in terms of precision, speed, environmental robustness, and easy to use~\cite{hashimoto2017current}.  
Without accurate measurement, the machine vision cannot successfully guide the robotic hand to perform automatic assembly of smartphone or mechanical watch ~\cite{qi2010vision,chang2018robotic,song2020robotic}.  
In the high-speed assembly, the average time of template matching is less than several microseconds for electronics manufacturing~\cite{zhang2015region,zhong2017blob}. 
The manufacturing techniques, including machining (milling, grinding, polishing), welding, injection molding, etc., can be improved with an auxiliary machine vision system.
Rrobotic grinding demonstrates flexibility, intelligence and cost efficiency under the guidance of machine vision~\cite{tam1999robotic}\cite{zhu2020robotic}. 
Equipped with machine vision system, the performance of intelligence robotic welding improves a lot \cite{wang2020intelligent}.
Robotic seam tracking is challenging due to the strong noise images(reflection, strong arc light and the large spatter), and a convolution neural network is proposed to extract the feature points of weld seam image~\cite{du2019strong}.
A machine vision approach is introduced to tracking the welding electrode tips for assessing welding arc stability and polarity in manual metal arc~\cite{Jamrozik2021assessing}.
\textcolor{black}{In general, the on-line visual feedback of machine vision measurement, including position, shape, velocity, etc., enables the machining, welding, injection molding techniques with improved manufacturing quality and speed. For example, \textit{Agilent}'s machine vision toolkit has enabled speed and scale through democratized access to advanced vision capabilities \cite{globalLighthouse2023}. \textit{Unilever} complemented machine vision supervision platform for people safety and food safety compliance, decreasing 78\% unsafe behaviours \cite{globalLighthouse2022}. }

\subsection{Networking facilitates ubiquitous connectivity}

Current network is the product of the deep integration of the new generation of information,  communication and modern industrial technologies. It has been one of the magic weapons to realize digital transformation of manufacturing.  This paper briefly introduces some related progresses of typical network technologies and applications, such as 5G, time sensitive networking, and identity resolution. 

The 5G network can connect autonomous industrial machines, ground vehicles, robots and software in an URLLC wireless communication, aiming at assistance of real-time control and management of industrial machinery towards smart factories \cite{shafi20175g}\cite{agiwal2016next}\cite{9522071}.
The authors in \cite{9522071} study 5G-enabled industrial Internet of Things, considering using machine learning to handle complex tasks at industrial machinery and 5G networks management,configuration,and control. 
An intelligent and green resource allocation mechanism for the IIoT under 5G heterogeneous networks is proposed \cite{9272839}, so as to ensure the high reliability and energy efficiency of IIoT services.
Networking slicing is an essential technique to support massive connectivity and service isolation for 5G, so as to realize scale-up applications in the industry \cite{afolabi2018network}\cite{alliance2016description}. \textcolor{black}{In practice, \textit{Exquisite Automotive Systems} has joined forces with China Unicom Hebei and Huawei to create the world's first flexible 5.5G production line in their Baoding factory, where the service interruption time per year reduces from 60 hours to 5.26 mins, and the annual output value increases \$112 million \cite{huawei5Gcase}.} 

Time-Sensitive Networking (TSN) \cite{8695835} is a standardized technology that provides deterministic information transmission on traditional Ethernet. 
In \cite{8610105}, the authors introduce Open Platform Communication Unified Architecture TSN (OPC UA TSN) as a new technology and present the current view.
In \cite{9142734}, the authors present an approach to integrating heterogeneous Industrial Ethernet applications over TSN,
and deploy different real-time schedulers that are embedded in a Software-defined Network (SDN)\cite{fang2013loss}\cite{hu2014SDNsurvey} controller to allow different applications with various traffic patterns and Quality-of-Service (QoS) requirements to share the same network.
In \cite{8715451}, the authors introduce the latest research progress of industrial communication systems, and analyze the technical challenges and future trends for TSN. 
\textcolor{black}{In practice, \textit{Moxa}, which is a leading provider of industrial networking, computing, and automation solutions, has successfully integrated TSN into the wafer manufacturing process of  ELS System Technology, a leading Taiwanese lithography solution provider \cite{moxa2023}. }

\subsection{Digital twin connects cyber-physical spaces}
The digital twin attracts much attention in recent years, \textcolor{black}{and some great progresses have achieved}. This paper reviews these contributions in terms of accurately modeling and interoperability. 

In terms of accurately modeling in manufacturing, current works mainly focus on single-dimension modeling, including structure modeling, attribute modeling, behavior modeling, and process modeling. Primary methods include data-driven modeling, knowledge-based modeling, and hybrid modeling. 
A neural network based modeling approach is proposed \cite{yu2021digital} for the uncertainty and complexity of the digital twin system. 
The digital twin technique is used into monitoring task of the tool Wear Monitoring of Micro-Milling \cite{kiswanto2020digital}. The model is based on the existing electrical and mechanical formulas of the three core modules of the micro-milling machine tool.
The work \cite{ladj2021knowledge} presents a digital twin model that makes use of data collected by sensors, so that the performance (accuracy and reliability)  can be improved by the knowledge and data analysis. 
A digital-twin-assisted fault diagnosis using deep transfer learning to analyze the operational conditions of machining tools is proposed\cite{deebak2021digital}. This proposed system has developed an intelligent tool-holder that integrates a k-type thermocouple and cloud data acquisition system over the WiFi module, and the tool-holder can provide better accuracy to demonstrate the performance improvement of milling and drilling operations of cutting tools.

Interoperability supports mutual execution between physical and digital spaces. It is an increasingly attractive topic. However, related researches are still at an early stage, and many problems remain about mutual intelligibility, interoperability, dynamic updating, data heterogeneity, submodel heterogeneity, etc.  Currently, some works are achieved that focus more on solving heterogeneous data related problems. The work \cite{liu2021multi} studies the multi-scale evolution mechanism of product quality for the fusion and use of multi-source heterogeneous data. 
Jiang \textit{et al.}\cite{jiang2021model} designed a data interaction mechanism that integrates message middleware, memory database and relational database for data interaction in virtual and physical space. The messaging middleware, such as Kafka and MQTT, is used to send production instructions and feedback execution results. The memory database, eg Redis, is used to store real-time operating data during production. The relational databases, eg MySQL, Oracle, SQL Server, and so on, are used for data management for simulation systems.   

\textcolor{black}{Although there are some challenges in the development of digital twin, its practical values cannot be ignored. According to the global lighthouse network 2023 \cite{globalLighthouse2023}, the digital twin has been used in many practical factories. For example, \textit{ACG Capsules} in India has introduced digital twin to power production planning and scheduling, and improved 13\% on-time delivery in full. \textit{Aramco} in Saudi Arabia has developed a digital twin model for energy consumption reduction, and realized 159\% profitability enhancement. }

\subsection{Deep learning boosts intelligence}
Deep learning methods use computational models that are composed of multiple processing layers to learn representations of data with multiple levels of abstraction \cite{lecun2015deep}. It does not depend on prior data processing and automatically extracts features by using the backpropagation algorithm \cite{rusk2016deep}. \textcolor{black}{In recent years, deep learning has brought great breakthroughs \cite{hu2020reinforcement,hu2021bidirectional,zhang2022truncated,sun2023learning,chen2023global} in the domains of scheduling and planning, computer vision, games, etc., and are speeding up the development of IIoT intelligence. } 
\textcolor{black}{Deep learning methods can be divided into three categories, supervised learning, unsupervised learning and deep reinforcement learning. 
This paper demonstrates the enabling effect of deep learning by researches in industrial machine vision as examples.}

\subsubsection{Supervised learning}
Given a certain amount of annotated images, the learned machine vision models are flexible with various types of output, including discrete labels, continuous values for every single pixel or the whole image. The supervised learning provides efficient solutions for parts of the field problems.
As an important step for planar measurement, the basic edge detection is achieved with Convolutional Neural Network (CNN) by classifying the pixel~\cite{wang2016edge}\cite{he2020bdcn}. The quality of the detected edges are significantly improved in comparison with the traditional edge detectors, e.g., Canny.
Compared to traditional fringe projection profilometry with several images, deep neural networks can map a single fringe image to its height/depth image~\cite{van2019deep}\cite{wang2021single}. It thus enables accurate 3D surface measurement for moving objects. 
Measuring the deformation field between two images also benefits the manufacturing processing, both digital image correlation (DIC) and particle image velocimetry (PIV) are modified with a modern deep regression neural network~\cite{lee2017piv,lee2021diffeomorphic,boukhtache2021deep}. The neural networks demonstrate improved accuracy and robustness.
These geometrical measurement techniques are the foundation for manufacturing, and has been significantly improved with the supervised learning. 


\subsubsection{Unsupervised learning}
One shortage of supervised learning is the need of annotation, which is often expensive. On the contrary, the raw data without labels is quite cheap for most of the industry scenes. As a result, the unsupervised learning based machine vision becomes popular in industry. 
As a representative of intelligent industrial tasks, defect detection is reviewed with recent progress in this part. One strategy of defect detection is to reconstruct the background image (defect-free), thus the difference between the images and corresponding reconstructed reference implies the defect. 
A denoise autoencoder networks is used to rebuild the defect free image, thus the reconstruction residuals of the training patches are used as the indicator for defect prediction~\cite{mei2018unsupervised}. Besides, a variational autoencoder (VAE) is also an optional to rebuild the reference images~\cite{wei2021real}.  Similarly, an modified deep convolutional generative adversarial network (DCGAN) is used to reconstruct a given query image such that no defects but only normal textures will be preserved in the reconstruction~\cite{hu2020unsupervised}. 
The two-stage frameworks build a generative one-class classifier on learned representations which is learned in a self-supervised manner~\cite{sohn2020learning}\cite{li2021cutpaste}. 
Generally speaking, the supervised methods perform significantly better than the unsupervised methods if sufficient annotation is available~\cite{lehr2021supervised}. The hybrid two-stage framework takes advantage of both unsupervised representation learning and supervised classification.

\subsubsection{Deep reinforcement learning}
Intelligent equipment needs the feedback provided by machine vision to automatically make decisions in complex tasks, including robotic assembly, intelligent welding, etc. 
The challenge of traditional machine vision is to find proper representation which can accurately describe the current state of environment and equipment, such as the robotic assembly with deformable objects~\cite{zheng1991strategies}. 
The traditional model-based methods depict the behavior of deformable objects with simplified application-dependent models~\cite{jimenez2012survey}.
Recently, reinforcement learning from high-dimensional sensory inputs is becoming feasible and popular~\cite{hafner2019dream}\cite{Chen2021MuMMI}. The general deformable objects manipulation might be achieved with high-dimensional reinforcement learning~\cite{lin2020softgym}. However, the performance of image-based reinforcement learning (Dreamer~\cite{hafner2019dream}, PlaNet~\cite{hafner2019learning}, etc) is far below the optimal performance of the state representation~\cite{lin2020softgym}. It points out that learning the dynamics of deformable objects is difficult due to the complex visual observations and dynamics.
Despite the challenges of real world~\cite{dulac2019challenges}, the industrial applications of deep reinforcement learning are broaden with the increasing reported cases~\cite{kostrikov2020image}\cite{belousov2021architectural}.

\textcolor{black}{Currently, the practical intelligent transformation cannot leave deep learning. In Global Lighthouse Network 2023 \cite{globalLighthouse2023}, deep learning has been applied in every process and realize impacts everywhere. For example, \textit{Ingrasys} deployed an AI-demand forecasting model, and the model has delivered 27\% more accuracy; \textit{Hengtong Alpha Optic-Electric} automatically optimizes perform and drawing parameters with a model trained on past strategies.}  

\subsection{Smart hardware builds intelligence foundation}
In recent years, smart hardware has realized great breakthroughs in smart sensors and powerful computing devices. 

\textcolor{black}{
The latest CMOS sensors, with approximately 95\% quantum efficiency, have replaced CCDs due to their commercial realization and satisfactory performance efficiency in sensor technology \cite{coffey2018machine}. 
Event-based cameras, known as silicon retinas, offer advantages such as low latency, high dynamic range, and low power consumption~\cite{kim2008simultaneous}\cite{taunyazov20event}. 
For low-light conditions, the quanta image sensor (QIS) serves as a valuable single-photon detector in unstructured industrial environments~\cite{chi2020dynamic}\cite{gnanasambandam2020hdr}.
Besides the optical sensors, computational imaging~\cite{mait2018computational} is another significant advancement in industrial image process.  It optimizes the processing capabilities of optics and electronics through joint design and optimization. Techniques include phase imaging, CT, MRI, ToF imaging, light-field imaging, time-gated imaging, synthetic aperture radar, coded-aperture imaging, and super-resolution.These sensors find wide application across industries \cite{van2020nondestructive,atefi2020robotic,zhang20203d,iacono2021computational}.
} 

The computing capability also keeps rapidly increasing due to Moore's Law and other computing technologies (e.g. domain specific architectures)~\cite{shalf2020future}\cite{leiserson2020there}. Therefore, the long data processing time required by computational imaging can be significantly shorter~\cite{qin2017gpu}\cite{yu2019gpu}.
Besides, the platforms equipped with Intel CPU, Nvidia GPU, or Google TPU enable fast training of deep learning models~\cite{you2018imagenet}\cite{you2019large}. 

\subsection{Cloud/Edge computing inspires business and deployment}


There are many surveys discussing the technical models of cloud computing \cite{kumar2019comprehensive,dang2019survey,arunarani2019task,katal2023energy} and edge computing \cite{luo2021resource,wang2020convergence,liu2019survey}. This paper will not repeat it, but focus on the contributions of cloud computing and edge computing in IIoT intelligence, including inspiring novel manufacturing business and deployment.
\subsubsection{Cloud manufacturing}
Cloud manufacturing \cite{xu2012cloud,liu2017workload,bello2021cloud} is a paradigm that leverages cloud computing technologies to enhance and optimize various aspects of the manufacturing process. It represents a shift from traditional, localized manufacturing models to a more connected and dynamic approach. In cloud manufacturing \cite{helo2021cloud}, resources, information, and capabilities are provided as services over the internet, enabling a more flexible, scalable, and collaborative manufacturing environment.

Cloud manufacturing as a novel business mode can offer many advantages. 
Firstly, manufacturing resources, such as equipment, software tools, and computational power, are shared and accessed through the cloud \cite{ghomi2019cloud,adamson2017cloud}. Advanced manufacturing resource virtualization and sharing strategies play a crucial role in optimizing production processes, improving efficiency, and making informed decisions \cite{liu2014multi,bouzary2020classification,zhang2017research}. This allows for more efficient utilization of resources and reduces the need for individual companies to invest heavily in specialized infrastructure. 
Secondly, manufacturers can access manufacturing services on a pay-as-you-go basis. This on-demand model allows companies to scale their production up or down based on market demand, without the need for significant capital investment.
Thirdly, the cloud enables the collection and analysis of large volumes of data generated throughout the manufacturing process. Data analysis tools can be applied to gain insights into production efficiency, quality control, and overall performance. 
\subsubsection{Cloud-based deployment}
Cloud-based deployment refers to the practice of hosting and delivering software applications, services, and resources over the internet through cloud computing infrastructure. In a cloud-based deployment model \cite{wang2021framework,okwudire2020three}, managers and clients of enterprises access and interact with applications and data remotely, typically through web browsers or client applications, rather than relying on on-premises hardware and infrastructure.

Cloud-based deployment offers a wide range of benefits across various industries. 
Firstly, cloud computing eliminates the need for significant upfront investments in physical hardware and infrastructure. Enterprises can leverage a pay-as-you-go model, paying only for the computing resources they use. This results in cost savings, especially for small and medium-sized businesses that may not have the capital for extensive IT infrastructure. 
Secondly, cloud-based deployment enable remote monitoring and management of manufacturing processes. This is particularly relevant in smart manufacturing, where sensors and IoT devices generate real-time data. Remote access allows stakeholders to monitor production, identify issues, and make informed decisions from anywhere with an internet connection.
Thirdly, cloud computing provides an environment for manufacturing enterprises to experiment and adopt innovative technologies. This includes integrating artificial intelligence, machine learning, and other emerging technologies into smart manufacturing processes. Cloud-based platforms offer the computational power needed for training and deploying advanced models.

\subsubsection{Edge-based deployment}
Unlike cloud-based deployment where industrial data is sent to centralized servers for processing, edge-based deployment shifts some of the processing tasks to local devices or edge servers. This approach is particularly beneficial in industrial settings where real-time processing, low latency, and reliable connectivity are crucial for operational efficiency and decision-making \cite{pundir2020designing,canete2020energy,canete2022supporting}.

Edge-based deployment can obtain advantages in low latency, bandwidth optimization, autonomous operation. Firstly, edge-based deployment processes data closer to the source reduces latency, enabling real-time or near-real-time responses. Secondly, edge-based deployment reduces the need to transmit large volumes of raw data to the cloud, optimizing bandwidth usage. This is especially important in scenarios where network bandwidth is limited or expensive.
Thirdly, edge-based deployment can operate autonomously even without continuous connectivity to the cloud. This ensures that critical functions can continue in the event of network disruptions, enhancing system reliability.

\textcolor{black}{In practice, multiple cloud companies are cooperating with manufacturing to realize business novelty and deployment optimization. For example, by utilizing Microsoft Azure high-performance computing and other Microsoft technologies,  STMicroelectronics is advancing its research and development, transforming its supply chain, scaling manufacturing, and giving its employees the right tools to secure its position as a world leader in semiconductor solutions \cite{zaureSTM}. 
Phillips chose to modernize its legacy on-premises Microsoft Dynamics solution by upgrading to the latest cloud capability of Microsoft Dynamics 365 and Dynamics 365 Field Service. The upgrade provides the company with real-time insights into the performance of the business, which has helped Phillips make process improvements and helped increase operational efficiency and accuracy of field resources \cite{zaurePhillips}. }

\section{Open Challenges and Future Trends} \label{section:challenges}
To identify existing challenges and circumvent potential misleading directions of IIoT intelligence, we think about questions:
\begin{itemize}
    \item What kinds of capabilities are required by future manufacturing? 
    \item Which capabilities have been established?
    \item How to establish or improve the capabilities that are required by future manufacturing?
\end{itemize}

After careful argumentation, We obtain keywords \textit{digital control, deterministic response, cost-friendly operation and deployment, IIoT intelligence proliferation} to answer the first question. The current IIoT intelligence has achieved certain capabilities, however, there is still room for improving them. 

\subsection{Digital control}
Digital control refers to the use of digital technologies and advanced control systems to monitor, manage and optimize various aspects of the manufacturing processes. This approach leverages digital information, connectivity, and data analysis to enhance control and decision-making in real-time. 

Current IIoT intelligence can support digital control to a certain extent. In the current architecture of IIoT intelligence, industrial sensors are integrated to provide real-time information about the production processes, and the data collected by sensors are processed for analysis. The automation systems in software layer, which are powered by programmable logic controllers (PLCs) and distributed control systems (DCS), allow for precise control over manufacturing equipment and processes based on the real-time data analysis. Remote control and monitoring also facilitate digital control by adjust manufacturing processes from remote locations. 

However, the current level of digital control is still far from achieving its ideal vision of providing real-time visibility into all aspects of the manufacturing process, and any remote or local commands can be quickly executed in visual virtual spaces and physical manufacturing processes. 
The real-time visibility indicates that future manufacturing has the ability to monitor the status of equipment, track the flow of materials and analyze production data as it is generated. 
The immediate response for commands indicates that the effective interaction across multiple layers of IIoT intelligence is realized, as well as the cooperation among robots, sensors, computation resources and network resources. 
More importantly, once the ideal vision of digital control is realized, managers can grasp various aspects of the manufacturing environment through real-time monitoring, adjustment, optimization, and control. 

\textcolor{black}{Although achieving digital control poses challenges, the digital twins connecting cyberspace and physical space offer opportunities once three key difficulties are addressed:
\begin{itemize}
    \item Overcoming issues between software and models, as well as standardization challenges in communication protocols within industrial environments, is crucial.
    \item Ensuring high-quality industrial data is essential for accurate analyses and reliable conclusions. Generating such data from practical datasets is vital for enabling intelligent data analysis.
    \item Building digital twins requires datasets covering the behavior of modeled objects under various conditions. However, it's impractical to collect perfect datasets for all scenarios. Thus, implementing update mechanisms is necessary to maintain high-performance operation of the digital twin system.
\end{itemize}
}


\subsection{Deterministic response}
The capability of deterministic response signifies an ability to produce predictable and certain outcomes or behavior based on specific inputs or initial conditions.  With deterministic response, there is no randomness or uncertainty, and the future state of the system can be precisely determined based on its current state and the inputs it receives. 

In the current IIoT intelligent architecture, Time Sensitive Networking (TSN) ensures deterministic response for network traffics. However, as the industrial digital transformation process accelerates, many time-critical computing tasks are generated by networked devices, such as sensor data analysis and motion planning for robots. Timely completion of the computing tasks is crucial for ensuring the correct and efficient operations of manufacturing systems. Moreover, as the cooperative relationships between robots, sensors, software and models in IIoT become more complex, there is an increasing expectation for the response delays of certain time-critical computing tasks to be deterministic. Current end-edge-cloud cooperative computing architecture, which relies on best-effort networks to forward data, leads to uncertainty and unpredictability in the response time of IIoT computing tasks. Therefore, the issue of ensuring the deterministic response of IIoT computing tasks has not been studied yet. It will become a future research direction.

\textcolor{black}{Building an innovative computing and networks convergence paradigm with the capability of deterministic responses will be helpful. Some researchers have started to explore such a paradigm \cite{hu20223cl,QingminDeterChina2022,YujiaoChina,jia2024deterministic}, but some problems are waiting to be deeply studied. 
\begin{itemize}
    \item How to schedule existing IIoT computation and network resources to ensure deterministic responses for scalable computing tasks?
    \item How to design sharing mechanisms that support deterministic scheduling of both communication traffics and computing tasks on the same IIoT networks?
\end{itemize}
}

\subsection{Cost-friendly operation and deployment}
Cost-friendly operation refers to the practice of managing and conducting business operations in a manner that optimizes costs, enhances efficiency, and ensures financial sustainability. Cloud manufacturing has provided such a cost-friendly operation that allows on-demand manufacturing. However, most researchers focus on fine-grained manufacturing resource scheduling in the perspective of cloud platform instead of manufacturing resource providers. Therefore, attention should be given to the pricing and supply strategies of manufacturing resources for manufacturing enterprises, in order to enable them to obtain higher profits at lower operating costs.

Cost-friendly deployment refers to the strategic and efficient implementation of systems or technologies with a focus on optimizing costs while achieving desired outcomes. 
The cost-friendly deployment of manufacturing processes is often realized by professionals who specialize in various aspects of industrial engineering, production management, and process optimization. 
We won't interfere with the deployment optimization of manufacturing processes, \textcolor{black}{but consider the cost-friendly deployment of computation and storage resources to ensure industrial data storage and analysis. } Cloud-based deployment and edge-based deployment are able to provide cost-friendly deployment strategies for industrial big data and systems. However, there are many time-critical tasks in IIoT. \textcolor{black}{Both cloud-based deployment and edge-based deployment can lead to failures in meeting the time requirements of tasks due to unexpected and unpredictable latency introduced by the usage of best-effort networks. Therefore, following two challenges should be studied in the future to build cost-friendly deployment. }
\textcolor{black}{
\begin{itemize}
    \item Given the personalized deployment costs associated with computing servers and network devices, how to design a cost-effective deployment solution for IIoT enterprises that lack a foundation for intelligent transformation?
    \item For IIoT enterprises with established intelligent transformation foundations, how can the current deployment structure of computing servers and network devices be optimized to cater to the personalized transformation needs?
\end{itemize}
}

\subsection{IIoT intelligence proliferation}
The intelligence proliferation refers to the development or creation of new intelligent systems, models, and algorithms through AI, to proceed IIoT tasks or emergency situation. The reasons why the intelligent proliferation is required in future smart manufacturing include:
\begin{itemize}
    \item Industrial manufacturing involves intricate production processes, equipment, and systems, making it challenging to accurately describe and address all possible scenarios through traditional manual rules. The diversity and complexity of smart manufacturing environments require systems to adapt to different conditions and requirements.
    \item  Designing and optimizing specialized algorithms for each possible scenario is a time-consuming task. Relying solely on manually designed case-by-case algorithms may not meet the demands of rapid changes and complexity in dynamic manufacturing environments.
    \item Intelligent emergence systems possess adaptability and learning capabilities, continuously optimizing their models and algorithms based on real-time industrial data. This enables the system to continually adapt to changing industrial environments and production requirements.
\end{itemize}

Inspired by the development of ChatGPT, which can understand and generate human-like text in conversations, retain context for coherent responses, perform tasks like answering questions and creative writing, and provide information based on its knowledge, \textcolor{black}{a large industrial generative intelligent model has potential to achieve IIoT intelligence proliferation, but it comes with several challenges:}
\textcolor{black}{
\begin{itemize}
    \item Building and maintaining large models can be costly, especially in terms of computational resources and data storage.
    \item Large models demand extensive and high-quality training data, which may be scarce or challenging to obtain in industrial settings.
    \item Deploying large models in real-world industrial environments can be complex due to resource constraints and real-time processing requirements.
\end{itemize}
}

\textcolor{black}{
\section{Ethical Implications and Environmental Impacts} \label{section:ethical}
In this section, we will explore the ethical implications and environmental impacts of adopting Industrial Internet of Things (IIoT) intelligence in manufacturing.}
\textcolor{black}{
\subsection{Ethical implications}
The ethical implications primarily include three aspects: job displacement and workforce transition, privacy and data security, social and economic equality. }
\textcolor{black}{
\subsubsection{Job displacement and workforce transition} The increased automation facilitated by IIoT intelligence may lead to job displacement for certain manual tasks, raising ethical concerns about the well-being and transition of affected workers. 
To avoid the unexpected situations, manufacturing enterprises that will realize transformation with IIoT intelligence are expected to carry out three steps \cite{glnworkers24}. Before the introduction of new technologies, companies should communicate the benefits and explain the why, make the bigger picture clear to workers of the technology, help workers to explore and become confident with the technologies, and include workers in the exchange of ideas. During the introduction of new technology, companies should make timelines clear early on, and provide a diversified skill training programmes to workers. After the introduction of new technology, companies should follow up the progress of technology applications, and encourage workers together to continue to explore new use cases. 
}
\textcolor{black}{
\subsubsection{Privacy and data security} IIoT intelligence involves vast operations of data collection and analysis,  raising privacy concerns for employees and stakeholders. 
To address this concern, companies should be transparent about their data collection, processing, and usage practices in IIoT deployments. This transparency helps build trust with employees, stakeholders, and the public. Furthermore, detailed privacy policies and data usage guidelines should be communicated clearly to all relevant parties, outlining the types of data collected, the purposes for which it is used, and the rights of individuals regarding their data. Additionally, encryption techniques should be employed to secure data both in transit and at rest, ensuring that it remains confidential and tamper-proof.}
\textcolor{black}{
\subsubsection{Social and economic equality} Larger corporations with greater financial resources and technical capabilities may have an advantage in adopting and leveraging IIoT solutions, leaving smaller manufacturers or those in less developed regions at a disadvantage. This unequal access can exacerbate existing disparities in productivity, competitiveness, and innovation within the manufacturing sector. 
To avoid the unequal access of IIoT intelligence, IIoT solution providers are expected to offer flexible and scalable offerings tailored to the needs and budget constraints of manufacturing businesses of all sizes. The development of open-source IIoT platforms and standards should be encouraged to facilitate interoperability and lowers barriers to entry for manufacturers of all scales. Initiatives such as establishing innovation clusters, technology parks, or shared manufacturing facilities can be established to encourage collaboration, knowledge sharing, and access to advanced technologies among local manufacturers. 
}
\textcolor{black}{
\subsection{Environmental Impacts}
Adopting IIoT intelligence in manufacturing can be both positive and negative. 
\subsubsection{Positive impacts}
IIoT intelligence enables better monitoring and optimization of resource usage, such as energy, water, and raw materials, leading to reduced waste and improved resource efficiency. 
Optimizing manufacturing processes through IIoT intelligence can lead to reduced emissions of greenhouse gases and other pollutants.
IIoT-enabled predictive maintenance and quality control measures reduce the likelihood of defects and production errors, resulting in less waste and scrap materials.
Additionally, IIoT intelligence provides greater visibility into supply chains, enabling more efficient logistics, inventory management, and transportation planning. 
\subsubsection{Negative impacts}
The proliferation of IIoT devices and sensors may contribute to a significant increase in electronic waste at the end of their lifecycle. 
While IIoT technologies can improve energy efficiency in manufacturing, the deployment of additional sensors, data centers, and network infrastructure may lead to increased overall energy consumption, especially, the deep application of the industrial large model will consume vase amount of electronic energy.  
The production of IIoT devices, including sensors, processors, and communication modules, requires the extraction of raw materials and energy-intensive manufacturing processes. }
\textcolor{black}{
\subsubsection{Mitigation strategies}
To reduce the side effects of IIoT intelligence on environment, some mitigation strategies are suggested. 
Firstly, IIoT devices for durability, repairability, and recyclability are encouraged to be designed to promote a circular economy approach, minimizing waste and resource consumption. Secondly, engaging suppliers committed to sustainable development, purchase environmentally friendly materials, reduce emissions related to transportation, and establish a more sustainable IIoT supply chain. Thirdly, conducting comprehensive lifecycle assessments and environmental impact analyses helps identify areas for improvement and inform decision-making throughout the IIoT deployment process. 
}

\section{Conclusion} \label{conclusion}
We propose a hierarchical development architecture for IIoT intelligence to build a systematic understanding of how IIoT intelligence deals with the issues in the transformation process of smart manufacturing. 
The architecture consists of equipment layer, networking layer, software layer, modeling layer, and analysis and optimization layer. The equipment layer builds the foundation of IIoT intelligence. The networking layer realizes ubiquitous connection among human-machine-thing. The software layer provides digital expression of industrial processes. The modeling layer achieves accurately modeling and interoperability improvement. The analysis and optimization layer supports case-by-case algorithm design to analyze industrial data and optimize industrial processes.  In addition, we also closely observe the lighthouse factory to demonstrate the positive effects of technologies at each layer. 

Following that, we identify seven kinds of technologies that accelerate the development of IIoT intelligence, including industrial robots, machine vision, networking technologies, digital twins, deep learning, smart hardware, cloud/edge computing. 
\textcolor{black}{Subsequently, we analyzes the ethical implications and environmental impacts of adopting IIoT intelligence in manufacturing. }

Finally, we explore the open challenges and discuss the future research directions from four aspects, including developing digital twins to realize ideal vision of digital control, exploring deterministic response strategies and mechanisms for IIoT computing tasks, optimizing computation and network resources deployment to reduce costs, and researching industrial generative AI large model to realize intelligence proliferation.

\bibliographystyle{IEEEtranbst}
\bibliography{huyujiao}

\begin{thebibliography}{100}
\providecommand{\url}[1]{#1}
\csname url@samestyle\endcsname
\providecommand{\newblock}{\relax}
\providecommand{\bibinfo}[2]{#2}
\providecommand{\BIBentrySTDinterwordspacing}{\spaceskip=0pt\relax}
\providecommand{\BIBentryALTinterwordstretchfactor}{4}
\providecommand{\BIBentryALTinterwordspacing}{\spaceskip=\fontdimen2\font plus
\BIBentryALTinterwordstretchfactor\fontdimen3\font minus \fontdimen4\font\relax}
\providecommand{\BIBforeignlanguage}[2]{{%
\expandafter\ifx\csname l@#1\endcsname\relax
\typeout{** WARNING: IEEEtran.bst: No hyphenation pattern has been}%
\typeout{** loaded for the language `#1'. Using the pattern for}%
\typeout{** the default language instead.}%
\else
\language=\csname l@#1\endcsname
\fi
#2}}
\providecommand{\BIBdecl}{\relax}
\BIBdecl

\bibitem{theproportionmanufacturing}
\BIBentryALTinterwordspacing
T.~world Bank, ``World development indicators: Structure of output,'' Website, 2019. [Online]. Available: \url{http://wdi.worldbank.org/table/4.2#}
\BIBentrySTDinterwordspacing

\bibitem{AmericanPolicy}
\BIBentryALTinterwordspacing
{National Science and Technology Council,Networking and Information Technology Research and Development Subcommittee}, ``The national artificial intelligence research and development strategic paln,'' Website, 2016. [Online]. Available: \url{https://www.nitrd.gov/pubs/national_ai_rd_strategic_plan.pdf}
\BIBentrySTDinterwordspacing

\bibitem{JapanesePolicy}
\BIBentryALTinterwordspacing
{The Headquarters for Japan’s Economic Revitalization}, ``New robot strategy,'' Website, 2015. [Online]. Available: \url{https://www.meti.go.jp/english/press/2015/pdf/0123_01b.pdf}
\BIBentrySTDinterwordspacing

\bibitem{GermanPolicy}
\BIBentryALTinterwordspacing
{Federal Ministry for Economic Affairs and Energy}, ``Industrial strategy 2030: Guidelines for a german and european industrial policy,'' Website, 2019. [Online]. Available: \url{https://www.bmwi.de/Redaktion/EN/Publikationen/Industry/industrial-strategy-2030.pdf?__blob=publicationFile&v=7}
\BIBentrySTDinterwordspacing

\bibitem{MadeInCHina2025}
\BIBentryALTinterwordspacing
{Ministry of Industry and Information Technology of the People's Republic of China}, ``{Made in China 2025},'' Website, 2015. [Online]. Available: \url{https://www.miit.gov.cn/zwgk/zcjd/art/2020/art_8f85af6a7785410b85c6acf1a73f357e.html}
\BIBentrySTDinterwordspacing

\bibitem{QingdaoHaier5GSmartFactory}
\BIBentryALTinterwordspacing
{Huawei}, ``{Qingdao Haier 5G Smart Factory},'' Website, 2020. [Online]. Available: \url{https://www.huawei.com/en/events/5g-core-summit-2020/videos/5g-smart-factory}
\BIBentrySTDinterwordspacing

\bibitem{MideaChinaUnicomandHuawei}
\BIBentryALTinterwordspacing
------, ``{Midea, China Unicom, and Huawei Jointly Release 5G Converged Positioning Solution},'' Website, 2021. [Online]. Available: \url{https://www.huawei.com/en/news/2021/7/midea-chinaunicom-5g-intelligent-manufacturing}
\BIBentrySTDinterwordspacing

\bibitem{qiu2020edge}
T.~Qiu, J.~Chi, X.~Zhou, Z.~Ning, M.~Atiquzzaman, and D.~O. Wu, ``Edge computing in industrial internet of things: Architecture, advances and challenges,'' \emph{IEEE Communications Surveys \& Tutorials}, vol.~22, no.~4, pp. 2462--2488, 2020.

\bibitem{boyes2018industrial}
H.~Boyes, B.~Hallaq, J.~Cunningham, and T.~Watson, ``The industrial internet of things (iiot): An analysis framework,'' \emph{Computers in industry}, vol. 101, pp. 1--12, 2018.

\bibitem{khan2020industrial}
W.~Z. Khan, M.~Rehman, H.~M. Zangoti, M.~K. Afzal, N.~Armi, and K.~Salah, ``Industrial internet of things: Recent advances, enabling technologies and open challenges,'' \emph{Computers \& electrical engineering}, vol.~81, p. 106522, 2020.

\bibitem{malik2021industrial}
P.~K. Malik, R.~Sharma, R.~Singh, A.~Gehlot, S.~C. Satapathy, W.~S. Alnumay, D.~Pelusi, U.~Ghosh, and J.~Nayak, ``Industrial internet of things and its applications in industry 4.0: State of the art,'' \emph{Computer Communications}, vol. 166, pp. 125--139, 2021.

\bibitem{8879484}
L.~Chettri and R.~Bera, ``A comprehensive survey on internet of things (iot) toward 5g wireless systems,'' \emph{IEEE Internet of Things Journal}, vol.~7, no.~1, pp. 16--32, 2020.

\bibitem{9548837}
A.~Mahmood, L.~Beltramelli, S.~Fakhrul~Abedin, S.~Zeb, N.~I. Mowla, S.~A. Hassan, E.~Sisinni, and M.~Gidlund, ``Industrial iot in 5g-and-beyond networks: Vision, architecture, and design trends,'' \emph{IEEE Transactions on Industrial Informatics}, vol.~18, no.~6, pp. 4122--4137, 2022.

\bibitem{finn2018introduction}
N.~Finn, ``Introduction to time-sensitive networking,'' \emph{IEEE Communications Standards Magazine}, vol.~2, no.~2, pp. 22--28, 2018.

\bibitem{messenger2018time}
J.~L. Messenger, ``Time-sensitive networking: An introduction,'' \emph{IEEE Communications Standards Magazine}, vol.~2, no.~2, pp. 29--33, 2018.

\bibitem{farkas2018time}
J.~Farkas, L.~L. Bello, and C.~Gunther, ``Time-sensitive networking standards,'' \emph{IEEE Communications Standards Magazine}, vol.~2, no.~2, pp. 20--21, 2018.

\bibitem{bello2019perspective}
L.~L. Bello and W.~Steiner, ``A perspective on ieee time-sensitive networking for industrial communication and automation systems,'' \emph{Proceedings of the IEEE}, vol. 107, no.~6, pp. 1094--1120, 2019.

\bibitem{9321458}
R.~A. Khalil, N.~Saeed, M.~Masood, Y.~M. Fard, M.-S. Alouini, and T.~Y. Al-Naffouri, ``Deep learning in the industrial internet of things: Potentials, challenges, and emerging applications,'' \emph{IEEE Internet of Things Journal}, vol.~8, no.~14, pp. 11\,016--11\,040, 2021.

\bibitem{wei2021consistency}
Y.~Wei, T.~Hu, T.~Zhou, Y.~Ye, and W.~Luo, ``Consistency retention method for cnc machine tool digital twin model,'' \emph{Journal of Manufacturing Systems}, vol.~58, pp. 313--322, 2021.

\bibitem{gregorio2021digital}
J.-L. Gr{\'e}gorio, C.~Lartigue, F.~Thi{\'e}baut, and R.~Lebrun, ``A digital twin-based approach for the management of geometrical deviations during assembly processes,'' \emph{Journal of Manufacturing Systems}, vol.~58, pp. 108--117, 2021.

\bibitem{Remotehumanrobotcollaboration2020}
H.~Liu and L.~Wang, ``Remote human--robot collaboration: A cyber--physical system application for hazard manufacturing environment,'' \emph{Journal of manufacturing systems}, vol.~54, pp. 24--34, 2020.

\bibitem{Germanrobots2017}
W.~Dauth, S.~Findeisen, J.~S{\"u}dekum, and N.~Woessner, ``German robots-the impact of industrial robots on workers,'' 2017.

\bibitem{Evolutionindustrialrobots2013}
B.~Singh, N.~Sellappan, and P.~Kumaradhas, ``Evolution of industrial robots and their applications,'' \emph{International Journal of emerging technology and advanced engineering}, vol.~3, no.~5, pp. 763--768, 2013.

\bibitem{ambika2020machine}
P.~Ambika, ``Machine learning and deep learning algorithms on the industrial internet of things (iiot),'' \emph{Advances in computers}, vol. 117, no.~1, pp. 321--338, 2020.

\bibitem{de2022deep}
S.~De, M.~Bermudez-Edo, H.~Xu, and Z.~Cai, ``Deep generative models in the industrial internet of things: a survey,'' \emph{IEEE Transactions on Industrial Informatics}, vol.~18, no.~9, pp. 5728--5737, 2022.

\bibitem{laroui2021edge}
M.~Laroui, B.~Nour, H.~Moungla, M.~A. Cherif, H.~Afifi, and M.~Guizani, ``Edge and fog computing for iot: A survey on current research activities \& future directions,'' \emph{Computer Communications}, vol. 180, pp. 210--231, 2021.

\bibitem{hamdan2020edge}
S.~Hamdan, M.~Ayyash, and S.~Almajali, ``Edge-computing architectures for internet of things applications: A survey,'' \emph{Sensors}, vol.~20, no.~22, p. 6441, 2020.

\bibitem{IndustrialRevolution4}
\BIBentryALTinterwordspacing
{World Economic Forum, McKinsey \& Company}, ``Fourth industrial revolution: Beacons of technology and innovation in manufacturing,'' Website, January 2019. [Online]. Available: \url{http://www3.weforum.org/docs/WEF_4IR_Beacons_of_Technology_and_Innovation_in_Manufacturing_report_2019.pdf}
\BIBentrySTDinterwordspacing

\bibitem{globalLighthouse}
\BIBentryALTinterwordspacing
------, ``Global lighthouse network: Insights from the forefront of the fourth industrial revolution,'' Website, December 2019. [Online]. Available: \url{http://www3.weforum.org/docs/WEF_Global_Lighthouse_Network.pdf}
\BIBentrySTDinterwordspacing

\bibitem{globalLighthouse2022}
\BIBentryALTinterwordspacing
------, ``Global lighthouse network: Shaping the next chapter of the forth industial revolution,'' Website, January 2023. [Online]. Available: \url{https://www.weforum.org/publications/global-lighthouse-network-shaping-the-next-chapter-of-the-fourth-industrial-revolution/}
\BIBentrySTDinterwordspacing

\bibitem{globalLighthouse2023}
\BIBentryALTinterwordspacing
------, ``Global lighthouse network: Adopting ai at speed and scale,'' Website, December 2023. [Online]. Available: \url{https://www.weforum.org/publications/manufacturing-lighthouses-and-the-path-to-impact-adopting-ai-at-speed-and-scale/}
\BIBentrySTDinterwordspacing

\bibitem{ChinesePopularity}
\BIBentryALTinterwordspacing
{Chinese National Bureau of Statistics}, Website, 2020. [Online]. Available: \url{https://data.stats.gov.cn/easyquery.htm?cn=C01&zb=A0401&sj=2020}
\BIBentrySTDinterwordspacing

\bibitem{AmericaAI2023Whitehouse}
\BIBentryALTinterwordspacing
{White House}, ``Executive order on the safe, secure, and trustworthy development and use of artificial intelligence,'' Website, October, 2023. [Online]. Available: \url{https://www.whitehouse.gov/briefing-room/presidential-actions/2023/10/30/executive-order-on-the-safe-secure-and-trustworthy-development-and-use-of-artificial-intelligence/}
\BIBentrySTDinterwordspacing

\bibitem{AmericaAI2023highlights}
\BIBentryALTinterwordspacing
L.~A. Harris and J.~Chris, ``Highlights of the 2023 executive order on artificial intelligence for congress,'' Website, November 2023. [Online]. Available: \url{https://crsreports.congress.gov/product/pdf/R/R47843}
\BIBentrySTDinterwordspacing

\bibitem{Europ2023GreenIndPlan}
\BIBentryALTinterwordspacing
{European Commission}, ``A green deal industrial plan for the net-zero age,'' Website, February, 2023. [Online]. Available: \url{https://commission.europa.eu/document/download/41514677-9598-4d89-a572-abe21cb037f4_en?filename=COM_2023_62_2_EN_ACT_A%20Green%20Deal%20Industrial%20Plan%20for%20the%20Net-Zero%20Age.pdf}
\BIBentrySTDinterwordspacing

\bibitem{Europ2024NetIndustrialPlan}
\BIBentryALTinterwordspacing
------, ``Commission welcomes political agreement to make clean technology manufacturing in the eu resilient and competitive,'' Website, February, 2024. [Online]. Available: \url{https://ec.europa.eu/commission/presscorner/detail/en/ip_24_680}
\BIBentrySTDinterwordspacing

\bibitem{ChinaStandarad2035}
\BIBentryALTinterwordspacing
{Chinese Government}, ``China standards 2035,'' Website, August 2023. [Online]. Available: \url{https://www.sac.gov.cn/xw/bzhdt/art/2023/art_adaa8006e0d149008617f1b2e07cbc77.html}
\BIBentrySTDinterwordspacing

\bibitem{ChinaNewplan2023}
\BIBentryALTinterwordspacing
{Global Times}, ``China releases implementation plan for new industries’ standards,'' Website, August 2023. [Online]. Available: \url{https://www.globaltimes.cn/page/202308/1296789.shtml}
\BIBentrySTDinterwordspacing

\bibitem{WhitePaperofFoxconn}
\BIBentryALTinterwordspacing
{Foxconn Industrial Internet, EqualOcean and Tencent CLoud}, ``Smart manufacturing milestones: The lighthouse factories leading digital transformation of chinese manufacturing,'' Website, June 2020. [Online]. Available: \url{http://www.d-long.com/eWebEditor/uploadfile/2020072018285832520761.pdf}
\BIBentrySTDinterwordspacing

\bibitem{technologyEcosystem}
\BIBentryALTinterwordspacing
{World Economic Forum}, ``Platforms and ecosystems: Enabling the digital economy,'' Website, February 2019. [Online]. Available: \url{http://www3.weforum.org/docs/WEF_Digital_Platforms_and_Ecosystems_2019.pdf}
\BIBentrySTDinterwordspacing

\bibitem{xu2012cloud}
X.~Xu, ``From cloud computing to cloud manufacturing,'' \emph{Robotics and computer-integrated manufacturing}, vol.~28, no.~1, pp. 75--86, 2012.

\bibitem{liu2017workload}
Y.~Liu, X.~Xu, L.~Zhang, L.~Wang, and R.~Y. Zhong, ``Workload-based multi-task scheduling in cloud manufacturing,'' \emph{Robotics and Computer-integrated manufacturing}, vol.~45, pp. 3--20, 2017.

\bibitem{bello2021cloud}
S.~A. Bello, L.~O. Oyedele, O.~O. Akinade, M.~Bilal, J.~M.~D. Delgado, L.~A. Akanbi, A.~O. Ajayi, and H.~A. Owolabi, ``Cloud computing in construction industry: Use cases, benefits and challenges,'' \emph{Automation in Construction}, vol. 122, p. 103441, 2021.

\bibitem{gupta2020overview}
B.~B. Gupta and M.~Quamara, ``An overview of internet of things (iot): Architectural aspects, challenges, and protocols,'' \emph{Concurrency and Computation: Practice and Experience}, vol.~32, no.~21, p. e4946, 2020.

\bibitem{IFRPresentationPPT}
\BIBentryALTinterwordspacing
{IFR}, ``World robotics 2020,'' IFR Press Conference, 24th September 2020. [Online]. Available: \url{https://ifr.org/downloads/press2018/Presentation_WR_2020.pdf}
\BIBentrySTDinterwordspacing

\bibitem{carbonfootprintrobots2020}
B.~He, X.~Cao, and Z.~Gu, ``Kinematics of underactuated robotics for product carbon footprint,'' \emph{Journal of Cleaner Production}, vol. 257, p. 120491, 2020.

\bibitem{nilakantan2017multi}
J.~M. Nilakantan, Z.~Li, Q.~Tang, and P.~Nielsen, ``Multi-objective co-operative co-evolutionary algorithm for minimizing carbon footprint and maximizing line efficiency in robotic assembly line systems,'' \emph{Journal of Cleaner Production}, vol. 156, pp. 124--136, 2017.

\bibitem{progressonprogramming2012}
Z.~Pan, J.~Polden, N.~Larkin, S.~Van~Duin, and J.~Norrish, ``Recent progress on programming methods for industrial robots,'' \emph{Robotics and Computer-Integrated Manufacturing}, vol.~28, no.~2, pp. 87--94, 2012.

\bibitem{vysocky2016human}
A.~Vysocky and P.~Novak, ``Human-robot collaboration in industry,'' \emph{MM Science Journal}, vol.~9, no.~2, pp. 903--906, 2016.

\bibitem{villani2018survey}
V.~Villani, F.~Pini, F.~Leali, and C.~Secchi, ``Survey on human--robot collaboration in industrial settings: Safety, intuitive interfaces and applications,'' \emph{Mechatronics}, vol.~55, pp. 248--266, 2018.

\bibitem{industrialsensors2023}
\BIBentryALTinterwordspacing
{Introduction to Industrial Sensors}, Website, 2023. [Online]. Available: \url{https://control.com/technical-articles/introduction-to-industrial-sensors/#:~:text=Fundamentally%2C%20every%20industrial%20sensor%20is,and%20converted%20to%20an%20output.}
\BIBentrySTDinterwordspacing

\bibitem{cloudcomputing-Azure}
\BIBentryALTinterwordspacing
{What is cloud computing}, Website, 2023. [Online]. Available: \url{https://azure.microsoft.com/en-us/resources/cloud-computing-dictionary/what-is-cloud-computing#:~:text=Simply%20put%2C%20cloud%20computing%20is,resources%2C%20and%20economies%20of%20scale.}
\BIBentrySTDinterwordspacing

\bibitem{GlobalNetworkingTrends2020}
\BIBentryALTinterwordspacing
{CISILION}, ``2020 global networking trends report,'' Website, 2020. [Online]. Available: \url{https://cdn2.hubspot.net/hubfs/302795/Cisco%20Global%20Networking%20Trends%20Report%202020.pdf?__hsfp=2303363933&__hssc=251652889.2.1632299031680&__hstc=251652889.4a87718e0ddc358f0f8252747dd7b7df.1632299031680.1632299031680.1632299031680.1}
\BIBentrySTDinterwordspacing

\bibitem{nasrallah2018ultra}
A.~Nasrallah, A.~S. Thyagaturu, Z.~Alharbi, C.~Wang, X.~Shao, M.~Reisslein, and H.~ElBakoury, ``Ultra-low latency (ull) networks: The ieee tsn and ietf detnet standards and related 5g ull research,'' \emph{IEEE Communications Surveys \& Tutorials}, vol.~21, no.~1, pp. 88--145, 2018.

\bibitem{WhitePaperZhongxingDeqin}
\BIBentryALTinterwordspacing
{ZTE, Deloitte}, ``White paper on 5g+ict industry trends: Innovation, survival and development,'' 2020. [Online]. Available: \url{https://res-www.zte.com.cn/mediares/zte/Files/PDF/white_book/202002210916.pdf?la=zh-CN}
\BIBentrySTDinterwordspacing

\bibitem{TFprediciton20200913}
\BIBentryALTinterwordspacing
S.~Haibing and L.~Xinjun, ``In-depth perspective on mes,'' Sept. 13, 2020. [Online]. Available: \url{https://www.vzkoo.com/doc/20022.html?a=4}
\BIBentrySTDinterwordspacing

\bibitem{fang2013loss}
S.~Fang, Y.~Yu, C.~H. Foh, and K.~M.~M. Aung, ``A loss-free multipathing solution for data center network using software-defined networking approach,'' \emph{IEEE transactions on magnetics}, vol.~49, no.~6, pp. 2723--2730, 2013.

\bibitem{hu2014SDNsurvey}
F.~Hu, Q.~Hao, and K.~Bao, ``A survey on software-defined network and openflow: From concept to implementation,'' \emph{IEEE Communications Surveys \& Tutorials}, vol.~16, no.~4, pp. 2181--2206, 2014.

\bibitem{xylomenos2013survey}
G.~Xylomenos, C.~N. Ververidis, V.~A. Siris, N.~Fotiou, C.~Tsilopoulos, X.~Vasilakos, K.~V. Katsaros, and G.~C. Polyzos, ``A survey of information-centric networking research,'' \emph{IEEE communications surveys \& tutorials}, vol.~16, no.~2, pp. 1024--1049, 2013.

\bibitem{vasilakos2015information}
A.~V. Vasilakos, Z.~Li, G.~Simon, and W.~You, ``Information centric network: Research challenges and opportunities,'' \emph{Journal of network and computer applications}, vol.~52, pp. 1--10, 2015.

\bibitem{shafi20175g}
M.~Shafi, A.~F. Molisch, P.~J. Smith, T.~Haustein, P.~Zhu, P.~De~Silva, F.~Tufvesson, A.~Benjebbour, and G.~Wunder, ``5g: A tutorial overview of standards, trials, challenges, deployment, and practice,'' \emph{IEEE journal on selected areas in communications}, vol.~35, no.~6, pp. 1201--1221, 2017.

\bibitem{agiwal2016next}
M.~Agiwal, A.~Roy, and N.~Saxena, ``Next generation 5g wireless networks: A comprehensive survey,'' \emph{IEEE Communications Surveys \& Tutorials}, vol.~18, no.~3, pp. 1617--1655, 2016.

\bibitem{afolabi2018network}
I.~Afolabi, T.~Taleb, K.~Samdanis, A.~Ksentini, and H.~Flinck, ``Network slicing and softwarization: A survey on principles, enabling technologies, and solutions,'' \emph{IEEE Communications Surveys \& Tutorials}, vol.~20, no.~3, pp. 2429--2453, 2018.

\bibitem{alliance2016description}
N.~Alliance, ``Description of network slicing concept,'' \emph{NGMN 5G P}, vol.~1, no.~1, 2016.

\bibitem{SoftwareWhitePaper2019}
\BIBentryALTinterwordspacing
Y.~Wang, ``{White Paper on Industrial Software Development in China},'' Website, 2019. [Online]. Available: \url{https://v1.cecdn.yun300.cn/site_1801180113%2F%E4%B8%AD%E5%9B%BD%E5%B7%A5%E4%B8%9A%E8%BD%AF%E4%BB%B6%E5%8F%91%E5%B1%95%E7%99%BD%E7%9A%AE%E4%B9%A6%282019%291565910127066.pdf}
\BIBentrySTDinterwordspacing

\bibitem{wu2017digital}
D.~Wu, J.~Terpenny, and D.~Schaefer, ``Digital design and manufacturing on the cloud: A review of software and services—retracted,'' \emph{AI EDAM}, vol.~31, no.~1, pp. 104--118, 2017.

\bibitem{mora2017design}
M.~Mora, R.~O'Connor, F.~Tsui, and J.~Marx~G{\'o}mez, ``Design methods for software architectures in the service-oriented computing and cloud paradigms,'' \emph{Software: Practice and Experience}, vol.~48, no.~2, pp. 263--267, 2017.

\bibitem{NationalDevelopmentandReformCommission2020}
{National Development and Reform Commission, Administration of the CPC Central Committee}, ``The implementation plan of promoting the action of "using big-data to empower wisdom in the cloud" for new economic development,'' Website, 2020-04-07, \url{https://www.ndrc.gov.cn/xxgk/zcfb/tz/202004/t20200410_1225542_ext.html} (accessed 10 September 2021).

\bibitem{DHL2019DHL}
{DHL Trend Research}, ``Digital twins in logistics: A dhl perspective on the impact of digital twins on the logistics industry,'' Website, 2019, \url{https://www.dhl.com/content/dam/dhl/global/core/documents/pdf/glo-core-digital-twins-in-logistics.pdf} (accessed 10 September 2021).

\bibitem{alsumait2002use}
A.~Alsumait, A.~Seffah, and T.~Radhakrishnan, ``Use case maps: A roadmap for usability and software integrated specification,'' in \emph{IFIP World Computer Congress, TC 13}.\hskip 1em plus 0.5em minus 0.4em\relax Springer, 2002, pp. 119--131.

\bibitem{buhr1998use}
R.~J. Buhr, ``Use case maps as architectural entities for complex systems,'' \emph{IEEE Transactions on Software Engineering}, vol.~24, no.~12, pp. 1131--1155, 1998.

\bibitem{shen2006agent}
W.~Shen, L.~Wang, and Q.~Hao, ``Agent-based distributed manufacturing process planning and scheduling: a state-of-the-art survey,'' \emph{IEEE Transactions on Systems, Man, and Cybernetics, Part C (Applications and Reviews)}, vol.~36, no.~4, pp. 563--577, 2006.

\bibitem{chang2000integrated}
P.-T. Chang and C.-H. Chang, ``An integrated artificial intelligent computer-aided process planning system,'' \emph{International Journal of Computer Integrated Manufacturing}, vol.~13, no.~6, pp. 483--497, 2000.

\bibitem{chen2023scheduling}
J.~Chen, P.~Han, Y.~Zhang, T.~You, and P.~Zheng, ``Scheduling energy consumption-constrained workflows in heterogeneous multi-processor embedded systems,'' \emph{Journal of Systems Architecture}, vol. 142, p. 102938, 2023.

\bibitem{niebel1965mechanized}
B.~W. Niebel, ``Mechanized process selection for planning new designs,'' \emph{ASME paper}, vol. 737, 1965.

\bibitem{yusof2014survey}
Y.~Yusof and K.~Latif, ``Survey on computer-aided process planning,'' \emph{The international journal of advanced manufacturing technology}, vol.~75, no. 1-4, pp. 77--89, 2014.

\bibitem{thong2015survey}
W.~J. Thong and M.~Ameedeen, ``A survey of petri net tools,'' in \emph{Advanced computer and communication engineering technology}.\hskip 1em plus 0.5em minus 0.4em\relax Springer, 2015, pp. 537--551.

\bibitem{zhang2011petri}
X.~Zhang, Q.~Lu, and T.~Wu, ``Petri-net based applications for supply chain management: an overview,'' \emph{International Journal of Production Research}, vol.~49, no.~13, pp. 3939--3961, 2011.

\bibitem{quintanilla2016petri}
F.~G. Quintanilla, O.~Cardin, A.~L’Anton, and P.~Castagna, ``A petri net-based methodology to increase flexibility in service-oriented holonic manufacturing systems,'' \emph{Computers in Industry}, vol.~76, pp. 53--68, 2016.

\bibitem{guo2017timed}
Z.~Guo, Y.~Zhang, X.~Zhao, and X.~Song, ``A timed colored petri net simulation-based self-adaptive collaboration method for production-logistics systems,'' \emph{Applied Sciences}, vol.~7, no.~3, p. 235, 2017.

\bibitem{kretschmer2017knowledge}
R.~Kretschmer, A.~Pfouga, S.~Rulhoff, and J.~Stjepandi{\'c}, ``Knowledge-based design for assembly in agile manufacturing by using data mining methods,'' \emph{Advanced Engineering Informatics}, vol.~33, pp. 285--299, 2017.

\bibitem{tsai2010knowledge}
Y.-L. Tsai, C.-F. You, J.-Y. Lin, and K.-Y. Liu, ``Knowledge-based engineering for process planning and die design for automotive panels,'' \emph{Computer-Aided Design and Applications}, vol.~7, no.~1, pp. 75--87, 2010.

\bibitem{li2011recent}
B.~M. Li, S.~Q. Xie, and X.~Xu, ``Recent development of knowledge-based systems, methods and tools for one-of-a-kind production,'' \emph{Knowledge-Based Systems}, vol.~24, no.~7, pp. 1108--1119, 2011.

\bibitem{wooldridge1995intelligent}
M.~J. Wooldridge and N.~R. Jennings, ``Intelligent agents: Theory and practice,'' \emph{The knowledge engineering review}, vol.~10, no.~2, pp. 115--152, 1995.

\bibitem{Iglesias1998A}
C.~A. Iglesias and M.~Garijo, ``A survey of agent-oriented methodologies,'' in \emph{International Workshop on Intelligent Agents V, Agent Theories, Architectures, and Languages}, 1998, pp. 317--330.

\bibitem{Tveit2001A}
A.~Tveit, ``A survey of agent-oriented software engineering,'' \emph{Journal of Computer Engineering Research}, 2001.

\bibitem{Sudeikat2004Evaluation}
J.~Sudeikat, L.~Braubach, A.~Pokahr, and W.~Lamersdorf, ``Evaluation of agent–oriented software methodologies – examination of the gap between modeling and platform,'' in \emph{International Conference on Agent-Oriented Software Engineering}, 2004, pp. 126--141.

\bibitem{Tonn2010ASGARD}
J.~Tonn and S.~Kaiser, ``Asgard – a graphical monitoring tool for distributed agent infrastructures,'' in \emph{Advances in Practical Applications of Agents and Multiagent Systems, International Conference on Practical Applications of Agents and Multiagent Systems, Paams 2010, Salamanca, Spain, 26-28 April}, 2010, pp. 163--173.

\bibitem{Lin2005Tool}
P.~Lin, J.~Thangarajah, and M.~Winikoff, ``Tool support for agent development using the prometheus methodology,'' in \emph{International Conference on Quality Software}, 2005, pp. 383--388.

\bibitem{Kravari2015A}
K.~Kravari and N.~Bassiliades, ``A survey of agent platforms,'' \emph{Journal of Artificial Societies \& Social Simulation}, vol.~18, no.~1, 2015.

\bibitem{Bergenti2017Agent}
F.~Bergenti, E.~Iotti, S.~Monica, and A.~Poggi, ``Agent-oriented model-driven development for jade with the jadel programming language,'' \emph{Computer Languages Systems \& Structures}, 2017.

\bibitem{zhang2007agent}
W.~Zhang and S.~Xie, ``Agent technology for collaborative process planning: a review,'' \emph{The International Journal of Advanced Manufacturing Technology}, vol.~32, no.~3, pp. 315--325, 2007.

\bibitem{li2010agent}
X.~Li, C.~Zhang, L.~Gao, W.~Li, and X.~Shao, ``An agent-based approach for integrated process planning and scheduling,'' \emph{Expert Systems with Applications}, vol.~37, no.~2, pp. 1256--1264, 2010.

\bibitem{sarkar2018multi}
A.~Sarkar and D.~{\v{S}}ormaz, ``Multi-agent system for cloud manufacturing process planning,'' \emph{Procedia manufacturing}, vol.~17, pp. 435--443, 2018.

\bibitem{torreno2017cooperative}
A.~Torre{\~n}o, E.~Onaindia, A.~Komenda, and M.~{\v{S}}tolba, ``Cooperative multi-agent planning: A survey,'' \emph{ACM Computing Surveys (CSUR)}, vol.~50, no.~6, pp. 1--32, 2017.

\bibitem{gendreau2005metaheuristics}
M.~Gendreau and J.-Y. Potvin, ``Metaheuristics in combinatorial optimization,'' \emph{Annals of Operations Research}, vol. 140, no.~1, pp. 189--213, 2005.

\bibitem{blum2011hybrid}
C.~Blum, J.~Puchinger, G.~R. Raidl, and A.~Roli, ``Hybrid metaheuristics in combinatorial optimization: A survey,'' \emph{Applied soft computing}, vol.~11, no.~6, pp. 4135--4151, 2011.

\bibitem{sigmund2013topology}
O.~Sigmund and K.~Maute, ``Topology optimization approaches,'' \emph{Structural and Multidisciplinary Optimization}, vol.~48, no.~6, pp. 1031--1055, 2013.

\bibitem{zuo2006manufacturing}
K.-T. Zuo, L.-P. Chen, Y.-Q. Zhang, and J.~Yang, ``Manufacturing-and machining-based topology optimization,'' \emph{The international journal of advanced manufacturing technology}, vol.~27, no. 5-6, pp. 531--536, 2006.

\bibitem{liu2016survey}
J.~Liu and Y.~Ma, ``A survey of manufacturing oriented topology optimization methods,'' \emph{Advances in Engineering Software}, vol. 100, pp. 161--175, 2016.

\bibitem{marck2012topology}
G.~Marck, M.~Nemer, J.-L. Harion, S.~Russeil, and D.~Bougeard, ``Topology optimization using the simp method for multiobjective conductive problems,'' \emph{Numerical Heat Transfer, Part B: Fundamentals}, vol.~61, no.~6, pp. 439--470, 2012.

\bibitem{long2016optimization}
H.~Long, Y.~Hu, X.~Jin, H.~Yu, and H.~Zhu, ``An optimization procedure for spot-welded structures based on simp method,'' \emph{Computational Materials Science}, vol. 117, pp. 602--607, 2016.

\bibitem{yamada2010topology}
T.~Yamada, K.~Izui, S.~Nishiwaki, and A.~Takezawa, ``A topology optimization method based on the level set method incorporating a fictitious interface energy,'' \emph{Computer Methods in Applied Mechanics and Engineering}, vol. 199, no. 45-48, pp. 2876--2891, 2010.

\bibitem{zhou2015minimum}
M.~Zhou, B.~S. Lazarov, F.~Wang, and O.~Sigmund, ``Minimum length scale in topology optimization by geometric constraints,'' \emph{Computer Methods in Applied Mechanics and Engineering}, vol. 293, pp. 266--282, 2015.

\bibitem{liu20153d}
J.~Liu and Y.-S. Ma, ``3d level-set topology optimization: a machining feature-based approach,'' \emph{Structural and Multidisciplinary Optimization}, vol.~52, no.~3, pp. 563--582, 2015.

\bibitem{liu2018uniform}
J.~Liu, L.~Li, and Y.~Ma, ``Uniform thickness control without pre-specifying the length scale target under the level set topology optimization framework,'' \emph{Advances in Engineering Software}, vol. 115, pp. 204--216, 2018.

\bibitem{diez2019data}
A.~Diez-Olivan, J.~Del~Ser, D.~Galar, and B.~Sierra, ``Data fusion and machine learning for industrial prognosis: Trends and perspectives towards industry 4.0,'' \emph{Information Fusion}, vol.~50, pp. 92--111, 2019.

\bibitem{kuo2016study}
C.-F.~J. Kuo and Y.~Juang, ``A study on the recognition and classification of embroidered textile defects in manufacturing,'' \emph{Textile Research Journal}, vol.~86, no.~4, pp. 393--408, 2016.

\bibitem{chang2012development}
C.-W. Chang, T.-M. Chao, J.-T. Horng, C.-F. Lu, and R.-H. Yeh, ``Development pattern recognition model for the classification of circuit probe wafer maps on semiconductors,'' \emph{IEEE Transactions on Components, Packaging and Manufacturing Technology}, vol.~2, no.~12, pp. 2089--2097, 2012.

\bibitem{vogl2014standards}
G.~W. Vogl, B.~A. Weiss, and M.~A. Donmez, ``Standards for prognostics and health management (phm) techniques within manufacturing operations,'' National Institute of Standards and Technology Gaithersburg United States, Tech. Rep., 2014.

\bibitem{shin2018framework}
I.~Shin, J.~Lee, J.~Y. Lee, K.~Jung, D.~Kwon, B.~D. Youn, H.~S. Jang, and J.-H. Choi, ``A framework for prognostics and health management applications toward smart manufacturing systems,'' \emph{International Journal of Precision Engineering and Manufacturing-Green Technology}, vol.~5, no.~4, pp. 535--554, 2018.

\bibitem{kumar2018big}
A.~Kumar, R.~Shankar, and L.~S. Thakur, ``A big data driven sustainable manufacturing framework for condition-based maintenance prediction,'' \emph{Journal of computational science}, vol.~27, pp. 428--439, 2018.

\bibitem{rastegari2014implementation}
A.~Rastegari and M.~Bengtsson, ``Implementation of condition based maintenance in manufacturing industry-a pilot case study,'' in \emph{2014 International Conference on Prognostics and Health Management}.\hskip 1em plus 0.5em minus 0.4em\relax IEEE, 2014, pp. 1--8.

\bibitem{wang2017new}
J.~Wang, L.~Zhang, L.~Duan, and R.~X. Gao, ``A new paradigm of cloud-based predictive maintenance for intelligent manufacturing,'' \emph{Journal of Intelligent Manufacturing}, vol.~28, no.~5, pp. 1125--1137, 2017.

\bibitem{he2017integrated}
Y.~He, C.~Gu, Z.~Chen, and X.~Han, ``Integrated predictive maintenance strategy for manufacturing systems by combining quality control and mission reliability analysis,'' \emph{International Journal of Production Research}, vol.~55, no.~19, pp. 5841--5862, 2017.

\bibitem{xia2018recent}
T.~Xia, Y.~Dong, L.~Xiao, S.~Du, E.~Pan, and L.~Xi, ``Recent advances in prognostics and health management for advanced manufacturing paradigms,'' \emph{Reliability Engineering \& System Safety}, vol. 178, pp. 255--268, 2018.

\bibitem{luo2008modelprognostic}
J.~Luo, K.~R. Pattipati, L.~Qiao, and S.~Chigusa, ``Model-based prognostic techniques applied to a suspension system,'' \emph{IEEE Transactions on Systems, Man, and Cybernetics-Part A: Systems and Humans}, vol.~38, no.~5, pp. 1156--1168, 2008.

\bibitem{liu2015novelhiddensemiMarkov}
Q.~Liu, M.~Dong, W.~Lv, X.~Geng, and Y.~Li, ``A novel method using adaptive hidden semi-markov model for multi-sensor monitoring equipment health prognosis,'' \emph{Mechanical Systems and Signal Processing}, vol.~64, pp. 217--232, 2015.

\bibitem{soualhi2016hiddenMarkov}
A.~Soualhi, G.~Clerc, H.~Razik, F.~Guillet \emph{et~al.}, ``Hidden markov models for the prediction of impending faults,'' \emph{IEEE Transactions on Industrial Electronics}, vol.~63, no.~5, pp. 3271--3281, 2016.

\bibitem{qian2017multiScaleApproach}
Y.~Qian, R.~Yan, and R.~X. Gao, ``A multi-time scale approach to remaining useful life prediction in rolling bearing,'' \emph{Mechanical Systems and Signal Processing}, vol.~83, pp. 549--567, 2017.

\bibitem{SiemensAndGENERA2022}
\BIBentryALTinterwordspacing
{Siemens}, ``Siemens and genera jointly accelerate the transformation to industrial serial applications in the field of additive manufacturing via digital light processing,'' Website, November 2022. [Online]. Available: \url{https://press.siemens.com/global/en/pressrelease/siemens-and-genera-jointly-accelerate-transformation-industrial-serial-applications}
\BIBentrySTDinterwordspacing

\bibitem{SiemensAndIntel2023}
\BIBentryALTinterwordspacing
------, ``Siemens and intel to collaborate on advanced semiconductor manufacturing,'' Website, Decmber 2023. [Online]. Available: \url{https://press.siemens.com/global/en/pressrelease/siemens-and-intel-collaborate-advanced-semiconductor-manufacturing}
\BIBentrySTDinterwordspacing

\bibitem{IFRchapter1}
\BIBentryALTinterwordspacing
{IFR, International Federation of Robotics}, ``World robotics industrial robots 2020: Chapter 1 reviews definitions and classifications of industrial robots and service robots.'' Website. [Online]. Available: \url{https://ifr.org/img/worldrobotics/WR_Industrial_Robots_2020_Chapter_1.pdf}
\BIBentrySTDinterwordspacing

\bibitem{jain1995machine}
R.~Jain, R.~Kasturi, and B.~G. Schunck, \emph{Machine vision}.\hskip 1em plus 0.5em minus 0.4em\relax McGraw-hill New York, 1995, vol.~5.

\bibitem{davies2004machine}
E.~R. Davies, \emph{Machine vision: theory, algorithms, practicalities}.\hskip 1em plus 0.5em minus 0.4em\relax Elsevier, 2004.

\bibitem{liu2022deep}
R.~W. Liu, Y.~Guo, Y.~Lu, K.~T. Chui, and B.~B. Gupta, ``Deep network-enabled haze visibility enhancement for visual iot-driven intelligent transportation systems,'' \emph{IEEE Transactions on Industrial Informatics}, vol.~19, no.~2, pp. 1581--1591, 2022.

\bibitem{hashimoto2017current}
M.~Hashimoto, Y.~Domae, and S.~Kaneko, ``Current status and future trends on robot vision technology,'' \emph{Journal of Robotics and Mechatronics}, vol.~29, no.~2, pp. 275--286, 2017.

\bibitem{qi2010vision}
Q.~Qi and R.~Du, ``A vision based micro-assembly system for assembling components in mechanical watch movements,'' in \emph{2010 International Symposium on Optomechatronic Technologies}.\hskip 1em plus 0.5em minus 0.4em\relax IEEE, 2010, pp. 1--5.

\bibitem{chang2018robotic}
W.-C. Chang, ``Robotic assembly of smartphone back shells with eye-in-hand visual servoing,'' \emph{Robotics and Computer-Integrated Manufacturing}, vol.~50, pp. 102--113, 2018.

\bibitem{song2020robotic}
R.~Song, F.~Li, T.~Fu, and J.~Zhao, ``A robotic automatic assembly system based on vision,'' \emph{Applied Sciences}, vol.~10, no.~3, p. 1157, 2020.

\bibitem{zhang2015region}
B.~Zhang, H.~Yang, and Z.~Yin, ``A region-based normalized cross correlation algorithm for the vision-based positioning of elongated ic chips,'' \emph{IEEE Transactions on Semiconductor Manufacturing}, vol.~28, no.~3, pp. 345--352, 2015.

\bibitem{zhong2017blob}
F.~Zhong, S.~He, and B.~Li, ``Blob analyzation-based template matching algorithm for led chip localization,'' \emph{The International Journal of Advanced Manufacturing Technology}, vol.~93, no.~1, pp. 55--63, 2017.

\bibitem{tam1999robotic}
H.-y. Tam, O.~C.-h. Lui, and A.~C. Mok, ``Robotic polishing of free-form surfaces using scanning paths,'' \emph{Journal of Materials Processing Technology}, vol.~95, no. 1-3, pp. 191--200, 1999.

\bibitem{zhu2020robotic}
D.~Zhu, X.~Feng, X.~Xu, Z.~Yang, W.~Li, S.~Yan, and H.~Ding, ``Robotic grinding of complex components: a step towards efficient and intelligent machining--challenges, solutions, and applications,'' \emph{Robotics and Computer-Integrated Manufacturing}, vol.~65, p. 101908, 2020.

\bibitem{wang2020intelligent}
B.~Wang, S.~J. Hu, L.~Sun, and T.~Freiheit, ``Intelligent welding system technologies: State-of-the-art review and perspectives,'' \emph{Journal of Manufacturing Systems}, vol.~56, pp. 373--391, 2020.

\bibitem{du2019strong}
R.~Du, Y.~Xu, Z.~Hou, J.~Shu, and S.~Chen, ``Strong noise image processing for vision-based seam tracking in robotic gas metal arc welding,'' \emph{The International Journal of Advanced Manufacturing Technology}, vol. 101, no.~5, pp. 2135--2149, 2019.

\bibitem{Jamrozik2021assessing}
\BIBentryALTinterwordspacing
W.~Jamrozik and J.~Górka, ``Assessing mma welding process stability using machine vision-based arc features tracking system,'' \emph{Sensors}, vol.~21, no.~1, 2021. [Online]. Available: \url{https://www.mdpi.com/1424-8220/21/1/84}
\BIBentrySTDinterwordspacing

\bibitem{9522071}
R.~W.~L. Coutinho and A.~Boukerche, ``Transfer learning for disruptive 5g-enabled industrial internet of things,'' \emph{IEEE Transactions on Industrial Informatics}, pp. 1--1, 2021.

\bibitem{9272839}
P.~Yu, M.~Yang, A.~Xiong, Y.~Ding, W.~Li, X.~Qiu, L.~Meng, M.~Kadoch, and M.~Cheriet, ``Intelligent-driven green resource allocation for industrial internet of things in 5g heterogeneous networks,'' \emph{IEEE Transactions on Industrial Informatics}, pp. 1--1, 2020.

\bibitem{huawei5Gcase}
\BIBentryALTinterwordspacing
Huawei, ``Bring 5.5g into reality,'' Website, 2023. [Online]. Available: \url{https://www-file.huawei.com/-/media/CORP2020/media-center/pdf/5G-advanced-brochure-EN-final.pdf}
\BIBentrySTDinterwordspacing

\bibitem{8695835}
L.~Lo~Bello and W.~Steiner, ``A perspective on ieee time-sensitive networking for industrial communication and automation systems,'' \emph{Proceedings of the IEEE}, vol. 107, no.~6, pp. 1094--1120, 2019.

\bibitem{8610105}
D.~Bruckner, ``An introduction to opc ua tsn for industrial communication systems,'' \emph{Proceedings of the IEEE}, vol. 107, no.~6, pp. 1121--1131, 2019.

\bibitem{9142734}
M.~A. Metaal, R.~Guillaume, R.~Steinmetz, and A.~Rizk, ``Integrated industrial ethernet networks: Time-sensitive networking over sdn infrastructure for mixed applications,'' in \emph{2020 IFIP Networking Conference (Networking)}, 2020, pp. 803--808.

\bibitem{8715451}
S.~Vitturi, C.~Zunino, and T.~Sauter, ``Industrial communication systems and their future challenges: Next-generation ethernet, iiot, and 5g,'' \emph{Proceedings of the IEEE}, vol. 107, no.~6, pp. 944--961, 2019.

\bibitem{moxa2023}
\BIBentryALTinterwordspacing
{Moxa}, ``How time-sensitive networking is being applied in real world manufacturing,'' Website, March 2023. [Online]. Available: \url{https://iebmedia.com/technology/tsn/how-tsn-as-being-applied-in-real-world-manufacturing/}
\BIBentrySTDinterwordspacing

\bibitem{yu2021digital}
J.~Yu, Y.~Song, D.~Tang, and J.~Dai, ``A digital twin approach based on nonparametric bayesian network for complex system health monitoring,'' \emph{Journal of Manufacturing Systems}, vol.~58, pp. 293--304, 2021.

\bibitem{kiswanto2020digital}
G.~Kiswanto \emph{et~al.}, ``Digital twin approach for tool wear monitoring of micro-milling,'' \emph{Procedia CIRP}, vol.~93, pp. 1532--1537, 2020.

\bibitem{ladj2021knowledge}
A.~Ladj, Z.~Wang, O.~Meski, F.~Belkadi, M.~Ritou, and C.~Da~Cunha, ``A knowledge-based digital shadow for machining industry in a digital twin perspective,'' \emph{Journal of Manufacturing Systems}, vol.~58, pp. 168--179, 2021.

\bibitem{deebak2021digital}
B.~Deebak and F.~Al-Turjman, ``Digital-twin assisted: Fault diagnosis using deep transfer learning for machining tool condition,'' \emph{International Journal of Intelligent Systems}, 2021.

\bibitem{liu2021multi}
S.~Liu, Y.~Lu, J.~Li, D.~Song, X.~Sun, and J.~Bao, ``Multi-scale evolution mechanism and knowledge construction of a digital twin mimic model,'' \emph{Robotics and Computer-Integrated Manufacturing}, vol.~71, p. 102123, 2021.

\bibitem{jiang2021model}
H.~Jiang, S.~Qin, J.~Fu, J.~Zhang, and G.~Ding, ``How to model and implement connections between physical and virtual models for digital twin application,'' \emph{Journal of Manufacturing Systems}, vol.~58, pp. 36--51, 2021.

\bibitem{lecun2015deep}
Y.~LeCun, Y.~Bengio, and G.~Hinton, ``Deep learning,'' \emph{nature}, vol. 521, no. 7553, pp. 436--444, 2015.

\bibitem{rusk2016deep}
N.~Rusk, ``Deep learning,'' \emph{Nature Methods}, vol.~13, no.~1, pp. 35--35, 2016.

\bibitem{hu2020reinforcement}
Y.~Hu, Y.~Yao, and W.~S. Lee, ``A reinforcement learning approach for optimizing multiple traveling salesman problems over graphs,'' \emph{Knowledge-Based Systems}, vol. 204, p. 106244, 2020.

\bibitem{hu2021bidirectional}
Y.~Hu, Z.~Zhang, Y.~Yao, X.~Huyan, X.~Zhou, and W.~S. Lee, ``A bidirectional graph neural network for traveling salesman problems on arbitrary symmetric graphs,'' \emph{Engineering Applications of Artificial Intelligence}, vol.~97, p. 104061, 2021.

\bibitem{zhang2022truncated}
Z.~Zhang, I.~Ng, D.~Gong, Y.~Liu, E.~Abbasnejad, M.~Gong, K.~Zhang, and J.~Q. Shi, ``Truncated matrix power iteration for differentiable dag learning,'' \emph{Advances in Neural Information Processing Systems}, vol.~35, pp. 18\,390--18\,402, 2022.

\bibitem{sun2023learning}
Q.~Sun, Y.~Yao, P.~Yi, Y.~Hu, Z.~Yang, G.~Yang, and X.~Zhou, ``Learning controlled and targeted communication with the centralized critic for the multi-agent system,'' \emph{Applied Intelligence}, vol.~53, no.~12, pp. 14\,819--14\,837, 2023.

\bibitem{chen2023global}
J.~Chen, T.~Li, Y.~Zhang, T.~You, Y.~Lu, P.~Tiwari, and N.~Kumar, ``Global-and-local attention-based reinforcement learning for cooperative behaviour control of multiple uavs,'' \emph{IEEE Transactions on Vehicular Technology}, 2023.

\bibitem{wang2016edge}
R.~Wang, ``Edge detection using convolutional neural network,'' in \emph{International Symposium on Neural Networks}.\hskip 1em plus 0.5em minus 0.4em\relax Springer, 2016, pp. 12--20.

\bibitem{he2020bdcn}
J.~He, S.~Zhang, M.~Yang, Y.~Shan, and T.~Huang, ``Bdcn: Bi-directional cascade network for perceptual edge detection,'' \emph{IEEE Transactions on Pattern Analysis and Machine Intelligence}, 2020.

\bibitem{van2019deep}
S.~Van~der Jeught and J.~J. Dirckx, ``Deep neural networks for single shot structured light profilometry,'' \emph{Optics express}, vol.~27, no.~12, pp. 17\,091--17\,101, 2019.

\bibitem{wang2021single}
F.~Wang, C.~Wang, and Q.~Guan, ``Single-shot fringe projection profilometry based on deep learning and computer graphics,'' \emph{Optics Express}, vol.~29, no.~6, pp. 8024--8040, 2021.

\bibitem{lee2017piv}
Y.~Lee, H.~Yang, and Z.~Yin, ``Piv-dcnn: cascaded deep convolutional neural networks for particle image velocimetry,'' \emph{Experiments in Fluids}, vol.~58, no.~12, p. 171, 2017.

\bibitem{lee2021diffeomorphic}
Y.~Lee and S.~Mei, ``Diffeomorphic particle image velocimetry,'' \emph{arXiv preprint arXiv:2108.07438}, 2021.

\bibitem{boukhtache2021deep}
S.~Boukhtache, K.~Abdelouahab, F.~Berry, B.~Blaysat, M.~Grediac, and F.~Sur, ``When deep learning meets digital image correlation,'' \emph{Optics and Lasers in Engineering}, vol. 136, p. 106308, 2021.

\bibitem{mei2018unsupervised}
S.~Mei, H.~Yang, and Z.~Yin, ``An unsupervised-learning-based approach for automated defect inspection on textured surfaces,'' \emph{IEEE Transactions on Instrumentation and Measurement}, vol.~67, no.~6, pp. 1266--1277, 2018.

\bibitem{wei2021real}
W.~Wei, D.~Deng, L.~Zeng, and C.~Zhang, ``Real-time implementation of fabric defect detection based on variational automatic encoder with structure similarity,'' \emph{Journal of Real-Time Image Processing}, vol.~18, no.~3, pp. 807--823, 2021.

\bibitem{hu2020unsupervised}
G.~Hu, J.~Huang, Q.~Wang, J.~Li, Z.~Xu, and X.~Huang, ``Unsupervised fabric defect detection based on a deep convolutional generative adversarial network,'' \emph{Textile Research Journal}, vol.~90, no. 3-4, pp. 247--270, 2020.

\bibitem{sohn2020learning}
K.~Sohn, C.-L. Li, J.~Yoon, M.~Jin, and T.~Pfister, ``Learning and evaluating representations for deep one-class classification,'' \emph{arXiv preprint arXiv:2011.02578}, 2020.

\bibitem{li2021cutpaste}
C.-L. Li, K.~Sohn, J.~Yoon, and T.~Pfister, ``Cutpaste: Self-supervised learning for anomaly detection and localization,'' in \emph{Proceedings of the IEEE/CVF Conference on Computer Vision and Pattern Recognition}, 2021, pp. 9664--9674.

\bibitem{lehr2021supervised}
J.~Lehr, J.~Philipps, V.~N. Hoang, D.~von Wrangel, and J.~Kr{\"u}ger, ``Supervised learning vs. unsupervised learning: A comparison for optical inspection applications in quality control,'' in \emph{IOP Conference Series: Materials Science and Engineering}, vol. 1140, no.~1.\hskip 1em plus 0.5em minus 0.4em\relax IOP Publishing, 2021, p. 012049.

\bibitem{zheng1991strategies}
Y.~F. Zheng, R.~Pei, and C.~Chen, ``Strategies for automatic assembly of deformable objects,'' in \emph{Proceedings. 1991 IEEE International Conference on Robotics and Automation}.\hskip 1em plus 0.5em minus 0.4em\relax IEEE Computer Society, 1991, pp. 2598--2599.

\bibitem{jimenez2012survey}
P.~Jim{\'e}nez, ``Survey on model-based manipulation planning of deformable objects,'' \emph{Robotics and computer-integrated manufacturing}, vol.~28, no.~2, pp. 154--163, 2012.

\bibitem{hafner2019dream}
D.~Hafner, T.~Lillicrap, J.~Ba, and M.~Norouzi, ``Dream to control: Learning behaviors by latent imagination,'' \emph{arXiv preprint arXiv:1912.01603}, 2019.

\bibitem{Chen2021MuMMI}
K.~Chen, Y.~Lee, and H.~Soh, ``Multi-modal mutual information (mummi) training for robust self-supervised deep reinforcement learning,'' in \emph{IEEE International Conference on Robotics and Automation (ICRA)}, 2021.

\bibitem{lin2020softgym}
X.~Lin, Y.~Wang, J.~Olkin, and D.~Held, ``Softgym: Benchmarking deep reinforcement learning for deformable object manipulation,'' \emph{arXiv preprint arXiv:2011.07215}, 2020.

\bibitem{hafner2019learning}
D.~Hafner, T.~Lillicrap, I.~Fischer, R.~Villegas, D.~Ha, H.~Lee, and J.~Davidson, ``Learning latent dynamics for planning from pixels,'' in \emph{International Conference on Machine Learning}.\hskip 1em plus 0.5em minus 0.4em\relax PMLR, 2019, pp. 2555--2565.

\bibitem{dulac2019challenges}
G.~Dulac-Arnold, D.~Mankowitz, and T.~Hester, ``Challenges of real-world reinforcement learning,'' \emph{arXiv preprint arXiv:1904.12901}, 2019.

\bibitem{kostrikov2020image}
I.~Kostrikov, D.~Yarats, and R.~Fergus, ``Image augmentation is all you need: Regularizing deep reinforcement learning from pixels,'' \emph{arXiv preprint arXiv:2004.13649}, 2020.

\bibitem{belousov2021architectural}
J.~S. T. S.~B. Belousov and G.~C. S. T.~B. Wibranek, ``Architectural assembly with tactile skills: Simulation and optimization,'' 2021.

\bibitem{coffey2018machine}
V.~C. Coffey, ``Machine vision: The eyes of industry 4.0,'' \emph{Optics and photonics news}, vol.~29, no.~7, pp. 42--49, 2018.

\bibitem{kim2008simultaneous}
H.~Kim, A.~Handa, R.~Benosman, S.-H. Ieng, and A.~J. Davison, ``Simultaneous mosaicing and tracking with an event camera,'' \emph{J. Solid State Circ}, vol.~43, pp. 566--576, 2008.

\bibitem{taunyazov20event}
T.~Taunyazoz, W.~Sng, H.~H. See, B.~Lim, J.~Kuan, A.~F. Ansari, B.~Tee, and H.~Soh, ``Event-driven visual-tactile sensing and learning for robots,'' in \emph{Proceedings of Robotics: Science and Systems}, July 2020.

\bibitem{chi2020dynamic}
Y.~Chi, A.~Gnanasambandam, V.~Koltun, and S.~H. Chan, ``Dynamic low-light imaging with quanta image sensors,'' in \emph{Computer Vision--ECCV 2020: 16th European Conference, Glasgow, UK, August 23--28, 2020, Proceedings, Part XXI 16}.\hskip 1em plus 0.5em minus 0.4em\relax Springer, 2020, pp. 122--138.

\bibitem{gnanasambandam2020hdr}
A.~Gnanasambandam and S.~H. Chan, ``Hdr imaging with quanta image sensors: Theoretical limits and optimal reconstruction,'' \emph{IEEE Transactions on Computational Imaging}, vol.~6, pp. 1571--1585, 2020.

\bibitem{mait2018computational}
J.~N. Mait, G.~W. Euliss, and R.~A. Athale, ``Computational imaging,'' \emph{Advances in Optics and Photonics}, vol.~10, no.~2, pp. 409--483, 2018.

\bibitem{van2020nondestructive}
T.~Van De~Looverbosch, M.~H.~R. Bhuiyan, P.~Verboven, M.~Dierick, D.~Van~Loo, J.~De~Beenbouwer, J.~Sijbers, and B.~Nicola{\"\i}, ``Nondestructive internal quality inspection of pear fruit by x-ray ct using machine learning,'' \emph{Food Control}, vol. 113, p. 107170, 2020.

\bibitem{atefi2020robotic}
A.~Atefi, Y.~Ge, S.~Pitla, and J.~Schnable, ``Robotic detection and grasp of maize and sorghum: Stem measurement with contact,'' \emph{Robotics}, vol.~9, no.~3, p.~58, 2020.

\bibitem{zhang20203d}
H.~Zhang and S.-B. Wen, ``3d photolithography through light field projections,'' \emph{Applied optics}, vol.~59, no.~27, pp. 8071--8076, 2020.

\bibitem{iacono2021computational}
S.~D. Iacono, G.~Di~Leo, and C.~Liguori, ``Computational imaging for drill bit wear estimation,'' in \emph{2021 IEEE International Instrumentation and Measurement Technology Conference (I2MTC)}.\hskip 1em plus 0.5em minus 0.4em\relax IEEE, 2021, pp. 1--6.

\bibitem{shalf2020future}
J.~Shalf, ``The future of computing beyond moore’s law,'' \emph{Philosophical Transactions of the Royal Society A}, vol. 378, no. 2166, p. 20190061, 2020.

\bibitem{leiserson2020there}
C.~E. Leiserson, N.~C. Thompson, J.~S. Emer, B.~C. Kuszmaul, B.~W. Lampson, D.~Sanchez, and T.~B. Schardl, ``There’s plenty of room at the top: What will drive computer performance after moore’s law?'' \emph{Science}, vol. 368, no. 6495, 2020.

\bibitem{qin2017gpu}
Y.~Qin, X.~Jin, and Q.~Dai, ``Gpu-based depth estimation for light field images,'' in \emph{2017 International Symposium on Intelligent Signal Processing and Communication Systems (ISPACS)}.\hskip 1em plus 0.5em minus 0.4em\relax IEEE, 2017, pp. 640--645.

\bibitem{yu2019gpu}
X.~Yu, H.~Wang, W.-c. Feng, H.~Gong, and G.~Cao, ``Gpu-based iterative medical ct image reconstructions,'' \emph{Journal of Signal Processing Systems}, vol.~91, no.~3, pp. 321--338, 2019.

\bibitem{you2018imagenet}
Y.~You, Z.~Zhang, C.-J. Hsieh, J.~Demmel, and K.~Keutzer, ``Imagenet training in minutes,'' in \emph{Proceedings of the 47th International Conference on Parallel Processing}, 2018, pp. 1--10.

\bibitem{you2019large}
Y.~You, J.~Li, S.~Reddi, J.~Hseu, S.~Kumar, S.~Bhojanapalli, X.~Song, J.~Demmel, K.~Keutzer, and C.-J. Hsieh, ``Large batch optimization for deep learning: Training bert in 76 minutes,'' \emph{arXiv preprint arXiv:1904.00962}, 2019.

\bibitem{kumar2019comprehensive}
M.~Kumar, S.~C. Sharma, A.~Goel, and S.~P. Singh, ``A comprehensive survey for scheduling techniques in cloud computing,'' \emph{Journal of Network and Computer Applications}, vol. 143, pp. 1--33, 2019.

\bibitem{dang2019survey}
L.~M. Dang, M.~J. Piran, D.~Han, K.~Min, and H.~Moon, ``A survey on internet of things and cloud computing for healthcare,'' \emph{Electronics}, vol.~8, no.~7, p. 768, 2019.

\bibitem{arunarani2019task}
A.~Arunarani, D.~Manjula, and V.~Sugumaran, ``Task scheduling techniques in cloud computing: A literature survey,'' \emph{Future Generation Computer Systems}, vol.~91, pp. 407--415, 2019.

\bibitem{katal2023energy}
A.~Katal, S.~Dahiya, and T.~Choudhury, ``Energy efficiency in cloud computing data centers: a survey on software technologies,'' \emph{Cluster Computing}, vol.~26, no.~3, pp. 1845--1875, 2023.

\bibitem{luo2021resource}
Q.~Luo, S.~Hu, C.~Li, G.~Li, and W.~Shi, ``Resource scheduling in edge computing: A survey,'' \emph{IEEE Communications Surveys \& Tutorials}, vol.~23, no.~4, pp. 2131--2165, 2021.

\bibitem{wang2020convergence}
X.~Wang, Y.~Han, V.~C. Leung, D.~Niyato, X.~Yan, and X.~Chen, ``Convergence of edge computing and deep learning: A comprehensive survey,'' \emph{IEEE Communications Surveys \& Tutorials}, vol.~22, no.~2, pp. 869--904, 2020.

\bibitem{liu2019survey}
F.~Liu, G.~Tang, Y.~Li, Z.~Cai, X.~Zhang, and T.~Zhou, ``A survey on edge computing systems and tools,'' \emph{Proceedings of the IEEE}, vol. 107, no.~8, pp. 1537--1562, 2019.

\bibitem{helo2021cloud}
P.~Helo, Y.~Hao, R.~Toshev, and V.~Boldosova, ``Cloud manufacturing ecosystem analysis and design,'' \emph{Robotics and Computer-Integrated Manufacturing}, vol.~67, p. 102050, 2021.

\bibitem{ghomi2019cloud}
E.~J. Ghomi, A.~M. Rahmani, and N.~N. Qader, ``Cloud manufacturing: challenges, recent advances, open research issues, and future trends,'' \emph{The International Journal of Advanced Manufacturing Technology}, vol. 102, pp. 3613--3639, 2019.

\bibitem{adamson2017cloud}
G.~Adamson, L.~Wang, M.~Holm, and P.~Moore, ``Cloud manufacturing--a critical review of recent development and future trends,'' \emph{International Journal of Computer Integrated Manufacturing}, vol.~30, no. 4-5, pp. 347--380, 2017.

\bibitem{liu2014multi}
N.~Liu, X.~Li, and W.~Shen, ``Multi-granularity resource virtualization and sharing strategies in cloud manufacturing,'' \emph{Journal of Network and Computer Applications}, vol.~46, pp. 72--82, 2014.

\bibitem{bouzary2020classification}
H.~Bouzary and F.~F. Chen, ``A classification-based approach for integrated service matching and composition in cloud manufacturing,'' \emph{Robotics and Computer-Integrated Manufacturing}, vol.~66, p. 101989, 2020.

\bibitem{zhang2017research}
Y.~Zhang, G.~Zhang, Y.~Liu, and D.~Hu, ``Research on services encapsulation and virtualization access model of machine for cloud manufacturing,'' \emph{Journal of Intelligent Manufacturing}, vol.~28, pp. 1109--1123, 2017.

\bibitem{wang2021framework}
L.-C. Wang, C.-C. Chen, J.-L. Liu, and P.-C. Chu, ``Framework and deployment of a cloud-based advanced planning and scheduling system,'' \emph{Robotics and Computer-Integrated Manufacturing}, vol.~70, p. 102088, 2021.

\bibitem{okwudire2020three}
C.~E. Okwudire, X.~Lu, G.~Kumaravelu, and H.~Madhyastha, ``A three-tier redundant architecture for safe and reliable cloud-based cnc over public internet networks,'' \emph{Robotics and Computer-Integrated Manufacturing}, vol.~62, p. 101880, 2020.

\bibitem{pundir2020designing}
S.~Pundir, M.~Wazid, D.~P. Singh, A.~K. Das, J.~J. Rodrigues, and Y.~Park, ``Designing efficient sinkhole attack detection mechanism in edge-based iot deployment,'' \emph{Sensors}, vol.~20, no.~5, p. 1300, 2020.

\bibitem{canete2020energy}
A.~Ca{\~n}ete, M.~Amor, and L.~Fuentes, ``Energy-efficient deployment of iot applications in edge-based infrastructures: A software product line approach,'' \emph{IEEE Internet of Things Journal}, vol.~8, no.~22, pp. 16\,427--16\,439, 2020.

\bibitem{canete2022supporting}
------, ``Supporting iot applications deployment on edge-based infrastructures using multi-layer feature models,'' \emph{Journal of Systems and Software}, vol. 183, p. 111086, 2022.

\bibitem{zaureSTM}
\BIBentryALTinterwordspacing
{Microsoft}, ``Stmicroelectronics transforms research and development, manufacturing, supply chain, and internal processes with azure,'' Website, April 2023. [Online]. Available: \url{https://customers.microsoft.com/en-us/story/1624918262220581625-stmicroelectronics-manufacturing-azure-teams-power-platform}
\BIBentrySTDinterwordspacing

\bibitem{zaurePhillips}
\BIBentryALTinterwordspacing
------, ``Phillips corporation modernizes service operations and increases revenue using dynamics 365 field service,'' Website, May 2023. [Online]. Available: \url{https://customers.microsoft.com/en-us/story/1627203871037668308-phillips-corporation-manufacturing-dynamics-365}
\BIBentrySTDinterwordspacing

\bibitem{hu20223cl}
Y.~Hu, Q.~Jia, H.~Liu, X.~Zhou, H.~Lai, and R.~Xie, ``3cl-net: A four-in-one networking paradigm for 6g system,'' in \emph{2022 5th International Conference on Hot Information-Centric Networking (HotICN)}.\hskip 1em plus 0.5em minus 0.4em\relax IEEE, 2022, pp. 132--136.

\bibitem{QingminDeterChina2022}
\BIBentryALTinterwordspacing
Q.~Jia, Y.~Hu, H.~Zhang, K.~Peng, P.~Chen, R.~Xie, and T.~Huang, ``Research on deterministic computing power network,'' \emph{Journal on Communications}, vol.~43, no.~10, p.~55, 2022. [Online]. Available: \url{https://www.infocomm-journal.com/txxb/EN/abstract/article_172861.shtml}
\BIBentrySTDinterwordspacing

\bibitem{YujiaoChina}
\BIBentryALTinterwordspacing
Y.~Hu, Q.~Jia, Q.~Sun, R.~Xie, and T.~Huang, ``Functional architecture to intelligent computing power network,'' \emph{Computer Science}, vol.~49, no.~9, pp. 249--259, 2022. [Online]. Available: \url{https://doi.org/10.11896/jsjkx.220500222}
\BIBentrySTDinterwordspacing

\bibitem{jia2024deterministic}
Q.~Jia, Y.~Hu, X.~Zhou, Q.~Ma, K.~Guo, H.~Zhang, R.~Xie, T.~Huang, and Y.~Liu, ``Deterministic computing power networking: Architecture, technologies and prospects,'' \emph{arXiv preprint arXiv:2401.17812}, 2024.

\bibitem{glnworkers24}
\BIBentryALTinterwordspacing
{World Economic Forum}, ``Views from the manufacturing front line: Workers’ insights on how to introduce new technology,'' Website, Jaunary 2024. [Online]. Available: \url{https://www.weforum.org/publications/views-from-the-manufacturing-front-line-workers-insights-on-how-to-introduce-new-technology/}
\BIBentrySTDinterwordspacing

\end{thebibliography}


\begin{IEEEbiography}[{\includegraphics[width=1in,height=1.25in,clip,keepaspectratio]{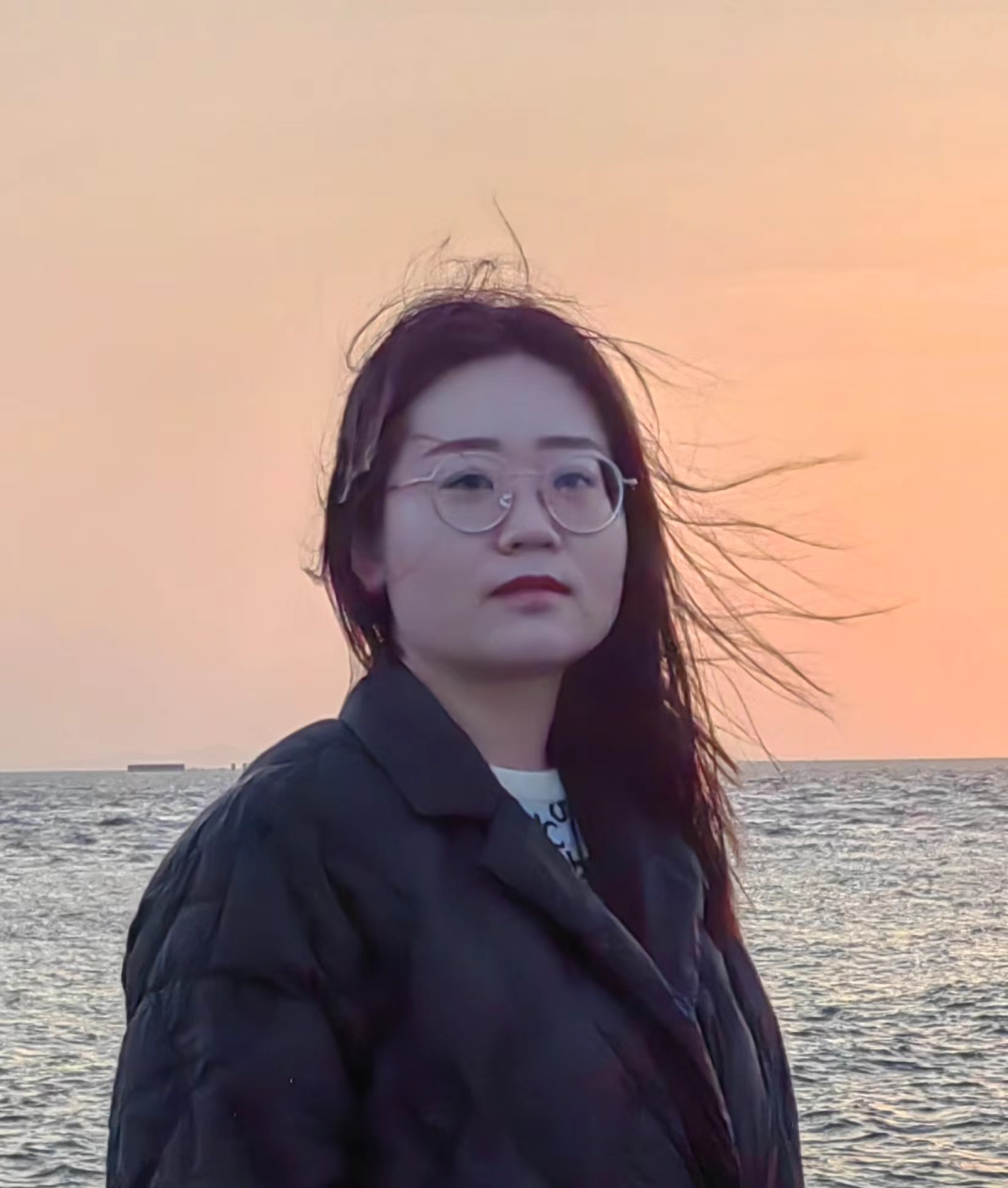}}]{Yujiao Hu}
    (Memeber, IEEE) received her Bachelor and PhD degrees from the Department of Computer Science of Northwestern Polytechnical University, Xi'an, China, in 2016 and 2021 respectively. From Nov. 2018 to March 2020, she was a visiting PhD student in National University of Singapore. Currently, she is a faculty member in Purple Mountain Laboratories. She focuses on deep learning, edge computing, multi-agent cooperation problems and time sensitive networks.    
\end{IEEEbiography}

\begin{IEEEbiography}[{\includegraphics[width=1in,height=1.25in,clip,keepaspectratio]{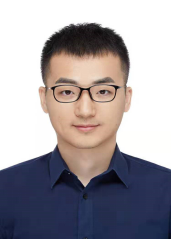}}]{Qingmin Jia}
	(Memeber, IEEE) is currently a Researcher in Future Network Research Center of Purple Mountain Laboratories. He received the B.S. degree from Qingdao University of Technology in 2014, and received the Ph.D. degree from Beijing University of Posts and Telecommunications (BUPT) in 2019. His current research interests include Edge Computing, Edge Intelligence, Industrial Internet of Things and Future Network Architecture. 
\end{IEEEbiography}

\begin{IEEEbiography}[{\includegraphics[width=1in,height=1.25in,clip,keepaspectratio]{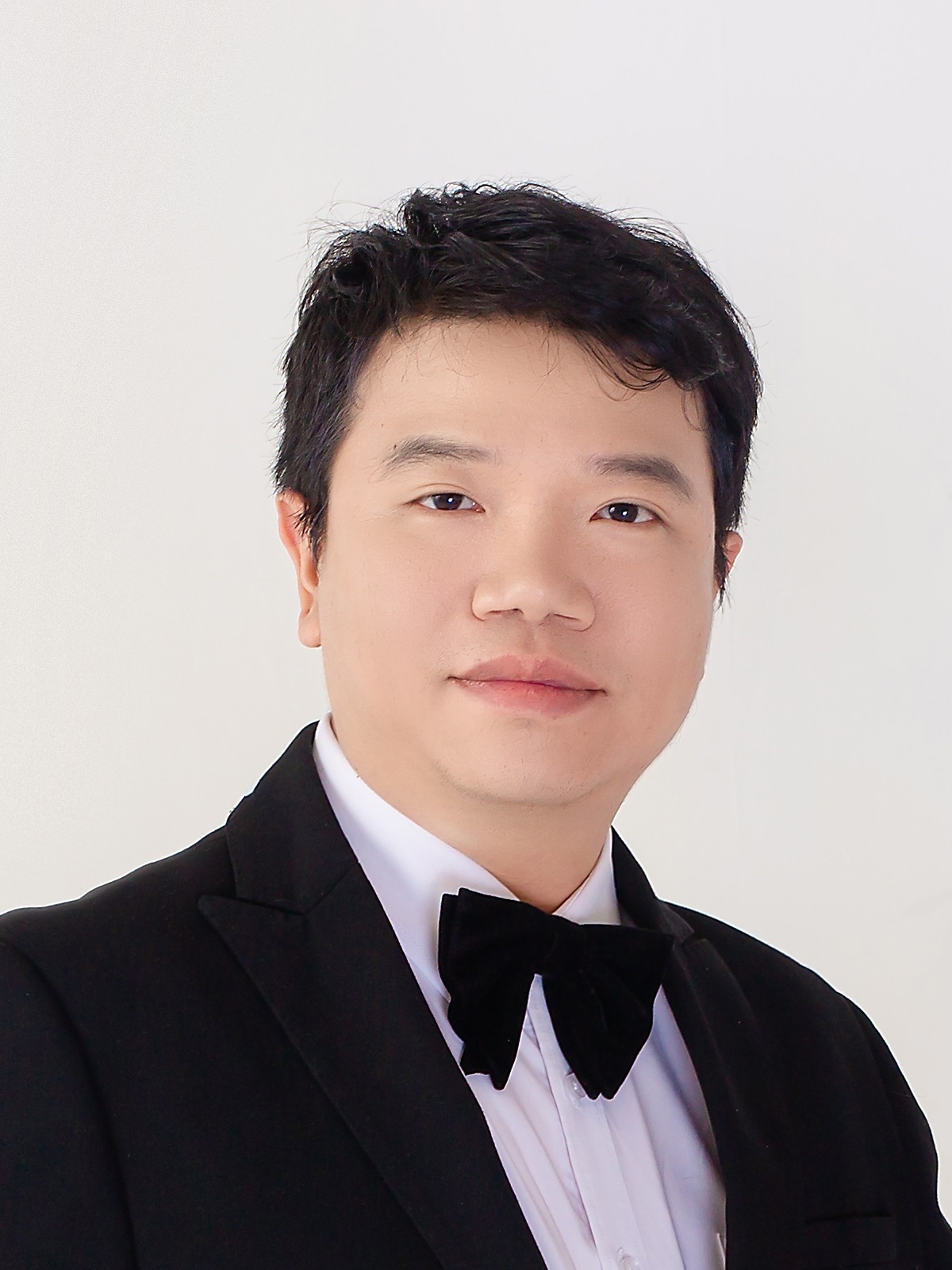}}]{Yuan Yao} (Memeber, IEEE) is currently an Associate Professor in the School of Computer Science, Northwestern Polytechnical University (NPU), Xi'an, China. He received the B.S. M.S., and Ph.D. degrees in computer science from NPU, in 2007, 2009 and 2015, respectively. He was a Postdoctoral Researcher in the Department of Computing at Polytechnic University, Hong Kong from 2016 to 2018. His research interests are in the area of real-time and embedded systems, swarm intelligence operating systems and cyber physical system.
\end{IEEEbiography}

\begin{IEEEbiography}[{\includegraphics[width=1in,height=1.25in,clip,keepaspectratio]{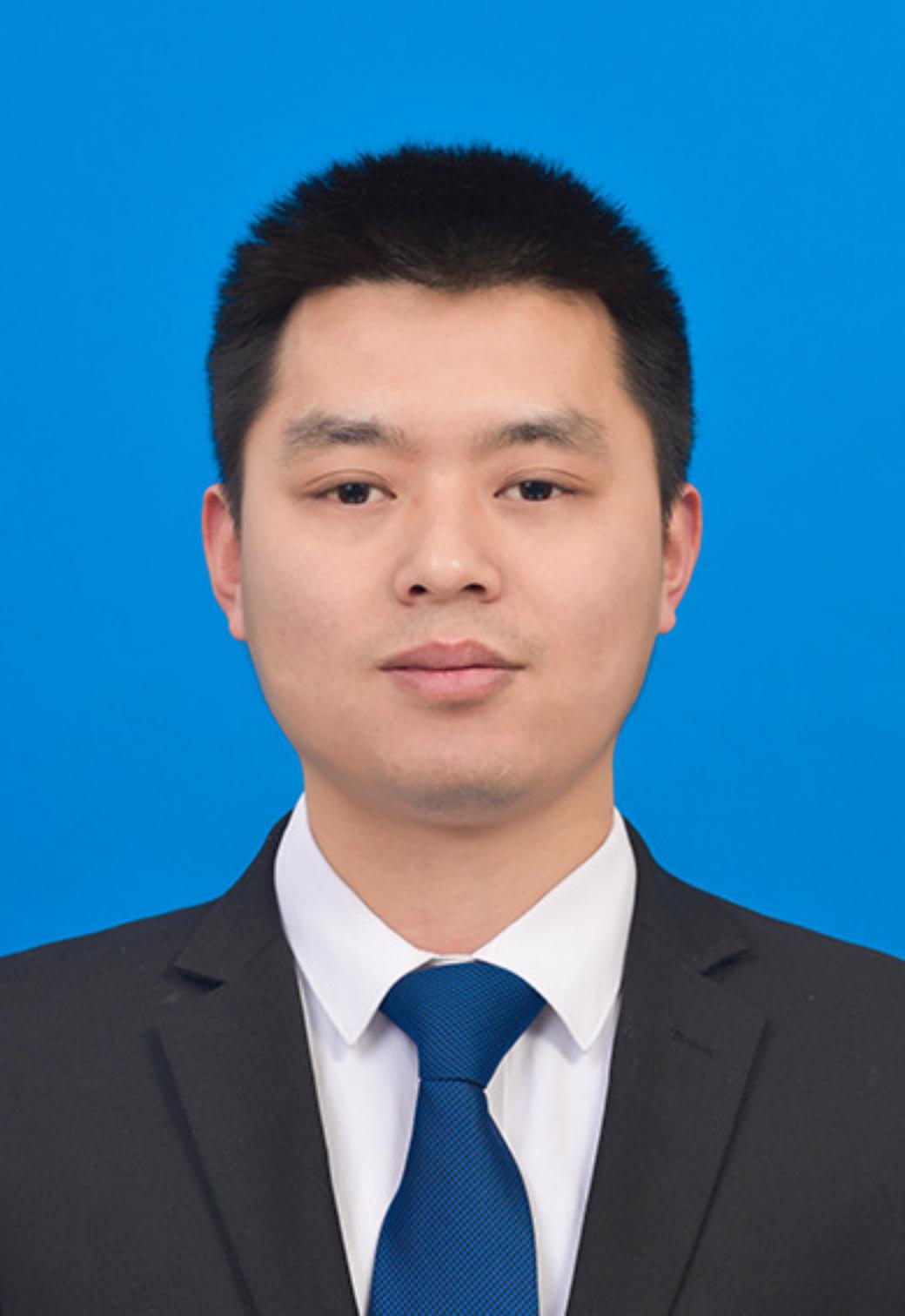}}]{Yong Lee} (Memeber, IEEE) is currently an lecturer of the Wuhan University of Technology since 2021. He received his BSc and Ph.D. degrees from Huazhong University of Science and Technology(HUST), Wuhan, P. R. China, in 2012 and 2018 respectively. He worked at Cobot as an algorithm scientist from 2018 to 2019. He was a Research Fellow with Department of Computer Science at National University of Singapore from 2019 to 2020. His research interests include image processing, particle image velocimetry and robotics.
\end{IEEEbiography}

\begin{IEEEbiography}[{\includegraphics[width=1in,height=1.25in,clip,keepaspectratio]{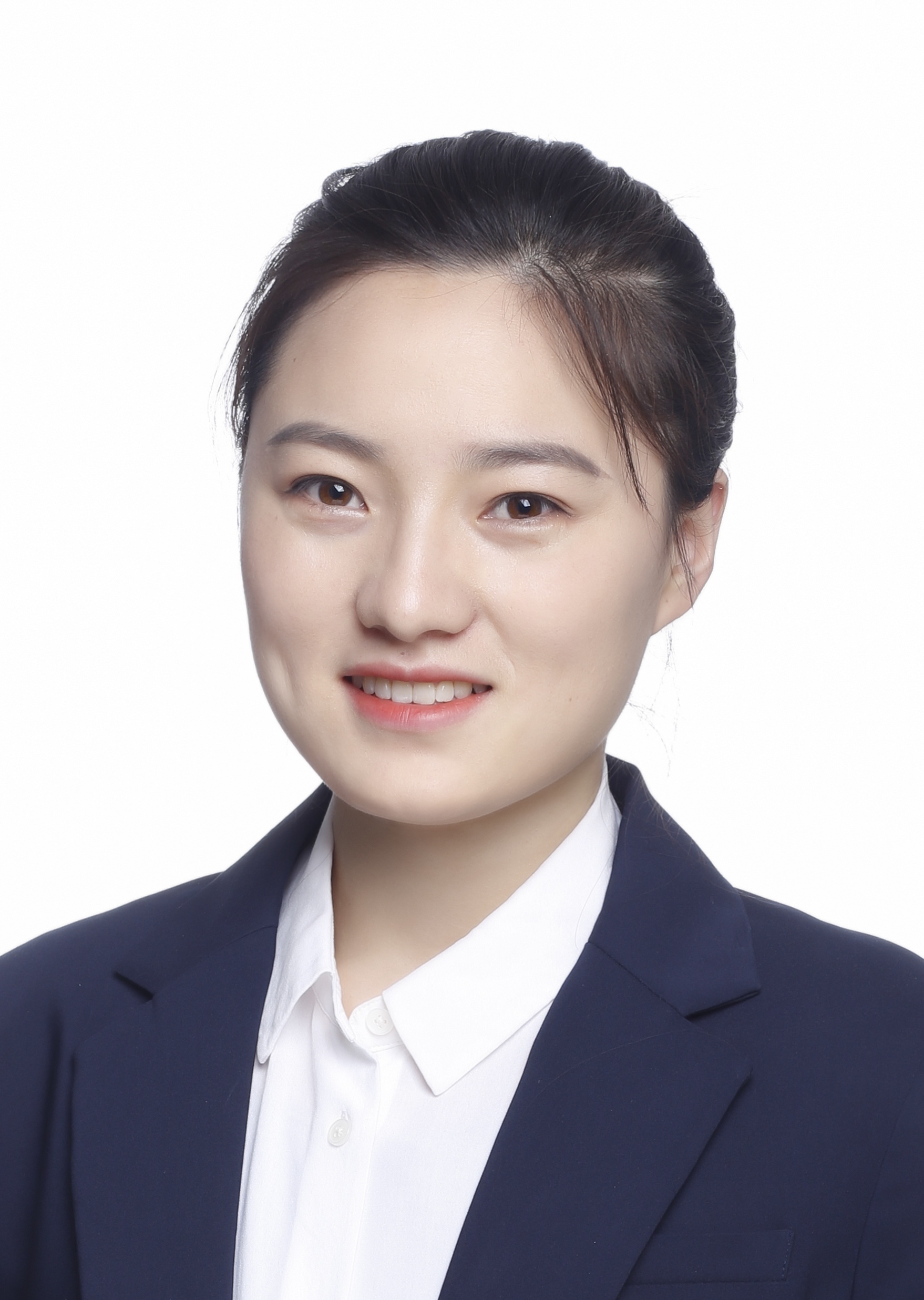}}]{Mengjie Lee} received her Bachelor and Master degrees from School of Computing in the Northwestern Polytechnical University Ming de College in 2018 and the Department of Software, Northwestern Polytechnic University (NPU) in 2021 respectively. She is currently a PhD student in the Department of Computer Science at NPU. She focuses on the accurately modeling and interoperability issues of the digital twin in the industry environment.
\end{IEEEbiography}

\begin{IEEEbiography}[{\includegraphics[width=1in,height=1.25in,clip,keepaspectratio]{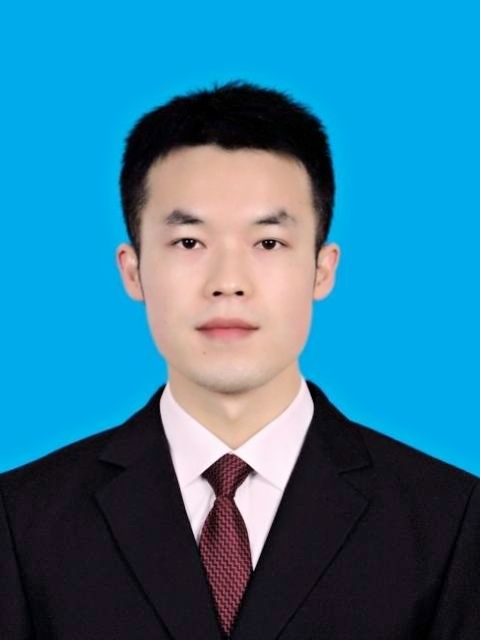}}]{Chenyi Wang} 
	received his Bachelor and Master degree from the Department of Computer Science of Northwestern Polytechnical University, Xi'an, P.R.China, in July2023. He has been a Ph.D. student in the Department of Computer Science of Peking University since September 2023, and his supervisor is Prof.~Gang Huang. His research interests focus on application of large language models in software engineering.
\end{IEEEbiography}

\begin{IEEEbiography}[{\includegraphics[width=1in,height=1.25in,clip,keepaspectratio]{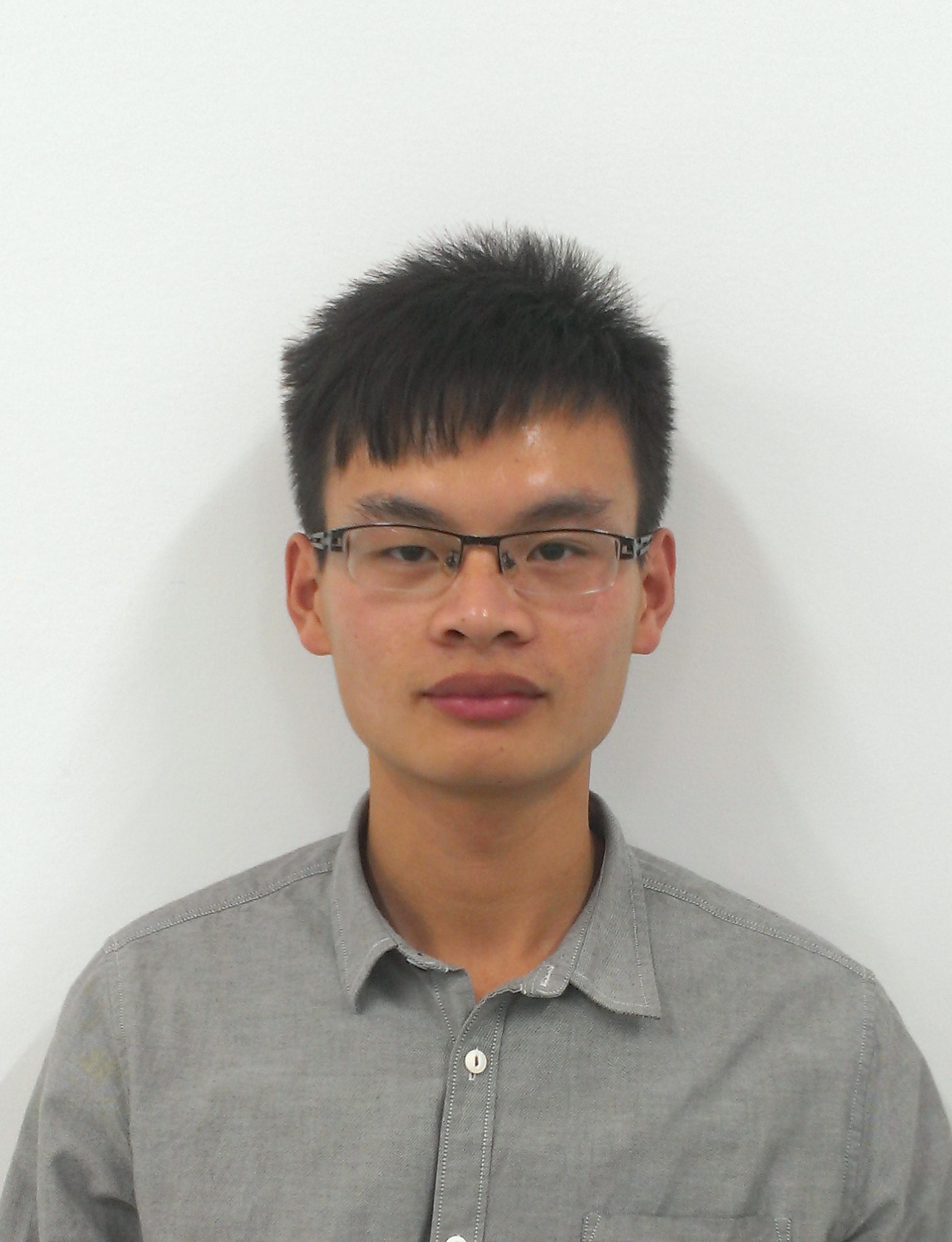}}]{Xiaomao Zhou} (Memeber, IEEE) 
	received his Bachelor and PhD degrees from the Department of Automation of Harbin Engineering University, Harbin, China, in 2014 and 2020, respectively. From Oct. 2016 to Oct. 2018, he was a visiting PhD student in Hamburg University, Germany. Currently, he is a faculty member in Purple Mountain Laboratories. He focuses on deep learning, edge intelligence, and computing network.
\end{IEEEbiography}

\begin{IEEEbiography}[{\includegraphics[width=1in,height=1.25in,clip,keepaspectratio]{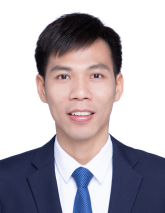}}]{Renchao Xie} 
	(Senior Member, IEEE) received the Ph.D. degree from the School of Information and Communication Engineering, Beijing University of Posts and Telecommunications (BUPT), Beijing, China, in 2012. He is a Professor with BUPT. 
 From November 2010 to November 2011, he visited Carleton University, Ottawa, ON, Canada, as a Visiting Scholar. His current research interests include 5G network and edge computing, information-centric networking, and future network architecture. 
\end{IEEEbiography}

\begin{IEEEbiography}[{\includegraphics[width=1in,height=1.25in,clip,keepaspectratio]{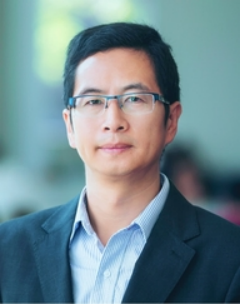}}]{F. Richard Yu } 
	(Fellow, IEEE) received the Ph.D. degree in electrical engineering from the University of British Columbia, Vancouver, BC, Canada, in 2003. From 2002 to 2006, he was with Ericsson, Lund, Sweden, and a start-up in California, USA. He joined Carleton University, Ottawa, ON, Canada, in 2007, where he is currently a Professor. His research interests include wireless cyber–physical systems, connected/autonomous vehicles, security, distributed ledger technology, and deep learning. He is a Distinguished Lecturer, the Vice President (Membership), and an Elected Member of the Board of Governors of the IEEE Vehicular Technology Society. He is a Fellow of the IEEE, Canadian Academy of Engineering (CAE), Engineering Institute of Canada (EIC), and Institution of Engineering and Technology (IET).
\end{IEEEbiography}
\end{document}